\documentclass[english]{article}
\usepackage[T1]{fontenc}
\usepackage[latin9]{inputenc}
\usepackage{color}
\usepackage{verbatim}
\usepackage{amsmath}
\usepackage{amsthm}
\usepackage{amssymb}
\usepackage{enumitem}
\usepackage{nicefrac}
\usepackage{pgfplots} 
\usepackage{subcaption}

\makeatletter
\theoremstyle{plain}
\newtheorem{thm}{\protect\theoremname}
\theoremstyle{plain}
\newtheorem*{rem}{\protect\remarkname}
\ifx\proof\undefined
\newenvironment{proof}[1][\protect\proofname]{\par
	\normalfont\topsep6\p@\@plus6\p@\relax
	\trivlist
	\itemindent\parindent
	\item[\hskip\labelsep\scshape #1]\ignorespaces
}{%
	\endtrivlist\@endpefalse
}
\providecommand{\proofname}{Proof}
\fi
\theoremstyle{plain}
\newtheorem{cor}[thm]{\protect\corollaryname}
\theoremstyle{plain}
\newtheorem{prop}[thm]{\protect\propositionname}

\theoremstyle{plain}
\newcounter{defin}
\newtheorem{definition}{Definition}[defin]

\theoremstyle{plain}
\newtheorem{assumption}{Assumption}[thm]

\newtheorem{lem}[thm]{Lemma}

 \PassOptionsToPackage{numbers}{natbib}

 \usepackage[preprint]{neurips_2020}

\usepackage[T1]{fontenc}    
\usepackage{hyperref}       
\usepackage{url}            
\usepackage{booktabs}       
\usepackage{amsfonts}       
\usepackage{nicefrac}       
\usepackage{microtype}      

\usepackage{wrapfig}

\title{Kernel Alignment Risk Estimator: \\ Risk Prediction from Training Data}

\author{%
  Arthur Jacot \\
  Ecole Polytechnique F\'ed\'erale de Lausanne \\
  \texttt{arthur.jacot@epfl.ch} \\
  \And
 Berfin \c{S}im\c{s}ek \\
  Ecole Polytechnique F\'ed\'erale de Lausanne \\
  \texttt{berfin.simsek@epfl.ch} \\
  \And
 Francesco Spadaro \\
  Ecole Polytechnique F\'ed\'erale de Lausanne \\
  \texttt{francesco.spadaro@epfl.ch} \\
  \And
 Cl\'ement Hongler \\
  Ecole Polytechnique F\'ed\'erale de Lausanne \\
  \texttt{clement.hongler@epfl.ch} \\
  \And
 Franck Gabriel \\
  Ecole Polytechnique F\'ed\'erale de Lausanne \\
  \texttt{franck.gabriel@epfl.ch} \\
}

\usepackage{amsmath}

\makeatother

\usepackage{babel}
\providecommand{\corollaryname}{Corollary}
\providecommand{\propositionname}{Proposition}
\providecommand{\remarkname}{Remark}
\providecommand{\theoremname}{Theorem}

\newcommand\krrf{\hat{f}^\epsilon_{\lambda}}
\newcommand\ye{y^\epsilon}
\newcommand\sctkl{\vartheta_{K,\lambda}}
\newcommand\sctl{\vartheta(\lambda)}
\newcommand\sct{\vartheta}
\newcommand\sctln{\vartheta(\lambda, N)}
\newcommand\psctl{\partial_\lambda\vartheta(\lambda)}
\newcommand\psct{\partial_\lambda\vartheta}
\newcommand\psctln{\partial_\lambda\vartheta(\lambda, N)}
\newcommand\sctklpi{\vartheta(\lambda, N, K, \pi)}
\newcommand\oon{\frac {1} {N} }
\newcommand\OO{\mathcal {O}}
\newcommand\OOT{\mathcal{O}^{T}}

\newcommand\kareylNG{\rho(\lambda, N, \ye, G)}

\newcommand\er{\hat R^\epsilon}
\newcommand\risk{R^\epsilon}

\newcommand\bra{\left\langle}
\newcommand\ket{\right\rangle}
\def\norm#1{\left\Vert{#1}\right\Vert}

\newcommand\tildeA{\tilde{A}_{\vartheta}}

\newcommand\Var{\mathrm{Var}}
\newcommand\Cov{\mathrm{Cov}}

\newcommand\EE{\mathbb E}
\newcommand\Tr{\mathrm{Tr}}

\begin{document}

\maketitle

\begin{abstract}
We study the risk (i.e. generalization error) of Kernel Ridge Regression (KRR) for a kernel $K$ with ridge $\lambda>0$ and i.i.d. observations. For this, we introduce two objects: the Signal Capture Threshold (SCT) and the Kernel Alignment Risk Estimator (KARE). The SCT $\sctkl$ is a function of the data distribution: it can be used to identify the components of the data that the KRR predictor captures, and to approximate the (expected) KRR risk. This then leads to a KRR risk approximation by the KARE $\rho_{K, \lambda}$, an explicit function of the training data, agnostic of the true data distribution.
We phrase the regression problem in a functional setting. The key results then follow from a finite-size analysis of the Stieltjes transform of general Wishart random matrices. Under a natural universality assumption (that the KRR moments depend asymptotically on the first two moments of the observations) we capture the mean and variance of the KRR predictor.
We numerically investigate our findings on the Higgs and MNIST datasets for various classical kernels: the KARE gives an excellent approximation of the risk, thus supporting our universality assumption.
Using the KARE, one can compare choices of Kernels and hyperparameters directly from the training set. The KARE thus provides a promising data-dependent procedure to select Kernels that generalize well.
\end{abstract}

\section{Introduction}
Kernel Ridge Regression (KRR) is a widely used statistical method to learn a function from its values on a training set \cite{Scholkopf,Shawe-Taylor}. It is a non-parametric generalization of linear regression to infinite-dimensional feature spaces. Given a positive-definite kernel function $K$ and (noisy) observations $\ye$ of a true function $ f^* $ at a list of points $ X = x_1, \ldots, x_N $, the $\lambda $-KRR estimator $ \krrf $ of $ f^* $ is defined by
\[
  \krrf(x)=\frac{1}{N}K(x,X) \left(\frac{1}{N}K(X,X)+\lambda I_N\right)^{-1} \ye,
\]
where $K(x,X)\!=\!(K(x,x_i))_{i=1,..,N} \in \mathbb{R}^N$ and $ K (X, X)\! =\! (K(x_i, x_j))_{i,j=1,.., N} \in \mathbb{R}^{N \times N}$.

Despite decades of intense mathematical progress, the rigorous analysis of the generalization of kernel methods remains a very active and challenging area of research. In recent years, many new kernels have been introduced for both regression and classification tasks; notably, a large number of kernels have been discovered in the context of deep learning, in particular through the so-called Scattering Transform \cite{mallat2012group}, and in close connection with deep neural networks \cite{cho2009, jacot2018neural}, yielding ever-improving performance for various practical tasks \cite{exact_arora2019, du2019, li2019, shankar2020}. Currently, theoretical tools to select the relevant kernel for a given task, i.e. to minimize the generalization error, are however lacking.

While a number of bounds for the risk of Linear Ridge Regression (LRR) or KRR \cite{caponnetto-07, sridharan-09, marteau-19} exist, most focus on the rate of convergence of the risk: these estimates typically involve constant factors which are difficult to control in practice. Recently, a number of more precise estimates have been given \cite{louart-17, dobriban2018, mei-19, liu-20, bordelon-2020}; however, these estimates typically require a priori knowledge of the data distribution. It remains a challenge to have estimates based on the training data alone, enabling one to make informed decisions on the choices of the ridge and of the kernel.

\subsection{Contributions}
We consider a generalization of the KRR predictor $ \krrf $: one tries to reconstruct a true function $f^*$ in a space of continuous functions $ \mathcal C $ from noisy observations $ \ye $ of the form $\left(o_1(f^*)+\epsilon e_1,\ldots,o_N(f^*)+\epsilon e_N\right)$, where the observations $o_i$ are i.i.d. linear forms $ \mathcal C \to \mathbb R $ sampled from a distribution $\pi$, $\epsilon$ is the level of noise, and the $e_1,\ldots,e_N$ are centered of unit variance. We work under the universality assumption that, for large $N$, only the first two moments of $\pi$ determine the behavior of the first two moments of $\krrf$. We obtain the following results:

\begin{enumerate}[leftmargin=*]
\item We introduce the Signal Capture Threshold (SCT) $ \sctklpi $,
which is determined by the ridge $\lambda$, the size of the training
set $N$, the kernel $K$, and the observations distribution $ \pi $ (more precisely, the dependence on $ \pi $ is only through its first two moments). We give approximations for the expectation and variance of the KRR predictor in terms of the SCT.

\item Decomposing $f^*$ along the kernel principal components of the data distribution, we observe that in expectation, the predictor $\krrf$ captures only the signal along the principal components with eigenvalues larger than the SCT. If $N$ increases or $\lambda$ decreases, the SCT $\vartheta$ shrinks, allowing the predictor to capture more signal. At the same time, the variance of $\krrf$ scales with the derivative $\partial_{\lambda}\vartheta$, which grows as $\lambda \to 0$, supporting the classical bias-variance tradeoff picture \cite{geman1992}.

\item We give an explicit approximation for the expected MSE risk $R^\epsilon (\krrf)$ and empirical MSE risk $\hat{R}^\epsilon (\krrf)$ for an arbitrary continuous true function $f^*$. We find that, surprisingly, the expected risk and expected empirical risk are approximately related by
\[
  \mathbb{E} [R^\epsilon (\krrf)] \approx \frac{\sctl^2}{\lambda^2} \mathbb{E} [\hat{R}^\epsilon (\krrf)].
\]

\item We introduce the Kernel Alignment Risk Estimator (KARE) as the ratio $ \rho $ defined by
\[
  \kareylNG=\frac{\frac{1}{N}\left(\ye\right)^{T}\left(\frac{1}{N}G+\lambda I_{N}\right)^{-2}\ye}{\left(\frac{1}{N}\mathrm{Tr}\left[\left(\frac{1}{N}G+\lambda I_{N}\right)^{-1}\right]\right)^{2}},
\]
where $G$ is the Gram matrix of $K$ on the observations. We show that the KARE approximates the expected risk; unlike the SCT, it is agnostic of the true data distribution. This result follows from the fact that $ \sctl \approx 1 / m_G(-\lambda) $, where $ m_G $ is the Stieltjes Transform of the Gram matrix $\frac{1}{N}G$.

\item Empirically, we find that the KARE predicts the risk on the Higgs and MNIST datasets. We see empirically that our results extend extremely well beyond the Gaussian observation setting, thus supporting our universality assumption (see Figure \ref{fig:KARE-predicts-risk}).
\end{enumerate}
Our proofs (see the Appendix) rely on a finite-size analysis of generalized Wishart matrices, in particular the complex Stieltjes transform $ m_G(z)$, evaluated at $ z = -\lambda $, and on fixed-point arguments.

\begin{figure}[!t]
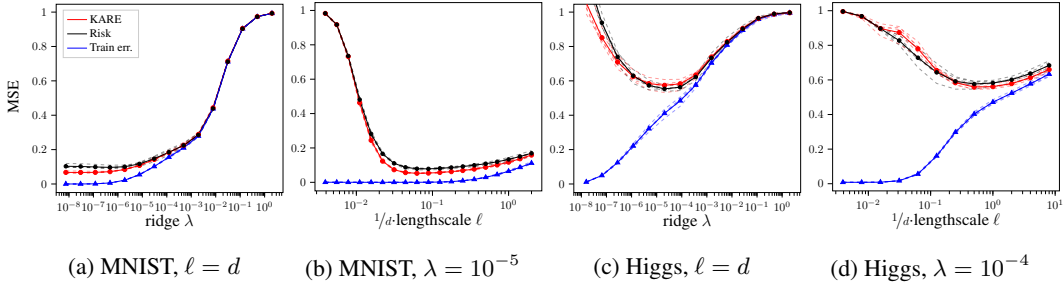

\vspace*{-0.3cm}
\rotatebox[origin=l]{90}{\tiny{\textcolor{white}{hey!!}MSE}}
\begin{subfigure}{.24\textwidth}
\input{MNIST-approx-lambd-N2000-RBF.tex}
\vspace*{-0.3cm}
\caption{{\small MNIST, $\ell = d$}}
\end{subfigure}
\begin{subfigure}{.24\textwidth}
\input{MNIST-approx-len-N2000-RBF.tex}
\vspace*{-0.3cm}
\caption{{\small MNIST, $\lambda = 10^{-5}$}}
\end{subfigure}
\begin{subfigure}{.24\textwidth}
\input{Higgs-approx-lambd-N1000-RBF.tex}
\vspace*{-0.3cm}
\caption{{\small Higgs, $\ell = d$}}
\end{subfigure}
\begin{subfigure}{.24\textwidth}
\begin{tikzpicture}[scale=0.44]

\begin{axis}[
legend cell align={left},
legend style={fill opacity=0.8, draw opacity=1, text opacity=1, draw=white!80!black},
log basis x={10},
tick align=outside,
tick pos=left,
x grid style={white!69.0196078431373!black},
xlabel={{\Large $\nicefrac{1}{d}\cdot$lengthscale $\ell$}},
xmin=0.00266804737647343, xmax=11.712685567565,
xmode=log,
xtick style={color=black},
y grid style={white!69.0196078431373!black},
ymin=-0.0414464953002449, ymax=1.04693679649874,
ytick style={color=black}
]
\addplot [semithick, black, dashed, opacity=0.4, forget plot]
table {%
0.00390625 0.994381227162768
0.0078125 0.968577068834081
0.015625 0.901874130155388
0.03125 0.82054124795334
0.0625 0.732778571463038
0.125 0.665207502340728
0.25 0.612030093551761
0.5 0.590192274392157
1 0.589464994644514
2 0.603161639029394
4 0.636248490377956
8 0.684668525724244
};
\addplot [semithick, blue, dashed, opacity=0.4, forget plot]
table {%
0.00390625 0.00823423340843671
0.0078125 0.00815227311870063
0.015625 0.00894800004707402
0.03125 0.0174312977136508
0.0625 0.0569415176361064
0.125 0.163057671321162
0.25 0.299443666456999
0.5 0.402178733032714
1 0.470795136728293
2 0.52516388379723
4 0.580254155402434
8 0.637994585354951
};
\addplot [semithick, red, dashed, opacity=0.4, forget plot]
table {%
0.00390625 0.994041806047238
0.0078125 0.967655259375288
0.015625 0.907840468058284
0.03125 0.890976616228274
0.0625 0.795560880130394
0.125 0.665654623847167
0.25 0.589150730353247
0.5 0.558924405335961
1 0.561208302012155
2 0.580678216034387
4 0.615990066668084
8 0.664455368255686
};
\addplot [semithick, black, dashed, opacity=0.4, forget plot]
table {%
0.00390625 0.997106603599078
0.0078125 0.969045513363782
0.015625 0.900079946764848
0.03125 0.843579165681347
0.0625 0.729465190320359
0.125 0.633698171829234
0.25 0.582061024225157
0.5 0.569621049226302
1 0.580533168817504
2 0.604509277463249
4 0.645632293599703
8 0.700051645154063
};
\addplot [semithick, blue, dashed, opacity=0.4, forget plot]
table {%
0.00390625 0.00824783836675048
0.0078125 0.00817740737121289
0.015625 0.00905030235423191
0.03125 0.0175931986867983
0.0625 0.0559405039421396
0.125 0.16002178178253
0.25 0.303027622811791
0.5 0.416012665202588
1 0.485302032475217
2 0.5362427211611
4 0.590374925371047
8 0.649858192355342
};
\addplot [semithick, red, dashed, opacity=0.4, forget plot]
table {%
0.00390625 0.997464828689692
0.0078125 0.975011935163114
0.015625 0.924762876689922
0.03125 0.90853118330713
0.0625 0.787771766831798
0.125 0.655711962180284
0.25 0.596039981159673
0.5 0.577627539491032
1 0.57799520817881
2 0.592498979980205
4 0.626535958892329
8 0.676752243276078
};
\addplot [semithick, black, dashed, opacity=0.4, forget plot]
table {%
0.00390625 0.994988647150328
0.0078125 0.965799758551122
0.015625 0.886911806189711
0.03125 0.80158310535927
0.0625 0.671950221404914
0.125 0.579719809666999
0.25 0.546600937386906
0.5 0.546503088835003
1 0.561404733374853
2 0.582519743645677
4 0.613998558800537
8 0.655199413069931
};
\addplot [semithick, blue, dashed, opacity=0.4, forget plot]
table {%
0.00390625 0.0082307187081884
0.0078125 0.0081100007459381
0.015625 0.00882741101318482
0.03125 0.0167490264261296
0.0625 0.054093262011211
0.125 0.156609813890588
0.25 0.290272363149271
0.5 0.395131904476477
1 0.465373872121515
2 0.521024651214784
4 0.577628898374286
8 0.637205353818899
};
\addplot [semithick, red, dashed, opacity=0.4, forget plot]
table {%
0.00390625 0.994674559025657
0.0078125 0.963795728844183
0.015625 0.897427276453792
0.03125 0.861624302425259
0.0625 0.760217067710734
0.125 0.639864560249143
0.25 0.570244678343305
0.5 0.548367912730951
1 0.554286131332906
2 0.57578600300578
4 0.613013448532338
8 0.663518976755539
};
\addplot [semithick, black, dashed, opacity=0.4, forget plot]
table {%
0.00390625 0.996866454233423
0.0078125 0.970126470730111
0.015625 0.891891346870936
0.03125 0.815885387268585
0.0625 0.720956606289312
0.125 0.644534338032873
0.25 0.599861927520169
0.5 0.583053065541199
1 0.582525209644837
2 0.595623385172997
4 0.626564188514909
8 0.669121321252371
};
\addplot [semithick, blue, dashed, opacity=0.4, forget plot]
table {%
0.00390625 0.00822186032023502
0.0078125 0.00802547250879969
0.015625 0.00834504879342117
0.03125 0.0155388496336264
0.0625 0.0518114749586712
0.125 0.155933123741463
0.25 0.290907985166847
0.5 0.392286431444086
1 0.460235117063856
2 0.512554948648565
4 0.565405025778857
8 0.61937408214849
};
\addplot [semithick, red, dashed, opacity=0.4, forget plot]
table {%
0.00390625 0.993759398542436
0.0078125 0.955185426197848
0.015625 0.853556328298997
0.03125 0.807680861189266
0.0625 0.733210878549337
0.125 0.63774698632899
0.25 0.571627459291476
0.5 0.544851479292881
1 0.548493889243451
2 0.56663807723219
4 0.600192132842085
8 0.645062915216225
};
\addplot [semithick, black, dashed, opacity=0.4, forget plot]
table {%
0.00390625 0.995825085052731
0.0078125 0.967844000179413
0.015625 0.905098997293434
0.03125 0.856594741156799
0.0625 0.785825823051806
0.125 0.695422256836818
0.25 0.618310346016451
0.5 0.592306519183395
1 0.597843914258836
2 0.620805963828327
4 0.660728098359091
8 0.710322204493474
};
\addplot [semithick, blue, dashed, opacity=0.4, forget plot]
table {%
0.00390625 0.00824089781769647
0.0078125 0.00816136835848743
0.015625 0.00894651127559009
0.03125 0.0181304803274305
0.0625 0.0609700081827185
0.125 0.168439919993277
0.25 0.304260186946338
0.5 0.408525506302938
1 0.473656392646148
2 0.522735158854107
4 0.573045698061171
8 0.623992187236944
};
\addplot [semithick, red, dashed, opacity=0.4, forget plot]
table {%
0.00390625 0.995696070967302
0.0078125 0.969217285970549
0.015625 0.90069400713358
0.03125 0.902288729992323
0.0625 0.822710966042849
0.125 0.673191765636152
0.25 0.59325631413725
0.5 0.565697922579539
1 0.563616379940341
2 0.577364883925024
4 0.608024074116222
8 0.649724093953302
};
\addplot [thick, blue]
table {%
0.00390625 0.00823510972426141
0.0078125 0.00812530442062775
0.015625 0.0088234546967004
0.03125 0.0170885705575271
0.0625 0.0559513533461693
0.125 0.160812462145804
0.25 0.297582364906249
0.5 0.402827048091761
1 0.471072510207006
2 0.523544272735157
4 0.577341740597559
8 0.633684880182925
};
\addlegendentry{trn}
\addplot [thick, red]
table {%
0.00390625 0.995127332654465
0.0078125 0.966173127110196
0.015625 0.896856191326915
0.03125 0.87422033862845
0.0625 0.779894311853022
0.125 0.654433979648347
0.25 0.58406383265699
0.5 0.559093851886073
1 0.561119982141533
2 0.578593232035517
4 0.612751136210212
8 0.659902719491366
};
\addlegendentry{estim}
\addplot [thick, black]
table {%
0.00390625 0.995833603439666
0.0078125 0.968278562331702
0.015625 0.897171245454863
0.03125 0.827636729483868
0.0625 0.728195282505886
0.125 0.64371641574133
0.25 0.591772865740089
0.5 0.576335199435611
1 0.582354404148109
2 0.601324001827929
4 0.636634325930439
8 0.683872621938817
};
\addlegendentry{risk}
\addplot [thick, blue, mark=triangle*, mark size=2.2, mark options={solid}, only marks, forget plot]
table {%
0.00390625 0.00823510972426141
0.0078125 0.00812530442062775
0.015625 0.0088234546967004
0.03125 0.0170885705575271
0.0625 0.0559513533461693
0.125 0.160812462145804
0.25 0.297582364906249
0.5 0.402827048091761
1 0.471072510207006
2 0.523544272735157
4 0.577341740597559
8 0.633684880182925
};
\addplot [thick, red, mark=*, mark size=1.8, mark options={solid}, only marks, forget plot]
table {%
0.00390625 0.995127332654465
0.0078125 0.966173127110196
0.015625 0.896856191326915
0.03125 0.87422033862845
0.0625 0.779894311853022
0.125 0.654433979648347
0.25 0.58406383265699
0.5 0.559093851886073
1 0.561119982141533
2 0.578593232035517
4 0.612751136210212
8 0.659902719491366
};
\addplot [thick, black, mark=*, mark size=1.5, mark options={solid}, only marks, forget plot]
table {%
0.00390625 0.995833603439666
0.0078125 0.968278562331702
0.015625 0.897171245454863
0.03125 0.827636729483868
0.0625 0.728195282505886
0.125 0.64371641574133
0.25 0.591772865740089
0.5 0.576335199435611
1 0.582354404148109
2 0.601324001827929
4 0.636634325930439
8 0.683872621938817
};
\legend{};
\end{axis};

\end{tikzpicture}
\vspace*{-0.3cm}
\caption{{\small Higgs, $\lambda=10^{-4}$}}
\end{subfigure}
\caption{Comparison between the KRR risk and the KARE for various choices of normalized lengthscale $\nicefrac{\ell}{d}$ and ridge $\lambda$ on the MNIST dataset (restricted to the digits $ 7 $ and $ 9 $, labeled by $ 1 $ and $ -1 $ respectively, $N=2000$) and on the Higgs dataset (classes `b' and `s', labeled by $-1$ and $1$, $N=1000$) with the RBF Kernel $K(x,x') = \exp(\nicefrac{-\|x-x'\|_2^2}{\ell})$.
KRR predictor risks, and KARE curves (shown as dashed lines, $5$ samples) concentrate around their respective averages (solid lines).}
\label{fig:KARE-predicts-risk}
\end{figure}


\subsection{Related Works}
The theoretical analysis of the risk of KRR has seen tremendous developments in the recent years. In particular, a number of upper and lower bounds for kernel risk have been obtained \cite{caponnetto-07, sridharan-09, marteau-19} in various settings: notably, convergence rates (i.e. without control of the constant factors) are obtained in general settings. This allows one to abstract away a number of details about the kernels (e.g. the lengthscale), which don't influence the asymptotic rates. However, this does not give access to the risk at finite data size (crucial to pick e.g. the correct lengthscale or the NTK depth \cite{jacot2018neural}).

A number of recent results have given precise descriptions of the risk for ridge regression \cite{dobriban2018,liu-20}, for random features \cite{mei-19,jacot-2020}, and in relation to neural networks \cite{louart-17, bordelon-2020}. These results rely on the analysis of the asymptotic spectrum of general Wishart random matrices, in particular through the Stieltjes transform \cite{silverstein-1995,bai-2008}. The limiting Stieltjes transform can be recovered from the formula for
the product of freely independent matrices \cite{gabriel-15}. To extend these asymptotic results to finite-size settings, we generalize and adapt the results of \cite{jacot-2020}.

While these techniques have given simple formulae for the KRR predictor expectation, approximating its variance has remained more challenging. For this reason the description of the expected risk in \cite{louart-17} is stated as a conjecture. In \cite{liu-20} only the bias component of the risk is approximated. In \cite{dobriban2018} the expected risk is given only for random true functions (in a Bayesian setting) with a specific covariance. In \cite{bordelon-2020}, the expected risk follows from a heuristic spectral analysis combining a PDE approximation and replica tricks. In this paper, we approximate the variance of the predictor along the principal components, giving an approximation of the risk for any continuous true function.

The SCT is related to a number of objects from previous works, such as the effective dimension of \cite{zhang-03, caponnetto-07}, the companion Stieltjes transform of \cite{dobriban2018,liu-20}, and particularly the effective ridge of \cite{jacot-2020}. The SCT can actually be viewed as a direct translation to the KRR risk setting of \cite{jacot-2020}.

\subsection{Outline}\label{subsec:outline}

In Section \ref{sec:setup}, we first introduce the Kernel Ridge Regression (KRR) predictor in functional space (Section \ref{subsec:krr-predictor}) and formulate its train error and risk for random observations (Section \ref{subsec:train-err-risk}).

The rest of the paper is then devoted to obtaining approximations for the KRR risk. In Section \ref{sec:SCT},the Signal Capture Threshold (SCT) is introduced and used to study the mean and variance of the KRR predictor (Sections \ref{subsec:mean-predictor} and \ref{subsec:variance-predictor}). An approximation of the SCT in terms of the observed data is then given (Section \ref{subsec:approximations}).
In Section \ref{sec:risk-prediction-with-kare}, the expected risk and the expected empirical risk are approximated in terms of the SCT and its derivative w.r.t. the ridge $ \lambda $. The SCT approximation of Section \ref{subsec:approximations}, together with the estimates of Section \ref{subsec:expected-risk}, leads to an approximation of the KRR risk by the Kernel Alignment Risk Estimator (KARE).


\section{Setup}\label{sec:setup}

Given a compact $\Omega \subset \mathbb R^d$, let $\mathcal{C}$ denote the space of continuous $f:\Omega\to\mathbb{R}$, endowed with the supremum norm $\left\Vert f\right\Vert _{\infty}=\sup_{x \in \Omega}\left|f(x)\right|$.
In the classical regression setting, we want to reconstruct a true function $f^{*} \in \mathcal C$ from its values on a training set $x_{1},\dots,x_{N}$, i.e. from the noisy labels $y^{\epsilon}=\left(f^{*}(x_{1})+\epsilon e_{1},\dots,f^{*}(x_{N})+\epsilon e_{N}\right)^{T}$ for some i.i.d. centered noise $e_{1},\dots,e_{N}$ of unit variance and noise level $\epsilon \geq 0$.

In this paper, the observed values (without noise) of the true function $ f^* $ consist in observations
$o_{1},\ldots,o_{N}\in\mathcal{C}^{*}$, where $\mathcal{C}^*$ is the dual space, i.e. the space of bounded linear functionals $\mathcal{C} \to \mathbb R$. We thus represent the training set of $N$ observations $o_{1},\dots,o_{N}$ by the \emph{sampling operator} $\mathcal{O}:\mathcal{C}\to\mathbb{R}^{N}$ which maps a function $f\in\mathcal{C}$ to the vector of observations $\mathcal{O}(f)=(o_{1}(f),\dots,o_{N}(f))^{T}$.

The classical setting corresponds to the case where the observations are evaluations of $ f^* $ at points $ x_1, \ldots, x_N \in \Omega $, i.e. $ o_i \left( f^* \right) = f^* (x_i) $ for $ i=1,\ldots,N $. In time series analysis (when $\Omega\subset\mathbb{R}$), the observations can be the averages $o_{i}(f^{*})=\frac{1}{b_{i}-a_{i}}\int_{a_{i}}^{b_{i}}f^{*}(t)dt$ over time intervals $[a_{i},b_{i}]\subset\mathbb{R}$.

\subsection{Kernel Ridge Regression Predictor}\label{subsec:krr-predictor}
The regression problem is now stated as follows: given noisy observations $\ye_i = o_{i}\left(f^{*}\right)+\epsilon e_{i}$ with i.i.d. centered noises $e_{1},\ldots,e_{N}$ of unit variance, how can one reconstruct $f^{*}$?

\begin{definition}
Consider a continuous positive kernel $K:\Omega\times\Omega\to\mathbb{R}$ and a ridge parameter $\lambda>0$. The Kernel Ridge Regression (KRR) predictor with ridge $\lambda$ is the function $\krrf : \Omega \to \mathbb{R}$
\[
  \krrf =\oon K\OOT (\oon \OO K\OOT+\lambda I_{N} )^{-1}\ye
\]
where $\OOT:\mathbb{R}^{N}\to\mathcal{C}^{*}$ is the adjoint of $ \OO $ defined by $(\OOT y)(f)=y^{T}\OO(f)$ and where we view $K$ as a map $ \mathcal{C^{*}}\to\mathcal{C}$ with $(K\mu)(x)=\mu(K(x,\cdot))$.
\end{definition}

We call the $N\times N$ matrix $G=\OO K \OOT$ the \emph{Gram matrix}: in the classical setting, when the observations are $o_{i}=\delta_{x_{i}}$ (with $\delta_{x} (f) = f(x) $), $ G$ is the usual Gram matrix, i.e. $G_{ij}=K(x_{i},x_{j})$.

\subsection{Training Error and Risk}\label{subsec:train-err-risk}

We consider the least-squares error (MSE loss) of the KRR predictor, taking into account randomness of: (1) the test point, random observation $ o $ to which is added a noise $ \epsilon e $ (2) the training data, made of $N$ observations $ o_i $ plus noises $ \epsilon e_i \sim \nu $, where $ o, o_1, \ldots, o_n \sim \pi $ and $ e, e_1,\ldots, e_N$ are i.i.d. The expected risk of the KRR predictor is thus taken w.r.t. the test and training observations and their noises. Unless otherwise specified, the expectations are taken w.r.t. all these sources of randomness.

For (fixed) observations $ o_1, \ldots, o_N $, the \emph{empirical risk} or \emph{training error} of the KRR predictor $ \krrf $ is
\[
    \er (\krrf) = \oon \sum_{i=1}^N (o_i(\krrf) - \ye_i)^2 = \oon \norm{\OO (\krrf) - \ye }^2.
\]

For a random observation $ o $ sampled from $ \pi $ and a noise $ \epsilon e $ (where $ e \sim \nu $ is centered of unit variance as before), the \emph{risk} $ \risk (\krrf) $ of the KRR predictor $ \krrf $ is defined by
\[
  \risk (\krrf) = \EE_{o\sim\pi,e\sim\nu}\left[ (o(f^{*}) + \epsilon e - o(\krrf))^{2} \right].
\]
Describing the observation variance by the bilinear form $\bra f,g \ket _{S}=\EE_{o \sim \pi}\left[o(f)o(g)\right]$ and the related semi-norm $\Vert f\Vert_{S}= \bra f,f \ket _{S}^{\nicefrac{1}{2}}$, the risk can be rewritten as
$ \risk(\krrf)=\| \krrf - f^*\|_{S}^{2}+\epsilon^2 $.

From now on, we will assume that $\bra\cdot, \cdot \ket _{S}$ is a scalar product; note that in the classical setting, when $o$ is the evaluation of $f^*$ at a point $x \in \Omega$ with $x \sim \sigma$, the $S$-norm is given by $\|f\|_{S}^{2}=\int_{\Omega} f(x)^{2} \sigma(dx)$.

The following three operators $ \mathcal C \to \mathcal C $ are central to our analysis:
\begin{definition} The KRR reconstruction operator $ A_\lambda: \mathcal C \to \mathcal C $, the KRR Integral Operator $ T_K: \mathcal C \to \mathcal C $, and its empirical version $ T^N_K : \mathcal C \to \mathcal C$ are defined by
	\begin{align*}
		\ A_\lambda &= \oon K\OOT (\oon \OO K\OOT + \lambda I_{N} )^{-1} \OO, \\
		\ (T_Kf)(x) &= \bra f, K(x,\cdot) \ket_S = \EE_{o\sim\pi}\left[ o(f)o(K(x,\cdot)) \right], \\
		\ (T_K^N f)(x) &= \oon K \OOT \OO f(x)= \oon \sum_{i=1}^N o_i(f)o_i(K(x,\cdot)).
	\end{align*}
\end{definition}
Note that in the noiseless regime (i.e. when $ \epsilon = 0 $), we have $ \krrf \big|_{\epsilon=0} = A_\lambda f^* $. Also note that $A_\lambda$ and $T^N_K $ are random operators, as they depend on the random observations. The operator $ T_K $ is the natural generalization to our framework of the integration operator $ f \mapsto \int K(x, \cdot) f(x) \sigma(d x) $, which is defined with random observations $ \delta_x $ with $ x \sim \sigma $ in the classical setting.

The reconstruction and empirical integral operators are linked by $  A_\lambda=T^N_K (T^N_K+\lambda I_\mathcal{C})^{-1}$,
which follows from the identity $\left(\oon\OO K \OOT+\lambda I_{N}\right)^{-1}\OO = \OO \left(\oon K \OOT \OO+\lambda I_{\mathcal{C}}\right)^{-1}$. As $N\to\infty$, we have that $ T^N_K \to T_K $, and it follows that
 \begin{align}\label{eqn:two-moments-tildeA}
	 \ A_\lambda \to \tilde A_\lambda:= T_K (T_K+\lambda I_\mathcal{C})^{-1}.
 \end{align}

\subsection{Eigendecomposition of the Kernel}\label{subsec:eigendecomposition}

We will assume that the kernel $K$ can be diagonalized by a countable family of eigenfunctions $(f^{(k)})_{k \in \mathbb{N}}$ in $\mathcal C$ with eigenvalues $(d_k)_{k \in \mathbb{N}}$, orthonormal with respect to the scalar product $ \bra\cdot,\cdot \ket_S $, such that we have (with uniform convergence):
\[
  K(x,x')=\sum_{k=1}^{\infty}d_{k}f^{(k)}(x)f^{(k)}(x').
\]
The functions $f^{(k)}$ are also eigenfunctions of $ T_K $: we have $ T_K f^{(k)} = d_k f^{(k)} $. We will also assume that $\Tr\left[T_K\right]=\sum_{k=1}^{\infty}\bra f^{(k)},T_{K}(f^{(k)})\ket _{S}=\sum_{k=1}^{\infty}d_{k}$ is finite.
Note that in the classical setting $K$ can be diagonalized as above (by Mercer's theorem), and $\Tr\left[T_K\right]=\EE_{x\sim\sigma}\left[K(x,x)\right]$ is finite.

\subsection{Gaussianity Assumption}
As seen in Equation (\ref{eqn:two-moments-tildeA}) above, $\tilde{A}_\lambda$ only depends on the first two moments of $\pi$ (through $\bra\cdot,\cdot\ket_S$), suggesting the following assumption, with which we will work in this paper:
\begin{assumption} \label{assumption:gaussianity} As far as one is concerned with the first two moments of the $A_\lambda$ operator, for large but finite $N$, one can assume that the observations $o_1,\ldots,o_N$ are Gaussian, i.e. that for any tuple of functions $(f_1,\ldots,f_N)$, the vector $\left(o_1(f_1),\ldots,o_N(f_N)\right)$ is a Gaussian vector.
\end{assumption}
Though our proofs use this assumption, the ideas in \cite{louart-17, benigni-2019} suggest a path to extend them beyond the Gaussian case, where our numerical experiments (see Figure \ref{fig:KARE-predicts-risk}) suggest that our results remain true.

\section{Predictor Moments and Signal Capture Threshold}\label{sec:SCT}
A central tool in our analysis of the KRR predictor $ \krrf $ is the Signal Capture Threshold (SCT):
\begin{definition}
For $ \lambda > 0 $, the \emph{Signal Capture Threshold} $\sctl= \sctklpi $ is the unique positive solution (see Section \ref{subsec:Appendix-Concentration-Stieltjes} in the Appendix) to the equation:
\[
  \sctl = \lambda+\frac{\sctl}{N}\mathrm{Tr}\left[T_{K}\left(T_{K}+ \sctl I_{\mathcal{C}}\right)^{-1}\right].
\]
\end{definition}
In this section, we use $ \sctl $ and the derivative $ \partial_\lambda \sctl $ for the estimation of the mean and variance of the KRR predictor $ \krrf $ upon which the Kernel Alignment Risk Estimator of Section \ref{sec:risk-prediction-with-kare} is based.

\subsection{Mean predictor}\label{subsec:mean-predictor}
The expected KRR predictor can be expressed in terms of the expected reconstruction operator $A_\lambda$
\[
  \EE[\krrf]=\EE[\oon K\OOT(\oon \OO K\OOT+\lambda I_{N})^{-1}\ye]=\EE\left[A_\lambda\right] f^*,
\]
where we used the fact that $\EE_{e_1,\ldots,e_N}[\ye]=\OO f^*$.

\begin{thm}[Theorem \ref{thm:approx_expectation_A} in the Appendix]
  \label{thm:expectation_A}
  The expected reconstruction operator $\EE[A_\lambda]$ is approximated by the operator $\tildeA=T_{K}\left(T_{K}+\sctl I_{\mathcal{C}}\right)^{-1}$ in the sense that for all $f,g\in\mathcal{C}$,
  \[
  \left|\bra f,\left(\EE\left[A_{\lambda}\right]-\tildeA\right)g\ket_{S}\right|\leq  \left(\frac{1}{N} + \boldsymbol{P}_0 (\frac{\Tr[T_K]}{\lambda N}) \right)  \left|\bra f,\tildeA(I_{\mathcal{C}}-\tildeA)g\ket_{S}  \right|,
  \]
  for a polynomial $\boldsymbol{P}_0$ with nonnegative coefficients and $ \boldsymbol{P}_0 (0) = 0 $.
\end{thm}

This theorem gives the following motivation for the name SCT: if the true function $ f^* $ is an eigenfunction of $ T_K $, i.e. $ T_K f^* = \delta f^* $, then $\tildeA f^* = \frac{\delta}{\sctl+\delta} f^*$ and we get:
\begin{itemize}[leftmargin=*]
\item if $\delta\gg\sctl$, then $\frac{\delta}{\sctl+\delta}\approx1$
and  $\EE\left[A_{\lambda}\right]f^{*}\approx f^{*}$,
i.e. the function is learned on average,
\item if $\delta\ll\sctl$, then $\frac{\delta}{\sctl+\delta}\approx0$
and $\EE\left[A_{\lambda}\right]f^{*}\approx0$,
i.e. the function is not learned on average.
\end{itemize}
More generally, if we decompose a true function $f^*$ along the principal components (i.e. eigenfunctions) of $ T_K $, the signal along the $k$-th principal component $f^{(k)}$ is captured whenever the corresponding eigenvalue $d_k\gg\sctl$ and lost when $d_k\ll\sctl$.

\subsection{Variance of the predictor}\label{subsec:variance-predictor}
We now estimate the variance of $ \krrf $ along each principal component in terms of the SCT $\sctl$ and its derivative $\psctl$. Along the eigenfunction $f^{(k)}$, the variance is estimated by $ V_k $, where
\[
V_k(f^*, \lambda, N, \epsilon) = \frac{\psctl}{N}\left(\left\Vert (I_{\mathcal{C}}-\tilde{A}_{\sct})f^{*}\right\Vert_{S}^2+\epsilon ^2 + \bra f^{(k)},f^*\ket_S^2\frac{\vartheta^2(\lambda)}{(\sctl+d_k)^2}\right)\frac{d_k^2}{(\sctl+d_k)^2}.
\]
\begin{thm}[Theorem \ref{thm:variance_A} in the Appendix]
\label{thm:variance_A} There is a constant $\boldsymbol{C}_1>0$ and a polynomial $\boldsymbol{P}_1$ with nonnegative coefficients and with $ \boldsymbol{P}_1 (0) = 0 $ such that
\[
\left|\mathrm{Var}\left(\bra f^{(k)},\krrf \ket _{S}\right)-V_k\right| \leq  \left(\frac{\boldsymbol{C}_1}{N} + \boldsymbol{P}_1 (\frac{\Tr[T_K]}{\lambda N^{\frac 1 2}} ) \right) V_k.
\]
\end{thm}

As shown in Section \ref{subsec:expected-risk}, understanding the variance along the principal components (rather than the covariances between the principal components) is enough to describe the risk.

\begin{figure}[!t]
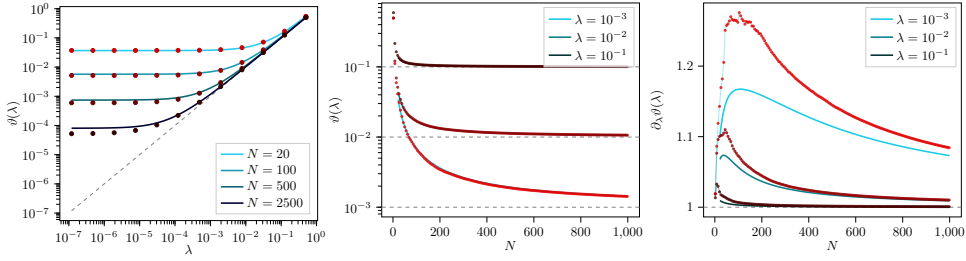

  \begin{subfigure}{.3\textwidth}
    \input{SCT-d20.tex}
  \end{subfigure}
  \begin{subfigure}{.3\textwidth}
    \input{SCT-d20-N-f.tex}
  \end{subfigure}
  \begin{subfigure}{.3\textwidth}
    \input{SCT-d20-der-N-f.tex}
  \end{subfigure}
  \caption{\textit{Signal Capture Threshold and Derivative.} We consider the RBF Kernel on the standard $d$-dimensional Gaussian with $\ell=d=20$. In blue lines, exact formulas for the SCT $\sctl$ and $\psctl$, computed using the eigenvalue decomposition of the integral operator $T_K$; in red dots, their approximation with Proposition \ref{prop:approx_stieltjes}.}
	\label{fig:behavior-SCT}
\end{figure}

\subsection{Behavior of the SCT}\label{subsec:bounds}
The behavior of the SCT can be controlled by the following (agnostic of the exact spectrum of $T_K$)
\begin{prop}[Proposition \ref{prop:appendix-bound-sct-universal} in the Appendix]\label{prop:bound-sct-universal} For any $\lambda > 0$, we have
\begin{align*}
\lambda < \sctln \leq \lambda + \oon \Tr[T_K], \quad \quad 1 \leq \psctln \leq \frac{1}{\lambda} \sctln,
\end{align*}
moreover $\sctln$ is decreasing as a function of $N$.
\end{prop}

\begin{rem}
As $N\to\infty$, we have $\sctln$ decreases down to $\lambda$ (see also Figure \ref{fig:behavior-SCT}), in agreement with the fact that $A_\lambda  \to \tilde{A}_\lambda $.
\end{rem}

As $\lambda \to 0$, the above upper bound for $ \psct $ becomes useless. Still, assuming that the spectrum of $K$ has a sufficiently fast power-law decay, we get:
\begin{prop}[Proposition  \ref{prop:appendix-bound-sct-polynomial} in the Appendix]\label{prop:bound-sct-polynomial-decay} If $d_k=\Theta(k^{-\beta})$ for some $\beta>1$, there exist $ c_0, c_1, c_2 > 0 $ such that for any $\lambda>0$
  \begin{align*}
    \lambda + c_0 N^{-\beta} \leq \sctln \leq\ c_2 \lambda + c_1 N^{-\beta}, \quad \quad 1 \leq \psctln \leq c_2.
  \end{align*}
\end{prop}

\subsection{Approximation of the SCT from the training data}\label{subsec:approximations}

The SCT $ \sct $ and its derivative $ \psct $ are functions of $\lambda, N$, and of the spectrum of $T_{K}$. In practice, the spectrum of $T_K$ is not known: for example, in the classical setting, one does not know the true data distribution $\sigma$. Fortunately, $ \sct $ can be approximated by $ 1 / m_G(-\lambda) $, where $ m_G $ is the \emph{Stieltjes Transform} of the Gram matrix, defined by $ m_G(z) = \Tr \left[ (\oon G - z I_N)^{-1} \right]$. Namely, we get:
\begin{prop}[Proposition \ref{prop:loose_concentration_Stieltjes_moments} in the Appendix]
\label{prop:approx_stieltjes} For any $\lambda > 0, s\in\mathbb N$, there is a $\boldsymbol{c}_s>0$ such that
\[ \EE\left[\left|\nicefrac{1}{\sctl} - m_G(-\lambda)\right|^{2s}\right] \leq \frac{\boldsymbol{c}_s (\Tr[T_K])^{2s}}{\lambda^{4s}N^{3s}}.
\]
\end{prop}
\begin{rem}
Likewise, we have $\psct \approx \left( \partial_z m_G(z) /  m_G(z)^2 \right)|_{z=-\lambda} $, as shown in the Appendix.
\end{rem}

\section{Risk Prediction with KARE}\label{sec:risk-prediction-with-kare}
In this section, we show that the Expected Risk $ \EE [\risk ( \krrf )] $ can be approximated in terms of the training data by the Kernel Alignment Risk Estimator (KARE).
\begin{definition}
  The Kernel Alignment Risk Estimator (KARE) $\rho$ is defined by
  \[
    \kareylNG=\frac{\oon\left(\ye\right)^{T}\left(\oon G+\lambda I_{N}\right)^{-2}\ye}{\left(\oon\Tr\left[\left(\oon G+\lambda I_{N}\right)^{-1}\right]\right)^{2}}.
  \]
\end{definition}
In the following, using Theorems \ref{thm:expectation_A} and \ref{thm:variance_A}, we give an approximation for the expected risk and expected empirical risk in terms of the SCT and the true function $f^*$. This yields the important relation (\ref {eq:kare-relation}) in Section \ref{subsec:kare}, which shows that the KARE can be used to efficiently approximate the kernel risk.

\subsection{Expected Risk and Expected Empirical Risk}\label{subsec:expected-risk}\label{subsec:empirical-risk}
The expected risk is approximated, in terms of the SCT and the true function $f^*$, by
\[
  \tilde R^\epsilon (f^*, \lambda, N, K, \pi) =  \psctl ( \| (I_{\mathcal{C}}-\tildeA )f^{*} \|_{S}^{2}+\epsilon^{2}),
\]
as shown by the following:
\begin{thm}[Theorem \ref{thm:appendix-expected-risk} in the Appendix]
\label{thm:expected-risk}There exists a constant $\boldsymbol{C}_2 > 0$ and a polynomial $\boldsymbol{P}_2$ with nonnegative coefficients and with $ \boldsymbol{P}_2 (0) = 0 $, such that we have
\begin{align*}
\left| \mathbb{E}[\risk(\krrf)] - \tilde R^\epsilon (f^*, \lambda, N, K, \pi) \right|  &\leq   \left(\frac{\boldsymbol{C}_2}{N} + \boldsymbol{P}_2 (\frac{\Tr[T_K]}{\lambda N^{\frac 1 2}} ) \right) \tilde R^\epsilon (f^*, \lambda, N, K, \pi).
\end{align*}
\end{thm}
\begin{proof} (Sketch; the full proof is given in the Appendix). From the bias-variance decomposition:
\[
  \mathbb{E}[R^{\epsilon}(\hat{f}_{\lambda}^{\epsilon})]  =
  R^{\epsilon}(\EE[\hat{f}_{\lambda}^{\epsilon}])+\sum_{k=1}^\infty \mathrm{Var}( \langle f^{(k)}, \krrf \rangle_S).
\]
By Theorem \ref{thm:expectation_A}, and a small calculation, the bias is approximately $  \| (I_{\mathcal{C}}-\tildeA )f^{*} \|_{S}^{2}+\epsilon^{2}.$
By Theorem \ref{thm:variance_A}, and a calculation, the variance is approximately $(\psctl-1) ( \| (I_{\mathcal{C}}-\tildeA )f^{*} \|_{S}^{2}+\epsilon^{2}).$
\end{proof}
The approximate expected risk $\tilde R^\epsilon (f^*, \lambda, N, K, \pi)$ is increasing in both $\sct$ and $\psct$. As $\lambda$ increases, the bias increases with $\sct$, while the variance decreases with $\psct$: this leads to the bias-variance tradeoff. On the other hand, as a function of $N$, $\sct$ is decreasing but $\psct$ is generally not monotone: this can lead to so-called multiple descent curves in the risk as a function of $N$ \cite{liang-20}.

\begin{rem}
For a decaying ridge $\lambda=cN^{-\gamma}$ for $0<\gamma<\frac 1 2$, as $N\to\infty$, by Proposition \ref{prop:bound-sct-universal}, we get  $\sctl \to 0$ and $\psctl \to 1$: this implies that $\mathbb{E}[R^{\epsilon}(\hat{f}_{\lambda}^{\epsilon})] \to \epsilon^2$. Hence the KRR can learn any continuous function $f^*$ as $N\to\infty$ (even if $f^*$ is not in the RKHS associated with $K$).
\end{rem}

\begin{rem}
In a Bayesian setting, assuming that $f^{*}$ is random with zero mean and covariance kernel $\Sigma$, the optimal choices for the KRR predictor are $K=\Sigma$ and $\lambda=\nicefrac{\epsilon^{2}}{N}$ (see Section \ref{subsec:Bayesian-setting} in the Appendix).
When $K=\Sigma$ and $\lambda= \nicefrac{\epsilon^{2}}{N}$, the formula of Theorem 6 simplifies (see Corollary \label{cor:Bayesian-setting} in the Appendix) to
\[
    \mathbb{E}\left[R^{\epsilon}\left(\krrf\right)\right] \approx N\vartheta\left(\frac{\epsilon^{2}}{N},\Sigma\right).
\]
\end{rem}

The empirical risk (or train error) $\hat{R}^{\epsilon}(\krrf)=\lambda^{2}(\ye)^{T}(\oon G+\lambda I_{N})^{-2}\ye$
can be analyzed with the same theoretical tools. Its approximation in terms of the SCT is given as follows:
\begin{thm}[Theorem \ref{thm:appendix-expected-empirical-risk} in the Appendix]\label{thm:train-error}
There exists a constant $\boldsymbol{C}_3 > 0$ and a polynomial $\boldsymbol{P}_3$ with nonnegative coefficients and with $ \boldsymbol{P}_3 (0) = 0 $ such that we have
\[
  \left| \EE [\hat{R}^{\epsilon} (\krrf)] - \frac{\lambda^2}{\sctl^2} \tilde{R}^{\epsilon}(\krrf, \lambda, N, K, \pi) \right|\leq  \left(\frac{1}{N} + \boldsymbol{P}_3  (\frac{\Tr[T_K]}{\lambda N} ) \right) \tilde R^\epsilon (f^*, \lambda, N, K, \pi).
\]
\end{thm}

\subsection{KARE: Kernel Alignment Risk Estimator}\label{subsec:kare}

While the above approximations (Theorems \ref{thm:expected-risk} and \ref{thm:train-error}) for the expected risk and empirical risk depend on $f^*$, their combination yields the following relation, which is surprisingly independent of $ f^* $:
\begin{equation} \label{eq:kare-relation}
\EE\left[R^{\epsilon}\left(\krrf\right)\right]\approx\frac{\vartheta^{2}}{\lambda^{2}}\EE\left[\hat{R}^{\epsilon}\left(\krrf\right)\right].
\end{equation}
Since $ \sct $ can be approximated from the training set (see Proposition \ref{prop:approx_stieltjes}), so can the expected risk. Assuming that the risk and empirical risk concentrate around their expectations, we get the KARE:
\[
R^{\epsilon}\left(\krrf\right) \approx \kareylNG = \frac{\oon\left(\ye\right)^{T}\left(\oon G+\lambda I_{N}\right)^{-2}\ye}{\left(\oon\Tr\left[\left(\oon G+\lambda I_{N}\right)^{-1}\right]\right)^{2}}.
\]

\begin{rem}
    As shown in the Appendix, estimating the risk of the expected predictor $ \EE[\krrf] $  yields:
		\[ \risk ( \EE[\krrf] ) \approx \varrho(\lambda, N, \ye, G) = \frac{(\ye)^T (\oon G + \lambda I_N )^{-2} \ye} {\Tr [ (\oon G + \lambda I_N)^{-2} ]}. \]
		Note that both $\rho$ and $\varrho$ are invariant (as is the risk) under the simultaneous rescaling $K, \lambda \leadsto \alpha K, \alpha \lambda$.
\end{rem}

\begin{figure}[!t]
\vspace*{-0.3cm}
\rotatebox[origin=l]{90}{\tiny{\textcolor{white}{he}ridge $\lambda$}}
\begin{subfigure}{.24\textwidth}
\begin{tikzpicture}[scale=0.36]

\begin{axis}[
colorbar,
colorbar style={ytick={0.1,0.2,0.3,0.4,0.5,0.6,0.7,0.8,0.9,1},
yticklabels={0.1,0.2,0.3,0.4,0.5,0.6,0.7,0.8,0.9,1.0},ylabel={}},
colormap={mymap}{[1pt]
  rgb(0pt)=(0.2298057,0.298717966,0.753683153);
  rgb(1pt)=(0.26623388,0.353094838,0.801466763);
  rgb(2pt)=(0.30386891,0.406535296,0.84495867);
  rgb(3pt)=(0.342804478,0.458757618,0.883725899);
  rgb(4pt)=(0.38301334,0.50941904,0.917387822);
  rgb(5pt)=(0.424369608,0.558148092,0.945619588);
  rgb(6pt)=(0.46666708,0.604562568,0.968154911);
  rgb(7pt)=(0.509635204,0.648280772,0.98478814);
  rgb(8pt)=(0.552953156,0.688929332,0.995375608);
  rgb(9pt)=(0.596262162,0.726149107,0.999836203);
  rgb(10pt)=(0.639176211,0.759599947,0.998151185);
  rgb(11pt)=(0.681291281,0.788964712,0.990363227);
  rgb(12pt)=(0.722193294,0.813952739,0.976574709);
  rgb(13pt)=(0.761464949,0.834302879,0.956945269);
  rgb(14pt)=(0.798691636,0.849786142,0.931688648);
  rgb(15pt)=(0.833466556,0.860207984,0.901068838);
  rgb(16pt)=(0.865395197,0.86541021,0.865395561);
  rgb(17pt)=(0.897787179,0.848937047,0.820880546);
  rgb(18pt)=(0.924127593,0.827384882,0.774508472);
  rgb(19pt)=(0.944468518,0.800927443,0.726736146);
  rgb(20pt)=(0.958852946,0.769767752,0.678007945);
  rgb(21pt)=(0.96732803,0.734132809,0.628751763);
  rgb(22pt)=(0.969954137,0.694266682,0.579375448);
  rgb(23pt)=(0.966811177,0.650421156,0.530263762);
  rgb(24pt)=(0.958003065,0.602842431,0.481775914);
  rgb(25pt)=(0.943660866,0.551750968,0.434243684);
  rgb(26pt)=(0.923944917,0.49730856,0.387970225);
  rgb(27pt)=(0.89904617,0.439559467,0.343229596);
  rgb(28pt)=(0.869186849,0.378313092,0.300267182);
  rgb(29pt)=(0.834620542,0.312874446,0.259301199);
  rgb(30pt)=(0.795631745,0.24128379,0.220525627);
  rgb(31pt)=(0.752534934,0.157246067,0.184115123);
  rgb(32pt)=(0.705673158,0.01555616,0.150232812)
},
point meta max=1,
point meta min=0.1,
tick align=outside,
tick pos=left,
x grid style={white!69.0196078431373!black},
xlabel={{\Large $\nicefrac{1}{d}\cdot$lengthscale $\ell$}},
xmin=0, xmax=12,
xtick style={color=black},
xtick={0.5,1.5,2.5,3.5,4.5,5.5,6.5,7.5,8.5,9.5,10.5,11.5},
xticklabel style = {rotate=90.0},
xticklabels={-8, -7, -6, -5, -4,
-3 , -2, -1, 0, 1, 2, 3},
y dir=reverse,
y grid style={white!69.0196078431373!black},
ymin=0, ymax=12,
ytick style={color=black},
ytick={0.5,1.5,2.5,3.5,4.5,5.5,6.5,7.5,8.5,9.5,10.5,11.5},
yticklabels={-20, -18, -16, -14,
-12, -10, -8, -6, -4, -2, 0, 2}
]
\addplot graphics [includegraphics cmd=\pgfimage,xmin=0, xmax=12, ymin=12, ymax=0] {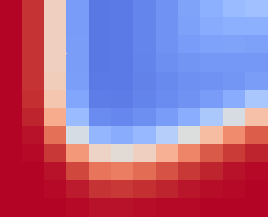};
\draw (axis cs:4.5,0.5) node[
  scale=2.25,
  text=white,
  rotate=0.0
]{*};
\end{axis}

\end{tikzpicture}
\vspace*{-0.3cm}
\caption{{\small Risk}}
\end{subfigure}
\begin{subfigure}{.24\textwidth}
\begin{tikzpicture}[scale=0.36]

\begin{axis}[
colorbar,
colorbar style={ytick={0.1,0.2,0.3,0.4,0.5,0.6,0.7,0.8,0.9,1},
yticklabels={0.1,0.2,0.3,0.4,0.5,0.6,0.7,0.8,0.9,1.0},ylabel={}},
colormap={mymap}{[1pt]
  rgb(0pt)=(0.2298057,0.298717966,0.753683153);
  rgb(1pt)=(0.26623388,0.353094838,0.801466763);
  rgb(2pt)=(0.30386891,0.406535296,0.84495867);
  rgb(3pt)=(0.342804478,0.458757618,0.883725899);
  rgb(4pt)=(0.38301334,0.50941904,0.917387822);
  rgb(5pt)=(0.424369608,0.558148092,0.945619588);
  rgb(6pt)=(0.46666708,0.604562568,0.968154911);
  rgb(7pt)=(0.509635204,0.648280772,0.98478814);
  rgb(8pt)=(0.552953156,0.688929332,0.995375608);
  rgb(9pt)=(0.596262162,0.726149107,0.999836203);
  rgb(10pt)=(0.639176211,0.759599947,0.998151185);
  rgb(11pt)=(0.681291281,0.788964712,0.990363227);
  rgb(12pt)=(0.722193294,0.813952739,0.976574709);
  rgb(13pt)=(0.761464949,0.834302879,0.956945269);
  rgb(14pt)=(0.798691636,0.849786142,0.931688648);
  rgb(15pt)=(0.833466556,0.860207984,0.901068838);
  rgb(16pt)=(0.865395197,0.86541021,0.865395561);
  rgb(17pt)=(0.897787179,0.848937047,0.820880546);
  rgb(18pt)=(0.924127593,0.827384882,0.774508472);
  rgb(19pt)=(0.944468518,0.800927443,0.726736146);
  rgb(20pt)=(0.958852946,0.769767752,0.678007945);
  rgb(21pt)=(0.96732803,0.734132809,0.628751763);
  rgb(22pt)=(0.969954137,0.694266682,0.579375448);
  rgb(23pt)=(0.966811177,0.650421156,0.530263762);
  rgb(24pt)=(0.958003065,0.602842431,0.481775914);
  rgb(25pt)=(0.943660866,0.551750968,0.434243684);
  rgb(26pt)=(0.923944917,0.49730856,0.387970225);
  rgb(27pt)=(0.89904617,0.439559467,0.343229596);
  rgb(28pt)=(0.869186849,0.378313092,0.300267182);
  rgb(29pt)=(0.834620542,0.312874446,0.259301199);
  rgb(30pt)=(0.795631745,0.24128379,0.220525627);
  rgb(31pt)=(0.752534934,0.157246067,0.184115123);
  rgb(32pt)=(0.705673158,0.01555616,0.150232812)
},
point meta max=1,
point meta min=0.1,
tick align=outside,
tick pos=left,
x grid style={white!69.0196078431373!black},
xlabel={{\Large $\nicefrac{1}{d}\cdot$lengthscale $\ell$}},
xmin=0, xmax=12,
xtick style={color=black},
xtick={0.5,1.5,2.5,3.5,4.5,5.5,6.5,7.5,8.5,9.5,10.5,11.5},
xticklabel style = {rotate=90.0},
xticklabels={-8, -7, -6, -5, -4,
-3 , -2, -1, 0, 1, 2, 3},
y dir=reverse,
y grid style={white!69.0196078431373!black},
ymin=0, ymax=12,
ytick style={color=black},
ytick={0.5,1.5,2.5,3.5,4.5,5.5,6.5,7.5,8.5,9.5,10.5,11.5},
yticklabels={-20, -18, -16, -14,
-12, -10, -8, -6, -4, -2, 0, 2}
]
\addplot graphics [includegraphics cmd=\pgfimage,xmin=0, xmax=12, ymin=12, ymax=0] {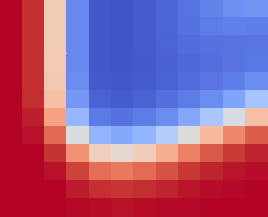};
\draw (axis cs:5.5,0.5) node[
  scale=2.25,
  text=white,
  rotate=0.0
]{*}; 
\end{axis}

\end{tikzpicture}
\vspace*{-0.3cm}
\caption{{\small KARE Predictions}}
\end{subfigure}
\begin{subfigure}{.24\textwidth}
\begin{tikzpicture}[scale=0.36]

\begin{axis}[
colorbar,
colorbar style={ytick={0.1,0.2,0.3,0.4,0.5,0.6,0.7,0.8,0.9,1},
yticklabels={0.1,0.2,0.3,0.4,0.5,0.6,0.7,0.8,0.9,1.0},ylabel={}},
colormap={mymap}{[1pt]
  rgb(0pt)=(0.2298057,0.298717966,0.753683153);
  rgb(1pt)=(0.26623388,0.353094838,0.801466763);
  rgb(2pt)=(0.30386891,0.406535296,0.84495867);
  rgb(3pt)=(0.342804478,0.458757618,0.883725899);
  rgb(4pt)=(0.38301334,0.50941904,0.917387822);
  rgb(5pt)=(0.424369608,0.558148092,0.945619588);
  rgb(6pt)=(0.46666708,0.604562568,0.968154911);
  rgb(7pt)=(0.509635204,0.648280772,0.98478814);
  rgb(8pt)=(0.552953156,0.688929332,0.995375608);
  rgb(9pt)=(0.596262162,0.726149107,0.999836203);
  rgb(10pt)=(0.639176211,0.759599947,0.998151185);
  rgb(11pt)=(0.681291281,0.788964712,0.990363227);
  rgb(12pt)=(0.722193294,0.813952739,0.976574709);
  rgb(13pt)=(0.761464949,0.834302879,0.956945269);
  rgb(14pt)=(0.798691636,0.849786142,0.931688648);
  rgb(15pt)=(0.833466556,0.860207984,0.901068838);
  rgb(16pt)=(0.865395197,0.86541021,0.865395561);
  rgb(17pt)=(0.897787179,0.848937047,0.820880546);
  rgb(18pt)=(0.924127593,0.827384882,0.774508472);
  rgb(19pt)=(0.944468518,0.800927443,0.726736146);
  rgb(20pt)=(0.958852946,0.769767752,0.678007945);
  rgb(21pt)=(0.96732803,0.734132809,0.628751763);
  rgb(22pt)=(0.969954137,0.694266682,0.579375448);
  rgb(23pt)=(0.966811177,0.650421156,0.530263762);
  rgb(24pt)=(0.958003065,0.602842431,0.481775914);
  rgb(25pt)=(0.943660866,0.551750968,0.434243684);
  rgb(26pt)=(0.923944917,0.49730856,0.387970225);
  rgb(27pt)=(0.89904617,0.439559467,0.343229596);
  rgb(28pt)=(0.869186849,0.378313092,0.300267182);
  rgb(29pt)=(0.834620542,0.312874446,0.259301199);
  rgb(30pt)=(0.795631745,0.24128379,0.220525627);
  rgb(31pt)=(0.752534934,0.157246067,0.184115123);
  rgb(32pt)=(0.705673158,0.01555616,0.150232812)
},
point meta max=1,
point meta min=0.1,
tick align=outside,
tick pos=left,
x grid style={white!69.0196078431373!black},
xlabel={{\Large $\nicefrac{1}{d}\cdot$lengthscale $\ell$}},
xmin=0, xmax=12,
xtick style={color=black},
xtick={0.5,1.5,2.5,3.5,4.5,5.5,6.5,7.5,8.5,9.5,10.5,11.5},
xticklabel style = {rotate=90.0},
xticklabels={-8, -7, -6, -5, -4,
-3 , -2, -1, 0, 1, 2, 3},
y dir=reverse,
y grid style={white!69.0196078431373!black},
ymin=0, ymax=12,
ytick style={color=black},
ytick={0.5,1.5,2.5,3.5,4.5,5.5,6.5,7.5,8.5,9.5,10.5,11.5},
yticklabels={-20, -18, -16, -14,
-12, -10, -8, -6, -4, -2, 0, 2}
]
\addplot graphics [includegraphics cmd=\pgfimage,xmin=0, xmax=12, ymin=12, ymax=0] {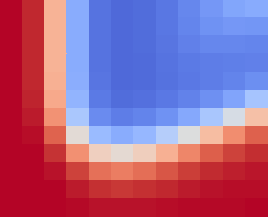};
\draw (axis cs:5.5,0.5) node[
  scale=2.25,
  text=white,
  rotate=0.0
]{*}; 
\end{axis}

\end{tikzpicture}
\vspace*{-0.3cm}
\caption{{\small Cross Val. Predictions}}
\end{subfigure}
\begin{subfigure}{.24\textwidth}
\begin{tikzpicture}[scale=0.36]

\begin{axis}[
colorbar,
colorbar style={ylabel={}},
colormap={mymap}{[1pt]
  rgb(0pt)=(0.2298057,0.298717966,0.753683153);
  rgb(1pt)=(0.26623388,0.353094838,0.801466763);
  rgb(2pt)=(0.30386891,0.406535296,0.84495867);
  rgb(3pt)=(0.342804478,0.458757618,0.883725899);
  rgb(4pt)=(0.38301334,0.50941904,0.917387822);
  rgb(5pt)=(0.424369608,0.558148092,0.945619588);
  rgb(6pt)=(0.46666708,0.604562568,0.968154911);
  rgb(7pt)=(0.509635204,0.648280772,0.98478814);
  rgb(8pt)=(0.552953156,0.688929332,0.995375608);
  rgb(9pt)=(0.596262162,0.726149107,0.999836203);
  rgb(10pt)=(0.639176211,0.759599947,0.998151185);
  rgb(11pt)=(0.681291281,0.788964712,0.990363227);
  rgb(12pt)=(0.722193294,0.813952739,0.976574709);
  rgb(13pt)=(0.761464949,0.834302879,0.956945269);
  rgb(14pt)=(0.798691636,0.849786142,0.931688648);
  rgb(15pt)=(0.833466556,0.860207984,0.901068838);
  rgb(16pt)=(0.865395197,0.86541021,0.865395561);
  rgb(17pt)=(0.897787179,0.848937047,0.820880546);
  rgb(18pt)=(0.924127593,0.827384882,0.774508472);
  rgb(19pt)=(0.944468518,0.800927443,0.726736146);
  rgb(20pt)=(0.958852946,0.769767752,0.678007945);
  rgb(21pt)=(0.96732803,0.734132809,0.628751763);
  rgb(22pt)=(0.969954137,0.694266682,0.579375448);
  rgb(23pt)=(0.966811177,0.650421156,0.530263762);
  rgb(24pt)=(0.958003065,0.602842431,0.481775914);
  rgb(25pt)=(0.943660866,0.551750968,0.434243684);
  rgb(26pt)=(0.923944917,0.49730856,0.387970225);
  rgb(27pt)=(0.89904617,0.439559467,0.343229596);
  rgb(28pt)=(0.869186849,0.378313092,0.300267182);
  rgb(29pt)=(0.834620542,0.312874446,0.259301199);
  rgb(30pt)=(0.795631745,0.24128379,0.220525627);
  rgb(31pt)=(0.752534934,0.157246067,0.184115123);
  rgb(32pt)=(0.705673158,0.01555616,0.150232812)
},
point meta max=0.1,
point meta min=0,
tick align=outside,
tick pos=left,
x grid style={white!69.0196078431373!black},
xlabel={{\Large $\nicefrac{1}{d}\cdot$lengthscale $\ell$}},
xmin=0, xmax=12,
xtick style={color=black},
xtick={0.5,1.5,2.5,3.5,4.5,5.5,6.5,7.5,8.5,9.5,10.5,11.5},
xticklabel style = {rotate=90.0},
xticklabels={-8, -7, -6, -5, -4,
-3 , -2, -1, 0, 1, 2, 3},
y dir=reverse,
y grid style={white!69.0196078431373!black},
ymin=0, ymax=12,
ytick style={color=black},
ytick={0.5,1.5,2.5,3.5,4.5,5.5,6.5,7.5,8.5,9.5,10.5,11.5},
yticklabels={-20, -18, -16, -14,
-12, -10, -8, -6, -4, -2, 0, 2}
]
\addplot graphics [includegraphics cmd=\pgfimage,xmin=0, xmax=12, ymin=12, ymax=0] {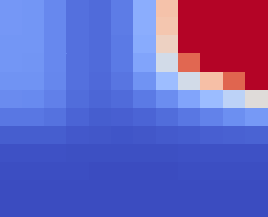};
\draw (axis cs:5.5,10.5) node[
  scale=2.25,
  text=white,
  rotate=0.0
]{*}; 
\end{axis}

\end{tikzpicture}
\vspace*{-0.3cm}
\caption{{\small Log-likelihood Estim.}}
\end{subfigure}
\caption{\textit{Comparision of risk predictors.} We calculate the risk (i.e. test error) of $\krrf$ on MNIST with the RBF Kernel for various values of $\ell$ and $\lambda$ on $N=200$ data points (same setup as Fig. \ref{fig:KARE-predicts-risk}).
We mark the minimum MSE achieved with a star.
We display the predictions of KARE and $4$-fold cross-validation; both find the hyper-parameters minimizing the risk.
We also show the (normalized) log-likehood estimator and observe that it favors large $\lambda$ values. Axes are $\log_2$ scale.}
\label{fig:comparision-predictors}
\end{figure}
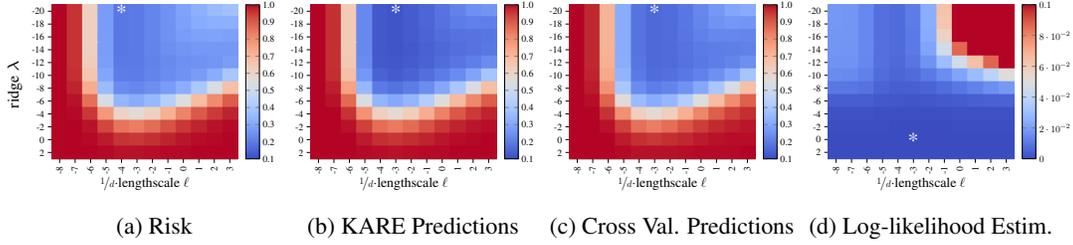

The KARE can be used to optimize the risk over the space of kernels, for instance to choose the ridge and length-scale. The most popular kernel selection techniques are (see Figure \ref{fig:comparision-predictors}):
\begin{itemize}[leftmargin=*]
  \item Cross-validation: accurate estimator of the risk on a test set, but costly to optimize (the predictor must be recomputed for each kernel and differentiating it in the space of kernels is hard).
  \item Kernel likelihood (Chapter 5 of \cite{Rasmussen-06-GaussianProcess}): efficient to optimize and takes into account the ridge, but not a risk estimator; unlike the risk, not invariant under the simultaneous rescaling $K, \lambda \leadsto \alpha K, \alpha \lambda$.
  \item Classical kernel alignment \cite{cristianini-2002}: very efficient to optimize and scale invariant, but not a risk estimator, not sensitive to small eigenvalues and inadequate to select hyperparameters such as the ridge.
\end{itemize}
The KARE combines the best features of the three above techniques:
\begin{itemize}[leftmargin=*]
  \item it can be computed efficiently on the training data, and optimized over the space of kernels;
  \item like the risk, it is invariant under the simultaneous rescaling $K, \lambda \leadsto \alpha K, \alpha \lambda$;
  \item it is sensitive to the small Gram matrix eigenvalues and to the ridge $ \lambda $.
\end{itemize}

\section{Conclusion}\label{sec:conclusion}
In this paper, we introduce new techniques to study the Kernel Ridge Regression (KRR) predictor and its risk. We obtain new precise estimates for the test and train error in terms of a new object, the Signal Capture Threshold (SCT), which identifies the components of a true function that are being learned by the KRR: our estimates reveal a remarkable relation, which leads one to the Kernel Alignment Risk Estimator (KARE). The KARE is a new efficient way to estimate the risk of a kernel predictor based on the training data only.
Numerically, we observe that the KARE gives a very accurate prediction of the risk for Higgs and MNIST datasets for a variety of classical kernels.

\section*{Broader Impact}
This work is fundamental and may be used in any research area using Kernel methods, possibly leading to indirect social impacts.
However, we do not predict any direct social impact.

\section*{Acknowledgements}
The authors wish to thank A. Montanari and M. Wyart for useful discussions. This work is partly supported by the ERC SG CONSTAMIS. C. Hongler acknowledges support from the Blavatnik Family Foundation, the Latsis Foundation, and the the NCCR Swissmap.

\bibliographystyle{abbrv}
\bibliography{../main}

\clearpage

\appendix
\section*{Appendix}

We organize the Appendix as follows:
\begin{enumerate}
\item In Section \ref{sec:numerical-results}, we present the details for the numerical results
presented in the main text (and in the Appendix) and we present additional experiments and some discussions.
\item In Section \ref{sec:proofs}, we present the proofs of the mathematical results
presented in the main text.
\end{enumerate}

\section{Numerical Results}\label{sec:numerical-results}

\subsection{Empirical Methods}

\textbf{For the MNIST dataset.}
We sample $N$ images of digits $7$ and $9$ from the MNIST training dataset (image size $d=24\times24$, edge pixels cropped, all pixels rescaled down to $[0,1]$ and recentered around the mean value) and label each of them with $+1$ and $-1$ labels.
We perform KRR with various ridge $\lambda$ on this dataset with the selected kernel $k$ times and calculate the MSE training error, risk, and the KARE for every trial ($k=10$ for small $N$ and $k=5$ for $N=2000$).
The risk is approximated using other $N_2 = 1000$ random samples of the MNIST training data. \\

\textbf{For the Higgs Dataset.}
We randomly choose $N$ samples among those that do not have any missing features marked with $-999$ from the Higgs training dataset.
The samples have $d=31$ features, and we normalize each feature column down to $[0, 1]$ by dividing by the maximum absolute value observed among the selected samples.
We replace the categorical labels `s' and `b' with regression values $+1$ and $-1$ respectively and perform KRR with various ridge $\lambda$.
We repeat this procedure $k$ times, which corresponds to sampling $k$ different training datasets of $N=1000$ samples to perform kernel regression, and calculate the MSE training error, the risk, and the KARE for every trial ($k=10$ for small $N$ and $k=5$ for $N=1000$).
The risk is approximated using other $N_2 = 1000$ random samples of the Higgs training data.

\clearpage

\subsection{KARE predicts risk for various Kernels}

\begin{figure}[!h]
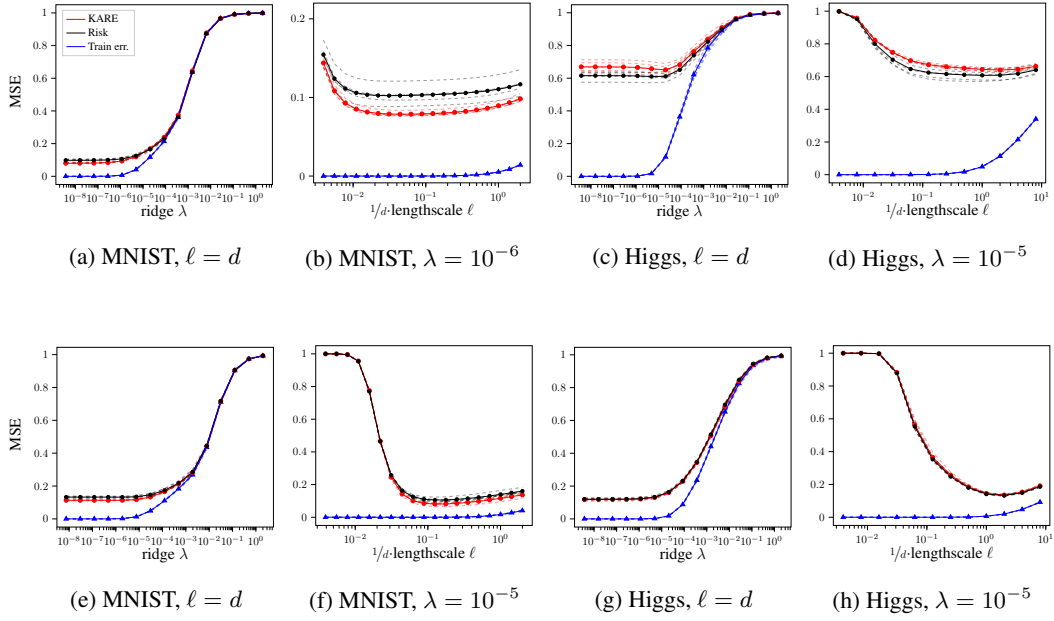

\rotatebox[origin=l]{90}{\tiny{\textcolor{white}{hey!!}MSE}}
\begin{subfigure}{.24\textwidth}
\input{MNIST-approx-lambd-N2000-lap.tex}
\vspace*{-0.3cm}
\caption{{\small MNIST, $\ell = d$}}
\end{subfigure}
\begin{subfigure}{.24\textwidth}
\input{MNIST-approx-len-N2000-lap.tex}
\vspace*{-0.3cm}
\caption{{\small MNIST, $\lambda = 10^{-6}$}}
\end{subfigure}
\begin{subfigure}{.24\textwidth}
\input{Higgs-approx-lambd-N1000-lap.tex}
\vspace*{-0.3cm}
\caption{{\small Higgs, $\ell = d$}}
\end{subfigure}
\begin{subfigure}{.24\textwidth}
\begin{tikzpicture}[scale=0.42]

\begin{axis}[
legend cell align={left},
legend style={fill opacity=0.8, draw opacity=1, text opacity=1, draw=white!80!black},
log basis x={10},
tick align=outside,
tick pos=left,
x grid style={white!69.0196078431373!black},
xlabel={{\Large $\nicefrac{1}{d}\cdot$lengthscale $\ell$}},
xmin=0.00266804737647343, xmax=11.712685567565,
xmode=log,
xtick style={color=black},
y grid style={white!69.0196078431373!black},
ymin=-0.0498606895861323, ymax=1.04914384695759,
ytick style={color=black},
]
\addplot [semithick, black, dashed, opacity=0.4, forget plot]
table {%
0.00390625 0.999189095296515
0.0078125 0.963580862888775
0.015625 0.830446160163866
0.03125 0.732603503988334
0.0625 0.669859498787029
0.125 0.645438213740387
0.25 0.635467260041071
0.5 0.629231431759497
1 0.624337550944668
2 0.62270617680658
4 0.629354421492725
8 0.649666531931558
};
\addplot [semithick, blue, dashed, opacity=0.4, forget plot]
table {%
0.00390625 9.78808526917807e-05
0.0078125 9.42479408647524e-05
0.015625 9.59911904001406e-05
0.03125 0.00017010408573458
0.0625 0.000465002686219926
0.125 0.00154553953326554
0.25 0.00529102530779814
0.5 0.0169694176200165
1 0.047613122717774
2 0.111063454172016
4 0.211657241738543
8 0.337052803068768
};
\addplot [semithick, red, dashed, opacity=0.4, forget plot]
table {%
0.00390625 0.998461594416508
0.0078125 0.955798136277195
0.015625 0.824718707979675
0.03125 0.746771185336392
0.0625 0.690836004569931
0.125 0.663458100730269
0.25 0.64973244115675
0.5 0.640604189646321
1 0.633527014423191
2 0.630395930372824
4 0.63639066224805
8 0.656965687951312
};
\addplot [semithick, black, dashed, opacity=0.4, forget plot]
table {%
0.00390625 0.997885989098802
0.0078125 0.944839896943824
0.015625 0.781969395353079
0.03125 0.674109030429809
0.0625 0.60970989612637
0.125 0.584383444043887
0.25 0.574697837720036
0.5 0.570841834897078
1 0.570722503848792
2 0.576074326523956
4 0.590958094210779
8 0.619309497940112
};
\addplot [semithick, blue, dashed, opacity=0.4, forget plot]
table {%
0.00390625 9.7892004240696e-05
0.0078125 9.40620749461562e-05
0.015625 9.6988779780971e-05
0.03125 0.000175183512339365
0.0625 0.000485455366592071
0.125 0.00162474953274187
0.25 0.00556983265542303
0.5 0.0178365770499053
1 0.0498866005441074
2 0.115846839902673
4 0.219348439808368
8 0.345951705818386
};
\addplot [semithick, red, dashed, opacity=0.4, forget plot]
table {%
0.00390625 0.998574038929051
0.0078125 0.953274690882638
0.015625 0.826893025951768
0.03125 0.757579914439937
0.0625 0.708530786911926
0.125 0.685060383001758
0.25 0.672501821308631
0.5 0.663366664535276
1 0.655820218139335
2 0.65183160243822
4 0.655818379379706
8 0.672127026385377
};
\addplot [semithick, black, dashed, opacity=0.4, forget plot]
table {%
0.00390625 0.998875979168163
0.0078125 0.958641364451749
0.015625 0.829206542063727
0.03125 0.747566943388767
0.0625 0.694198482074129
0.125 0.671049479351775
0.25 0.660558115965156
0.5 0.654277953647265
1 0.650219320266713
2 0.65018202995415
4 0.658351835370454
8 0.678319099323309
};
\addplot [semithick, blue, dashed, opacity=0.4, forget plot]
table {%
0.00390625 9.79313614077522e-05
0.0078125 9.45480541198534e-05
0.015625 9.52679887082169e-05
0.03125 0.000167762204309381
0.0625 0.000462883387710768
0.125 0.00155149591720581
0.25 0.0053354773918597
0.5 0.0171585471438252
1 0.0482337555260079
2 0.112551674247541
4 0.213783523458017
8 0.337464423999228
};
\addplot [semithick, red, dashed, opacity=0.4, forget plot]
table {%
0.00390625 0.99897116426229
0.0078125 0.958952824052502
0.015625 0.82089589898121
0.03125 0.741653159673176
0.0625 0.693826509755323
0.125 0.672239145804145
0.25 0.661108836961915
0.5 0.653064396822965
1 0.64623516644282
2 0.642250923519312
4 0.645173938795732
8 0.659337307600765
};
\addplot [semithick, black, dashed, opacity=0.4, forget plot]
table {%
0.00390625 0.997987981276778
0.0078125 0.946546596595342
0.015625 0.773434213390064
0.03125 0.673051686192354
0.0625 0.618085164403321
0.125 0.597106032442697
0.25 0.588201768490058
0.5 0.582662488095282
1 0.578770025097874
2 0.578816506715694
4 0.588196875527057
8 0.61281743493473
};
\addplot [semithick, blue, dashed, opacity=0.4, forget plot]
table {%
0.00390625 9.7940856417819e-05
0.0078125 9.47884465167985e-05
0.015625 9.54082460319555e-05
0.03125 0.000170040966853914
0.0625 0.000470950291266491
0.125 0.00157918343512002
0.25 0.00542823746213216
0.5 0.0174393702703179
1 0.0489613802750339
2 0.114178051025048
4 0.217194337924263
8 0.344479511853387
};
\addplot [semithick, red, dashed, opacity=0.4, forget plot]
table {%
0.00390625 0.999076975605939
0.0078125 0.961334137573119
0.015625 0.81893739191208
0.03125 0.745313109830598
0.0625 0.698392140103858
0.125 0.676621154855844
0.25 0.665336694787957
0.5 0.657156292645379
1 0.650363275468553
2 0.647088123189558
4 0.652197594856724
8 0.670768325658788
};
\addplot [semithick, black, dashed, opacity=0.4, forget plot]
table {%
0.00390625 0.998118871770168
0.0078125 0.947481536935785
0.015625 0.78803516931202
0.03125 0.689637529688298
0.0625 0.638643874438759
0.125 0.623632056811407
0.25 0.618951036472365
0.5 0.615876034461704
1 0.613195029358836
2 0.613148670958981
4 0.620771754183597
8 0.64173345078619
};
\addplot [semithick, blue, dashed, opacity=0.4, forget plot]
table {%
0.00390625 9.79486407793603e-05
0.0078125 9.45324403833002e-05
0.015625 9.67865692993926e-05
0.03125 0.000173487762356874
0.0625 0.0004757818162814
0.125 0.00158320341618668
0.25 0.00541281996263009
0.5 0.017304801216535
1 0.0483353246430107
2 0.11217850544299
4 0.212718279279855
8 0.337012593203936
};
\addplot [semithick, red, dashed, opacity=0.4, forget plot]
table {%
0.00390625 0.999150854130598
0.0078125 0.957988681512429
0.015625 0.824726113936043
0.03125 0.748495022805265
0.0625 0.691835531955502
0.125 0.664680263158025
0.25 0.650666353891728
0.5 0.640840816181286
1 0.632924730598077
2 0.629048131975774
4 0.634283476984506
8 0.653470129966561
};
\addplot [thick, blue]
table {%
0.00390625 9.79187431074817e-05
0.0078125 9.44357913661721e-05
0.015625 9.60885548441353e-05
0.03125 0.000171315706318823
0.0625 0.000472014709614131
0.125 0.00157683436690398
0.25 0.00540747855596863
0.5 0.01734174266012
1 0.0486060367411868
2 0.113163704958054
4 0.214940364441809
8 0.340392207588741
};
\addlegendentry{Train err.}
\addplot [thick, red]
table {%
0.00390625 0.998846925468877
0.0078125 0.957469694059576
0.015625 0.823234227752155
0.03125 0.747962478417074
0.0625 0.696684194659308
0.125 0.672411809510008
0.25 0.659869229621396
0.5 0.651006471966245
1 0.643774081014395
2 0.640122942299137
4 0.644772810452944
8 0.662533695512561
};
\addlegendentry{KARE}
\addplot [thick, black]
table {%
0.00390625 0.998411583322085
0.0078125 0.952218051563095
0.015625 0.800618296056551
0.03125 0.703393738737512
0.0625 0.646099383165922
0.125 0.624321845278031
0.25 0.615575203737738
0.5 0.610577948572165
1 0.607448885903376
2 0.608185542191872
4 0.617526596156922
8 0.64036920298318
};
\addlegendentry{Risk}
\addplot [thick, blue, mark=triangle*, mark size=2.2, mark options={solid}, only marks, forget plot]
table {%
0.00390625 9.79187431074817e-05
0.0078125 9.44357913661721e-05
0.015625 9.60885548441353e-05
0.03125 0.000171315706318823
0.0625 0.000472014709614131
0.125 0.00157683436690398
0.25 0.00540747855596863
0.5 0.01734174266012
1 0.0486060367411868
2 0.113163704958054
4 0.214940364441809
8 0.340392207588741
};
\addplot [thick, red, mark=*, mark size=1.8, mark options={solid}, only marks, forget plot]
table {%
0.00390625 0.998846925468877
0.0078125 0.957469694059576
0.015625 0.823234227752155
0.03125 0.747962478417074
0.0625 0.696684194659308
0.125 0.672411809510008
0.25 0.659869229621396
0.5 0.651006471966245
1 0.643774081014395
2 0.640122942299137
4 0.644772810452944
8 0.662533695512561
};
\addplot [thick, black, mark=*, mark size=1.5, mark options={solid}, only marks, forget plot]
table {%
0.00390625 0.998411583322085
0.0078125 0.952218051563095
0.015625 0.800618296056551
0.03125 0.703393738737512
0.0625 0.646099383165922
0.125 0.624321845278031
0.25 0.615575203737738
0.5 0.610577948572165
1 0.607448885903376
2 0.608185542191872
4 0.617526596156922
8 0.64036920298318
};
\legend{};
\end{axis}

\end{tikzpicture}
\vspace*{-0.3cm}
\caption{{\small Higgs, $\lambda=10^{-5}$}}
\end{subfigure}
\bigskip
\vspace{0.5cm}
\rotatebox[origin=l]{90}{\tiny{\textcolor{white}{hey!!}MSE}}
\begin{subfigure}{.24\textwidth}
\input{MNIST-approx-lambd-N2000-one-ker.tex}
\vspace*{-0.3cm}
\caption{{\small MNIST, $\ell = d$}}
\end{subfigure}
\begin{subfigure}{.24\textwidth}
\input{MNIST-approx-len-N2000-one-ker.tex}
\vspace*{-0.3cm}
\caption{{\small MNIST, $\lambda = 10^{-5}$}}
\end{subfigure}
\begin{subfigure}{.24\textwidth}
\input{Higgs-approx-lambd-N1000-one-ker.tex}
\vspace*{-0.3cm}
\caption{{\small Higgs, $\ell = d$}}
\end{subfigure}
\begin{subfigure}{.24\textwidth}
\begin{tikzpicture}[scale=0.42]

\begin{axis}[
legend cell align={left},
legend style={fill opacity=0.8, draw opacity=1, text opacity=1, draw=white!80!black},
log basis x={10},
tick align=outside,
tick pos=left,
x grid style={white!69.0196078431373!black},
xlabel={{\Large $\nicefrac{1}{d}\cdot$lengthscale $\ell$}},
xmin=0.00266804737647343, xmax=11.712685567565,
xmode=log,
xtick style={color=black},
y grid style={white!69.0196078431373!black},
ymin=-0.0497952835974343, ymax=1.0457009872722,
ytick style={color=black}
]
\addplot [semithick, black, dashed, opacity=0.4, forget plot]
table {%
0.00390625 0.999999991795288
0.0078125 0.999980978298632
0.015625 0.996810399502379
0.03125 0.874744018794644
0.0625 0.53958613747425
0.125 0.342377982193503
0.25 0.242631874883769
0.5 0.177112024906478
1 0.139520144197166
2 0.13064302913677
4 0.1467522761559
8 0.181507006065428
};
\addplot [semithick, blue, dashed, opacity=0.4, forget plot]
table {%
0.00390625 9.80296045038744e-05
0.0078125 9.80281675800562e-05
0.015625 9.77534237915674e-05
0.03125 8.73415001303433e-05
0.0625 7.10360363763141e-05
0.125 0.0001183231197214
0.25 0.000389390784520986
0.5 0.00162562508641408
1 0.00628963312043568
2 0.0193391616064205
4 0.0464291791042704
8 0.0901061738769343
};
\addplot [semithick, red, dashed, opacity=0.4, forget plot]
table {%
0.00390625 0.999999995544028
0.0078125 0.999985331774175
0.015625 0.997150611990982
0.03125 0.881662127309366
0.0625 0.556992550523794
0.125 0.36084432949969
0.25 0.254217101271485
0.5 0.183382600789964
1 0.142951983203804
2 0.133132170488125
4 0.150560299963331
8 0.188562921884853
};
\addplot [semithick, black, dashed, opacity=0.4, forget plot]
table {%
0.00390625 0.999999994180901
0.0078125 0.999982742355365
0.015625 0.996527066031307
0.03125 0.871327836668677
0.0625 0.543654390442027
0.125 0.349008093217666
0.25 0.245956786872868
0.5 0.177404890925543
1 0.138626628755391
2 0.128772288539018
4 0.144709627050172
8 0.181600968832562
};
\addplot [semithick, blue, dashed, opacity=0.4, forget plot]
table {%
0.00390625 9.80295996175869e-05
0.0078125 9.80280558429101e-05
0.015625 9.77994153690565e-05
0.03125 8.71883514511671e-05
0.0625 7.12565261430527e-05
0.125 0.000120089231874694
0.25 0.000403697778622917
0.5 0.00168843903703062
1 0.00647679716754514
2 0.0198418728281226
4 0.0476876737764176
8 0.0927956736355705
};
\addplot [semithick, red, dashed, opacity=0.4, forget plot]
table {%
0.00390625 0.999999945698252
0.0078125 0.99998414124962
0.015625 0.99761638903521
0.03125 0.880400431665241
0.0625 0.559013594177697
0.125 0.36632473910466
0.25 0.263551183217651
0.5 0.190352734631997
1 0.146980780872973
2 0.136291893002501
4 0.154320974247649
8 0.193909537385051
};
\addplot [semithick, black, dashed, opacity=0.4, forget plot]
table {%
0.00390625 0.999999934092736
0.0078125 0.999971527781344
0.015625 0.997032250193662
0.03125 0.894070589755647
0.0625 0.588464687673022
0.125 0.369222804356223
0.25 0.253299398836029
0.5 0.181923482858109
1 0.142480590122438
2 0.133520394442682
4 0.150754042817669
8 0.188637404598047
};
\addplot [semithick, blue, dashed, opacity=0.4, forget plot]
table {%
0.00390625 9.80296045083581e-05
0.0078125 9.80281400953093e-05
0.015625 9.77394043632402e-05
0.03125 8.7926801761003e-05
0.0625 7.27378361388394e-05
0.125 0.000119593204418117
0.25 0.00039024034853632
0.5 0.00163890632215867
1 0.00640069250247142
2 0.0196792160123804
4 0.0468477494016625
8 0.090612183109664
};
\addplot [semithick, red, dashed, opacity=0.4, forget plot]
table {%
0.00390625 0.999999995589764
0.0078125 0.99998505245383
0.015625 0.997010460360798
0.03125 0.88709579103249
0.0625 0.56601060959372
0.125 0.358335643454508
0.25 0.24870598211111
0.5 0.180111192588272
1 0.142137766885693
2 0.133237275977366
4 0.150371886894879
8 0.188473164257123
};
\addplot [semithick, black, dashed, opacity=0.4, forget plot]
table {%
0.00390625 0.999999998073485
0.0078125 0.999988737593598
0.015625 0.996765115873263
0.03125 0.881804458648087
0.0625 0.558995572010963
0.125 0.351112787675949
0.25 0.242944003338666
0.5 0.175022459894793
1 0.13751451498645
2 0.129250087718362
4 0.147012233142882
8 0.18491549237193
};
\addplot [semithick, blue, dashed, opacity=0.4, forget plot]
table {%
0.00390625 9.80296054348151e-05
0.0078125 9.80288551425171e-05
0.015625 9.77608504898173e-05
0.03125 8.81486504919895e-05
0.0625 7.62144742153094e-05
0.125 0.000130727708841835
0.25 0.000425110032999267
0.5 0.00173323776479927
1 0.00663436178057722
2 0.020406372388672
4 0.0489557514102553
8 0.0947181008095441
};
\addplot [semithick, red, dashed, opacity=0.4, forget plot]
table {%
0.00390625 1.00000000504051
0.0078125 0.999992325095326
0.015625 0.99720685200182
0.03125 0.889244400608462
0.0625 0.596201681382709
0.125 0.397390155617589
0.25 0.276416839729758
0.5 0.194583556171815
1 0.150063560769455
2 0.139927097572536
4 0.158298484635713
8 0.197819551494312
};
\addplot [semithick, black, dashed, opacity=0.4, forget plot]
table {%
0.00390625 0.999999994201513
0.0078125 0.999990012318384
0.015625 0.996965335076894
0.03125 0.870502963553008
0.0625 0.535427499378242
0.125 0.354587405726463
0.25 0.257932415642241
0.5 0.189925997791323
1 0.149556915024322
2 0.140133997769565
4 0.158069951844177
8 0.195822777869554
};
\addplot [semithick, blue, dashed, opacity=0.4, forget plot]
table {%
0.00390625 9.8029600791613e-05
0.0078125 9.80263177825163e-05
0.015625 9.76688683824524e-05
0.03125 8.67773861229216e-05
0.0625 6.93839016988577e-05
0.125 0.000115689400894628
0.25 0.000383041442927438
0.5 0.00162686400641169
1 0.0064228905640946
2 0.0199121250279636
4 0.0473915237806941
8 0.0908209960236509
};
\addplot [semithick, red, dashed, opacity=0.4, forget plot]
table {%
0.00390625 0.999999957675067
0.0078125 0.999966418323694
0.015625 0.996253857458156
0.03125 0.874103828869215
0.0625 0.536705257835064
0.125 0.344294225798806
0.25 0.243208412164308
0.5 0.178924450101258
1 0.143183857799526
2 0.135363591637009
4 0.152548172998391
8 0.189250039610299
};
\addplot [thick, red]
table {%
0.00390625 0.999999979909524
0.0078125 0.999982653779329
0.015625 0.997047634169393
0.03125 0.882501315896955
0.0625 0.562984738702597
0.125 0.36543781869505
0.25 0.257219903698863
0.5 0.185470906856661
1 0.14506358990629
2 0.135590405735507
4 0.153219963747992
8 0.191603042926328
};
\addlegendentry{KARE}
\addplot [thick, black]
table {%
0.00390625 0.999999982468785
0.0078125 0.999982799669464
0.015625 0.996820033335501
0.03125 0.878489973484013
0.0625 0.553225657395701
0.125 0.353261814633961
0.25 0.248552895914715
0.5 0.180277771275249
1 0.141539758617154
2 0.132463959521279
4 0.14945962620216
8 0.186496729947504
};
\addlegendentry{risk}
\addplot [thick, blue]
table {%
0.00390625 9.80296029712495e-05
0.0078125 9.80279072886618e-05
0.015625 9.77443924792267e-05
0.03125 8.74765379914849e-05
0.0625 7.21257549144746e-05
0.125 0.000120884533150135
0.25 0.000398296077521385
0.5 0.00166261444336287
1 0.00644487502702481
2 0.0198357495727118
4 0.04746237549466
8 0.0918106254910727
};
\addlegendentry{train err.}
\addplot [thick, blue, mark=triangle*, mark size=2.2, mark options={solid}, only marks, forget plot]
table {%
0.00390625 9.80296029712495e-05
0.0078125 9.80279072886618e-05
0.015625 9.77443924792267e-05
0.03125 8.74765379914849e-05
0.0625 7.21257549144746e-05
0.125 0.000120884533150135
0.25 0.000398296077521385
0.5 0.00166261444336287
1 0.00644487502702481
2 0.0198357495727118
4 0.04746237549466
8 0.0918106254910727
};
\addplot [thick, red, mark=*, mark size=1.8, mark options={solid}, only marks, forget plot]
table {%
0.00390625 0.999999979909524
0.0078125 0.999982653779329
0.015625 0.997047634169393
0.03125 0.882501315896955
0.0625 0.562984738702597
0.125 0.36543781869505
0.25 0.257219903698863
0.5 0.185470906856661
1 0.14506358990629
2 0.135590405735507
4 0.153219963747992
8 0.191603042926328
};
\addplot [thick, black, mark=*, mark size=1.5, mark options={solid}, only marks, forget plot]
table {%
0.00390625 0.999999982468785
0.0078125 0.999982799669464
0.015625 0.996820033335501
0.03125 0.878489973484013
0.0625 0.553225657395701
0.125 0.353261814633961
0.25 0.248552895914715
0.5 0.180277771275249
1 0.141539758617154
2 0.132463959521279
4 0.14945962620216
8 0.186496729947504
};
\legend{};
\end{axis};

\end{tikzpicture}
\vspace*{-0.3cm}
\caption{{\small Higgs, $\lambda=10^{-5}$}}
\end{subfigure}
\caption{Comparison between the KRR risk and the KARE for various choices of normalized lengthscale $\nicefrac{\ell}{d}$ and ridge $\lambda$ on the MNIST dataset (restricted to the digits $ 7 $ and $ 9 $, labeled by $ 1 $ and $ -1 $ respectively, $N=2000$) and on the Higgs dataset (classes `b' and `s', labeled by $-1$ and $1$, $N=1000$). We present the results for the Laplacian Kernel $K(x,x') = \exp(\nicefrac{-\|x-x'\|_2}{\ell})$ (top row) and the $\ell_1$-norm Kernel $K(x,x') = \exp(\nicefrac{-\|x-x'\|_1}{\ell})$ (bottom row).
KRR predictor risks, and KARE curves (shown as dashed lines, $5$ samples) concentrate around their respective averages (solid lines).}
\end{figure}

\subsection{KRR predictor in function space}

\begin{figure}[h]
    \centering
    {
        \includegraphics[width=0.23\textwidth]{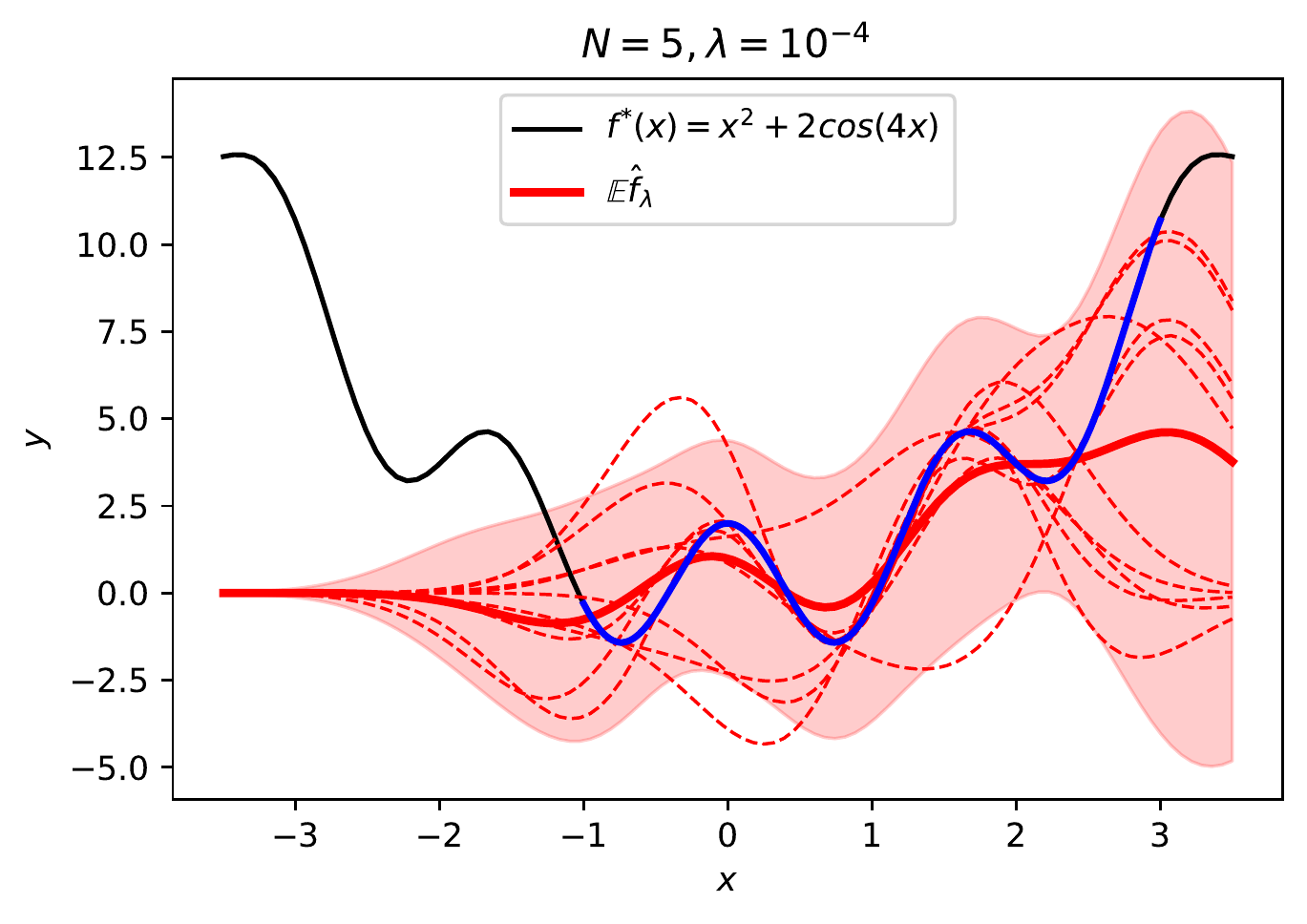}
        } \hfill
   {
        \includegraphics[width=0.23\textwidth]{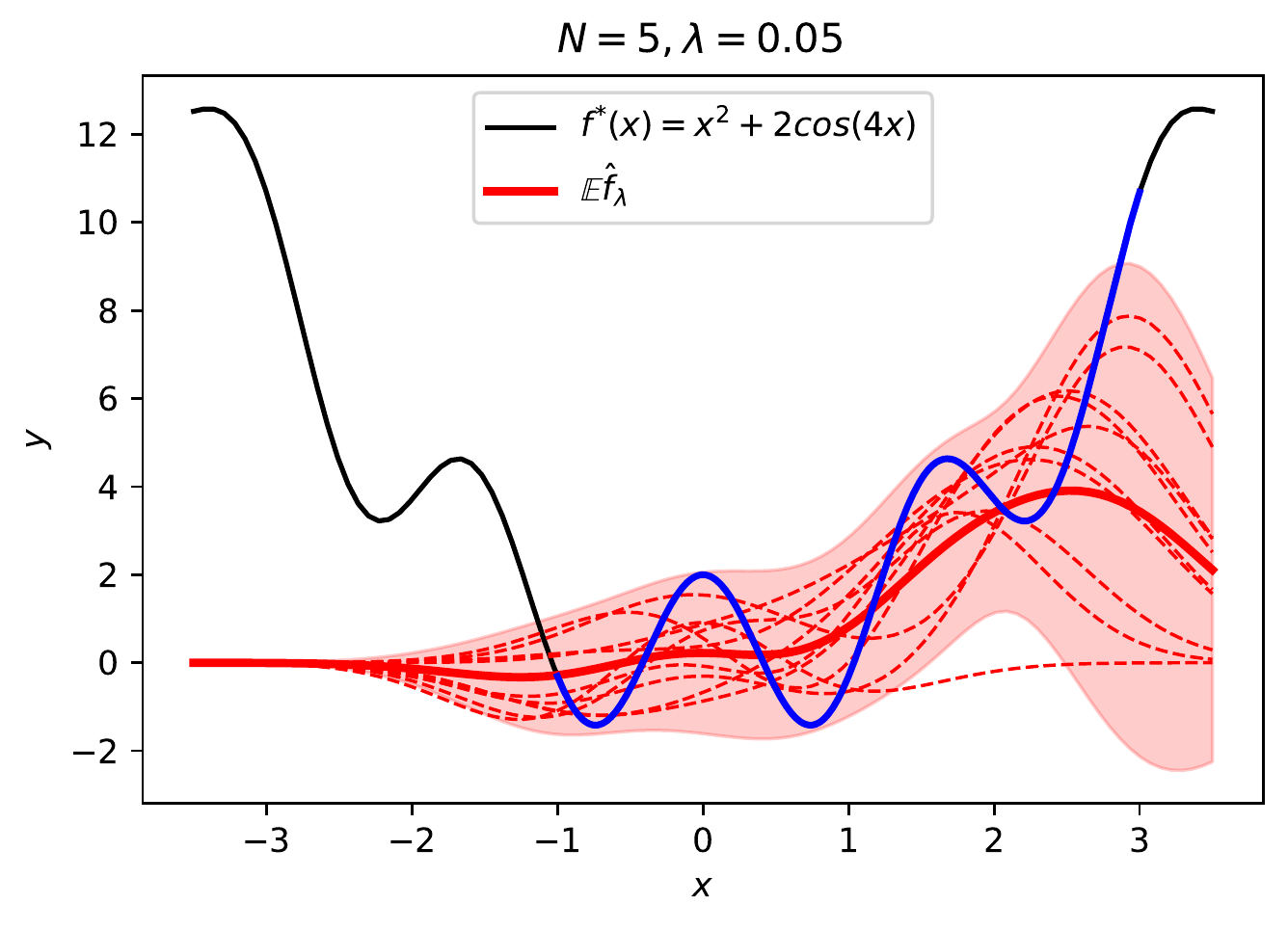}
        } \hfill
    {
        \includegraphics[width=0.23\textwidth]{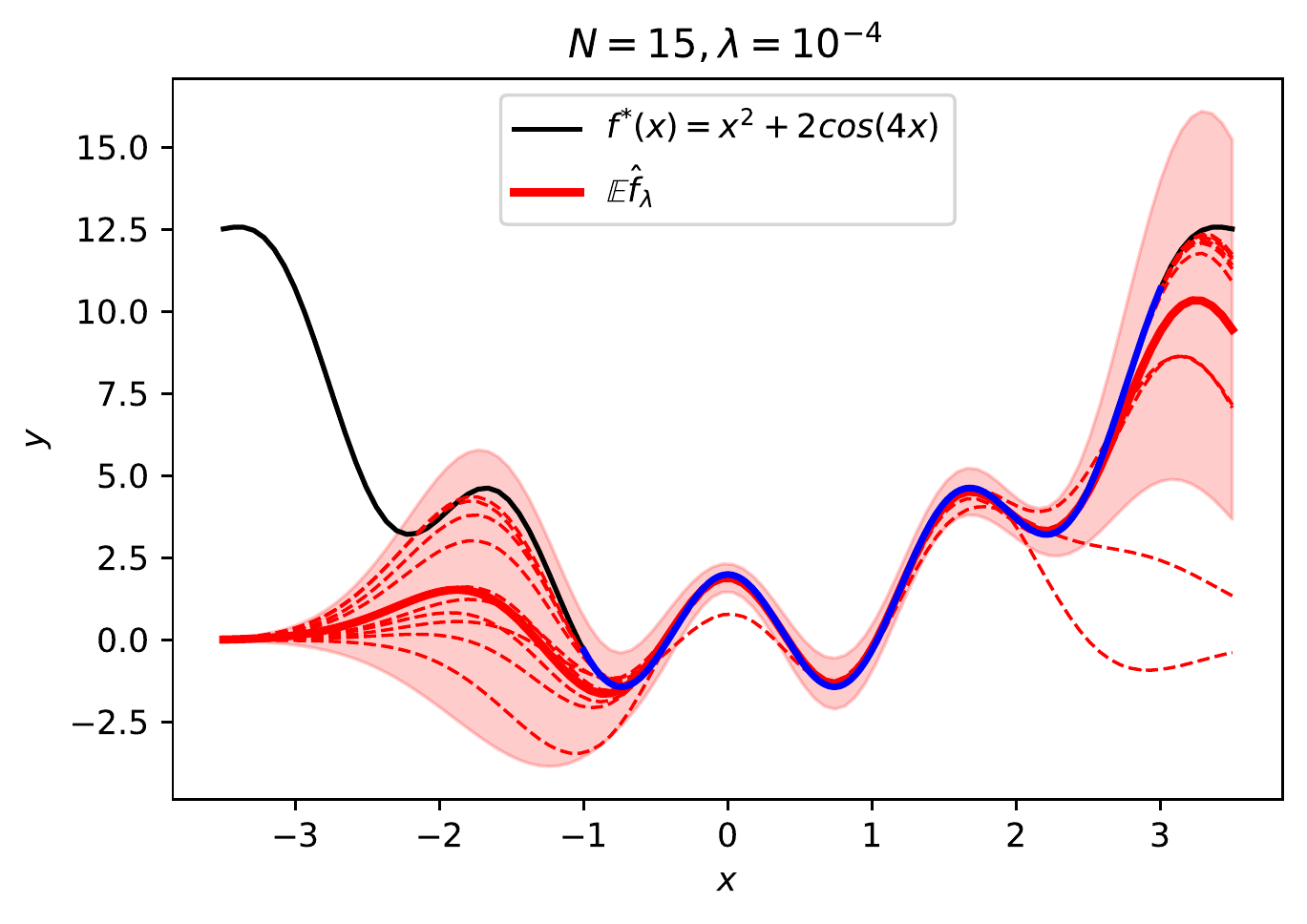}
        } \hfill
    {
          \includegraphics[width=0.23\textwidth]{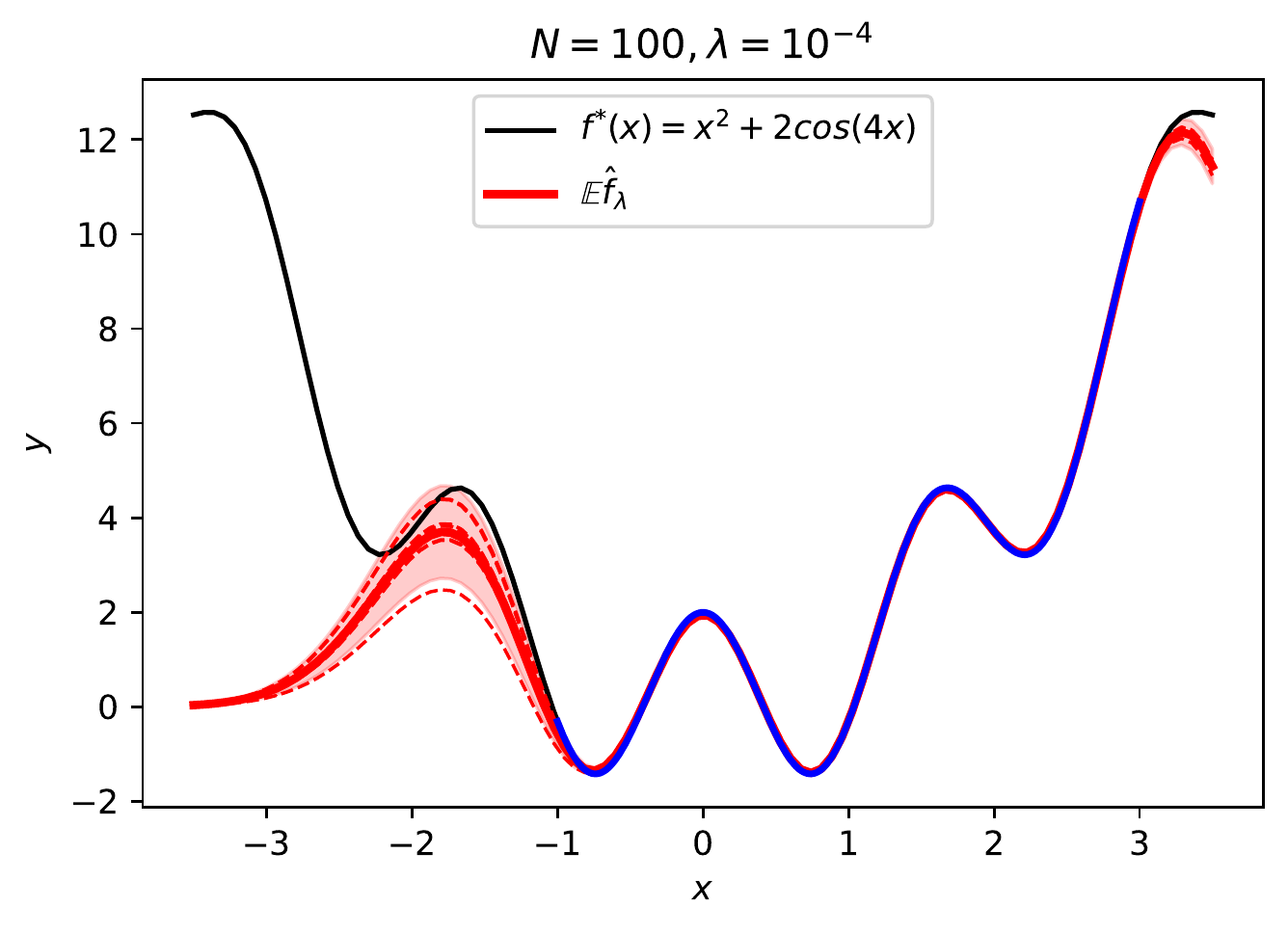}
        } \vfill
		{
        \includegraphics[width=0.23\textwidth]{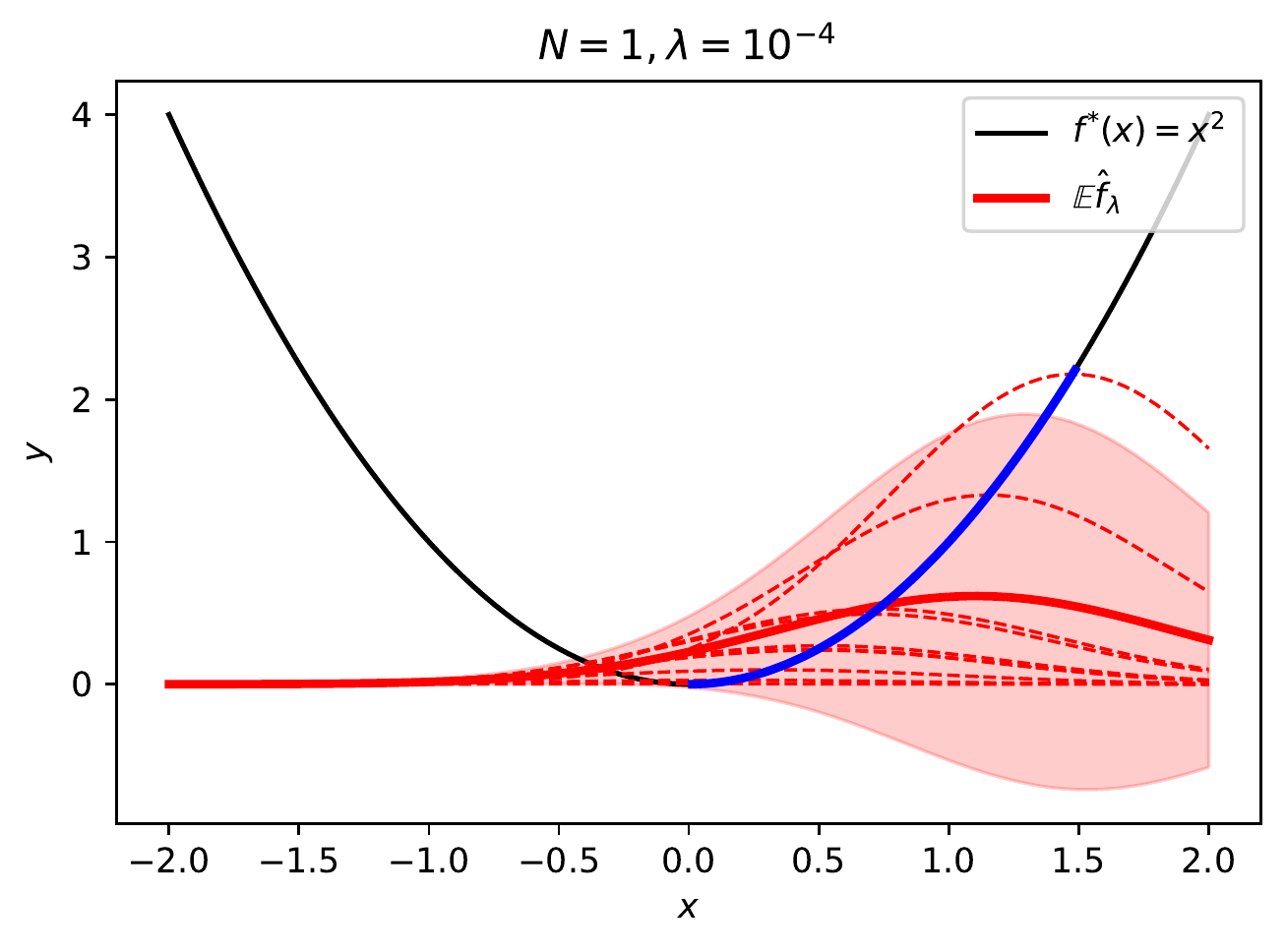}
        } \hfill
    {
        \includegraphics[width=0.23\textwidth]{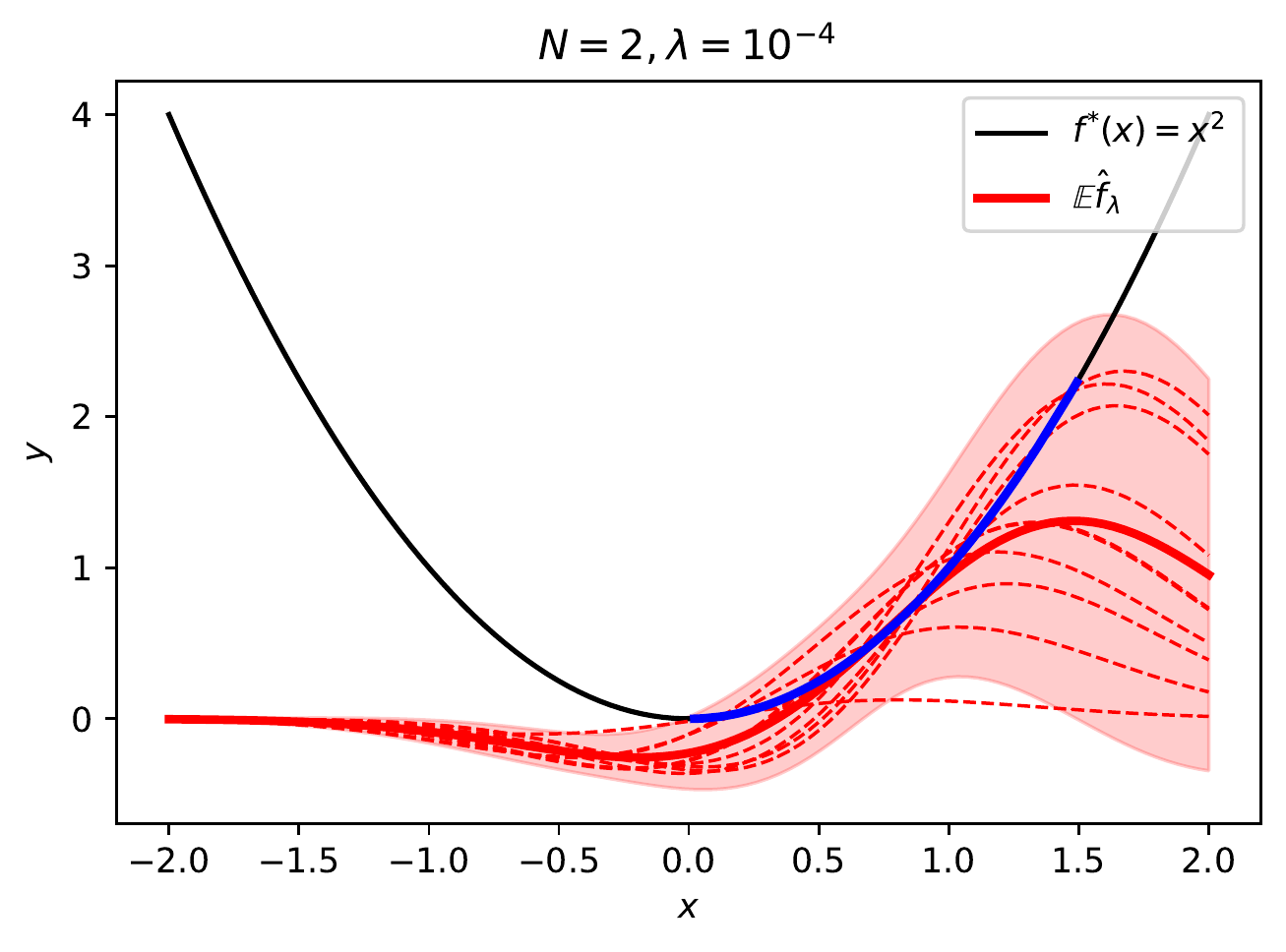}
        } \hfill
    {
        \includegraphics[width=0.23\textwidth]{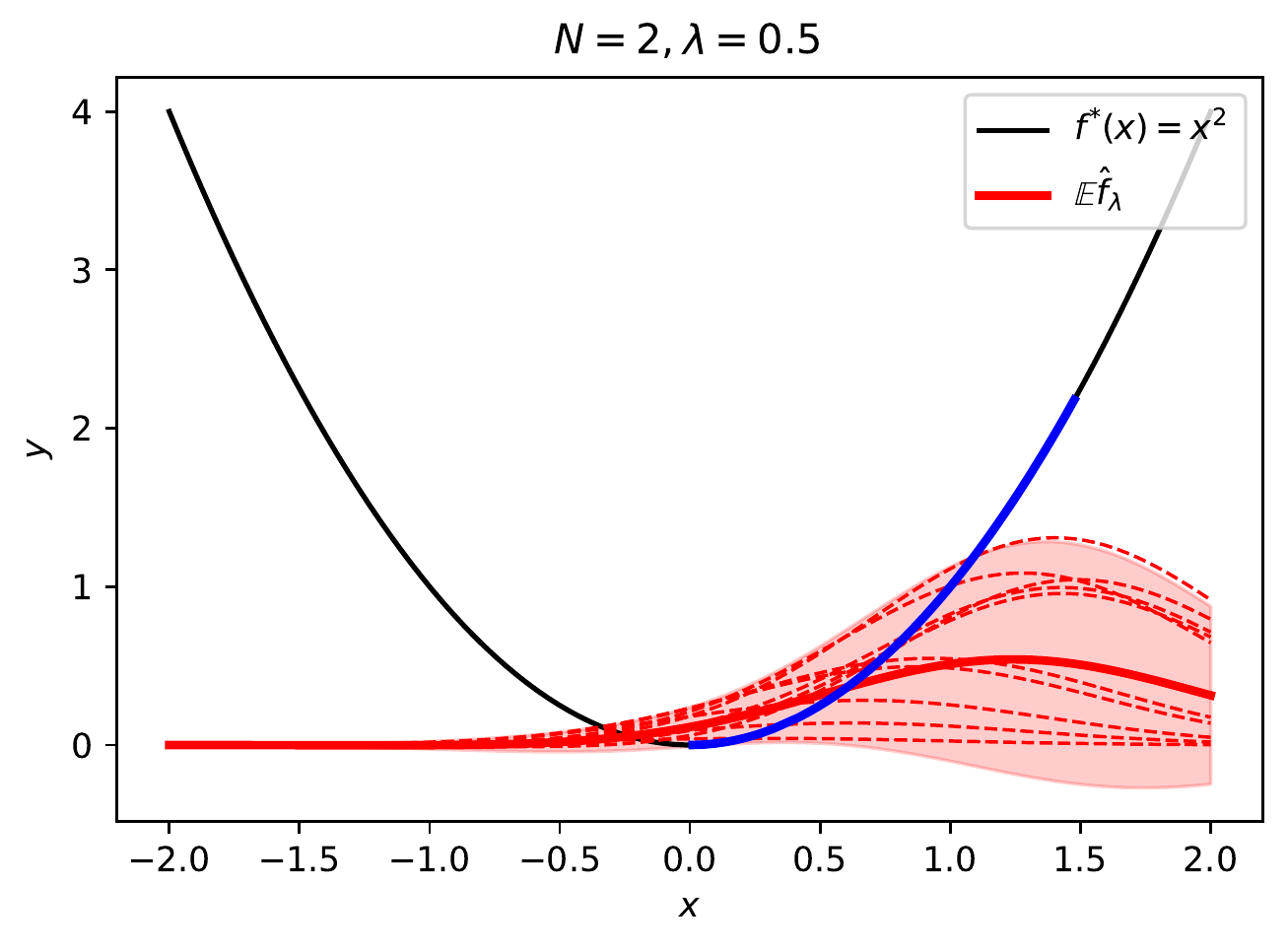}
        } \hfill
    {
         \includegraphics[width=0.23\textwidth]{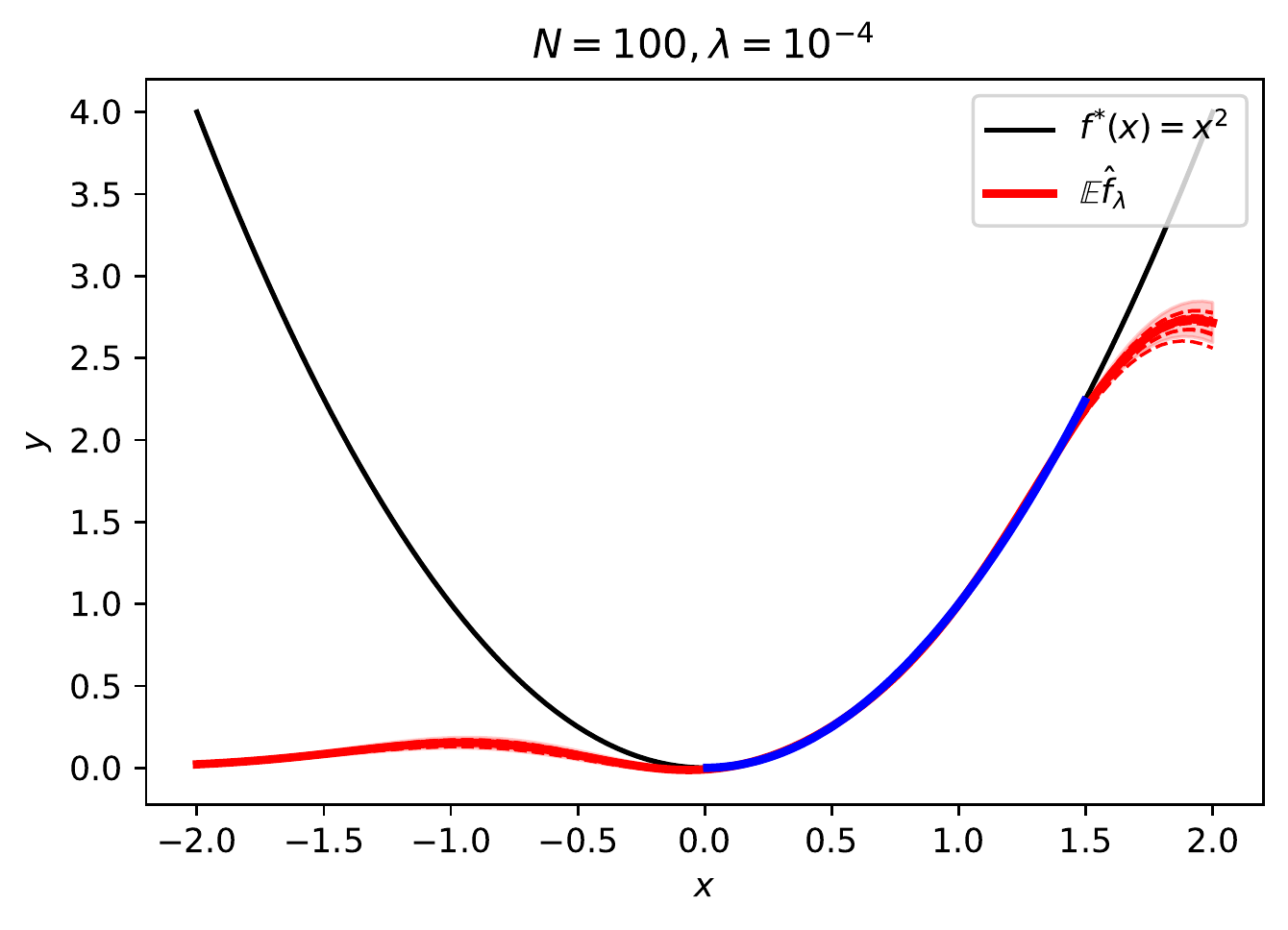}
        }
    \caption{\textit{KRR predictor in function space for various $N$ and $\lambda$ for the RBF Kernel $K$ with $\ell=d=1$.} Observations $o=\delta_x$ are sampled with uniform distribution on $x \sim U[-1, 3]$ (shown in blue)
    $\krrf$ is calculated $500$ times for different realizations of the training data ($10$ example predictors are shown in dashed lines), its mean and $\pm2$ standard deviation are shown in red. The true function $f^{*}(x) = x^2 + 2 \cos(4x)$ is shown in black.
		\textit{Second row.} Observations $o=\delta_x$ are sampled with uniform distribution $x \sim U [0, 1.5]$ (shown in blue) and $\krrf$ is calculated $100$ times.
		The true function $f^{*}(x) = x^2$ is shown in black.}
\end{figure}

\clearpage

\subsection{KARE predicts risk in average for small $N$}

\begin{figure}[!h]
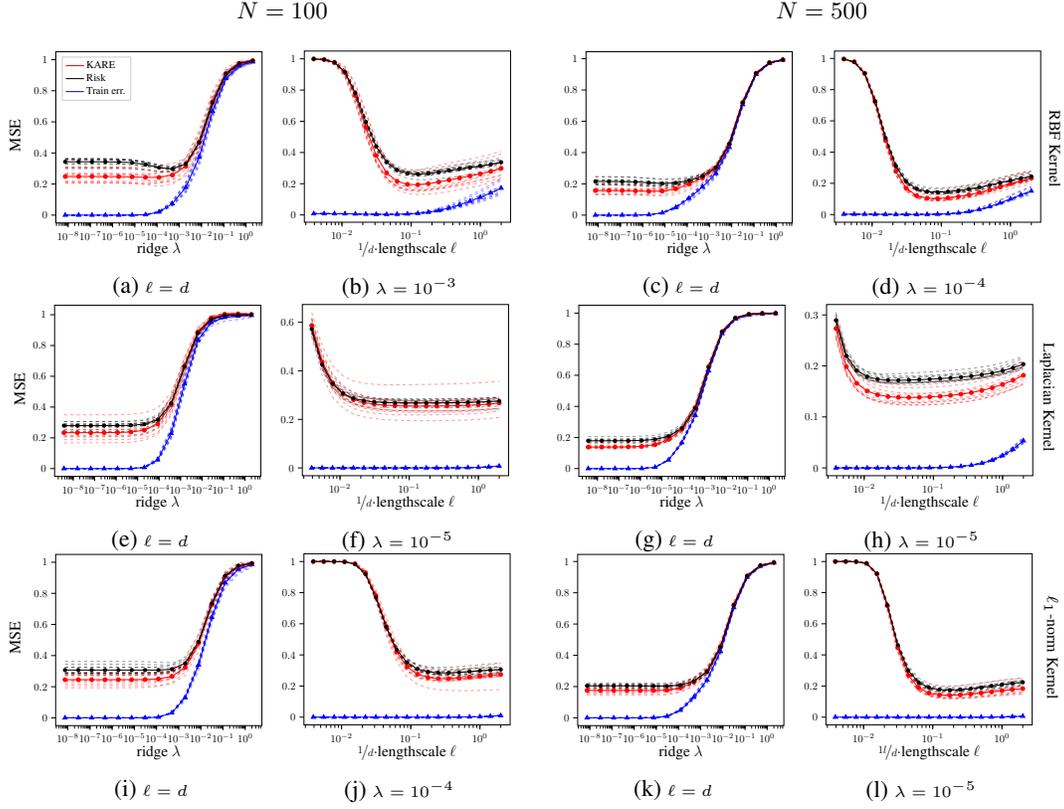

	{\small \textcolor{white}{h} \hspace{2.7cm} $N=100$ \hspace{5.8cm} $N=500$ \hspace{2cm}} \vspace{0.4cm} \\
	\rotatebox[origin=l]{90}{\tiny{\textcolor{white}{hey!}MSE}}
	\begin{subfigure}{.23\textwidth}
	\input{MNIST-approx-lambd-N100-RBF.tex}
	\vspace*{-0.4cm}
	\caption{{\tiny $\ell = d$}}
	\end{subfigure}
	\begin{subfigure}{.23\textwidth}
	\input{MNIST-approx-len-N100-RBF.tex}
	\vspace*{-0.4cm}
	\caption{{\tiny $\lambda = 10^{-3}$}}
	\end{subfigure}
	\hspace*{0.35cm}
	\begin{subfigure}{.23\textwidth}
	\input{MNIST-approx-lambd-N500-RBF.tex}
	\vspace*{-0.4cm}
	\caption{{\tiny $\ell = d$}}
	\end{subfigure}
	\begin{subfigure}{.23\textwidth}
	\input{MNIST-approx-len-N500-RBF.tex}
	\vspace*{-0.4cm}
	\caption{{\tiny $\lambda = 10^{-4}$}}
	\end{subfigure}
	\hspace*{-0.2cm}
	\rotatebox[origin=l]{270}{\tiny{\hspace*{-0.9cm}RBF Kernel}}
	\bigskip
	\vspace{1.5cm}
	\hspace*{-0.1cm}
	\rotatebox[origin=l]{90}{\tiny{\textcolor{white}{hey!}MSE}}
	\hspace*{-0.15cm}
	\begin{subfigure}{.23\textwidth}
	\input{MNIST-approx-lambd-N100-lap.tex}
	\vspace*{-0.4cm}
	\caption{{\tiny $\ell = d$}}
	\end{subfigure}
	\begin{subfigure}{.23\textwidth}
	\input{MNIST-approx-len-N100-lap.tex}
	\vspace*{-0.4cm}
	\caption{{\tiny $\lambda = 10^{-5}$}}
	\end{subfigure}
	\hspace*{0.3cm}
	\begin{subfigure}{.23\textwidth}
	\input{MNIST-approx-lambd-N500-lap.tex}
	\vspace*{-0.4cm}
	\caption{{\tiny $\ell = d$}}
	\end{subfigure}
	\begin{subfigure}{.23\textwidth}
	\input{MNIST-approx-len-N500-lap.tex}
	\vspace*{-0.4cm}
	\caption{{\tiny $\lambda = 10^{-5}$}}
	\end{subfigure}
	\hspace*{-0.2cm}
	\rotatebox[origin=l]{270}{\tiny{\hspace*{-1.2cm}Laplacian Kernel}}
	\bigskip
	\vspace{-2.4cm}
	\rotatebox[origin=l]{90}{\tiny{\textcolor{white}{hey!}MSE}}
	\begin{subfigure}{.23\textwidth}
	\input{MNIST-approx-lambd-N100-one-ker.tex}
	\vspace*{-0.4cm}
	\caption{{\tiny $\ell = d$}}
	\end{subfigure}
	\begin{subfigure}{.23\textwidth}
	\input{MNIST-approx-len-N100-one-ker.tex}
	\vspace*{-0.4cm}
	\caption{{\tiny $\lambda = 10^{-4}$}}
	\end{subfigure}
	\hspace*{0.25cm}
	\begin{subfigure}{.23\textwidth}
	\input{MNIST-approx-lambd-N500-one-ker.tex}
	\vspace*{-0.4cm}
	\caption{{\tiny $\ell = d$}}
	\end{subfigure}
	\begin{subfigure}{.23\textwidth}
	\input{MNIST-approx-len-N500-one-ker.tex}
	\vspace*{-0.4cm}
	\caption{{\tiny $\lambda = 10^{-5}$}}
	\end{subfigure}
	\hspace*{-0.15cm}
	\rotatebox[origin=l]{270}{\tiny{\hspace*{-1.15cm}$\ell_1$-norm Kernel}}
  \caption{\textit{The estimation predicts the risk in average for small $N=\{100,500\}$ on MNIST data.} In the top row, we used the RBF Kernel $K(x,z) = \exp(\nicefrac{-\|x-z\|_2^2}{\ell})$, in the second row, we used the Laplacian Kernel $K(x,z) = \exp(\nicefrac{-\|x-z\|_2}{\ell})$, and in the bottom row, we used the $\ell_1$-norm Kernel $K(x,z) = \exp(\nicefrac{-\|x-z\|_1}{\ell})$ for various choices of $\ell$ and $\lambda$.
  The optimal predictor is calculated using $N$ random samples ($N=100$ for the plots on the left and $N=500$ for the ones on the right) from the training data $10$ times (dashed curves) and their average is plotted in the solid curves.}
\end{figure}

\clearpage

\subsection{SCT and its behavior}
In general, it is hard to compute the spectrum $(d_k)_{k \in \mathbb{N}}$ of $T_K$ even when one has the knowledge of the true data distribution.
Luckily, following an adaptation from \cite{williams2000effect, fasshauer2012stable}, we can obtain an explicit formula for $d_k$ for centered $d$-dimensional Gaussian distribution with covariance matrix $\sigma^2 I_d$, and RBF Kernel $K(x, x') = \exp(-\nicefrac{\|x-x'\|^2}{\ell})$. The formula for the distinct eigenvalues $\lambda_k$ is
  \begin{align}\label{eq:eigen-formula-num}
    \ \lambda_k = \left(\sqrt{\frac{1}{2A \sigma^2}}\right)^{d} B^k,
  \end{align}
where $A = \frac{1}{4 \sigma^2} + \frac{1}{\ell} + c$, $B = \frac{1}{A \ell}$ with $c = \frac{1}{2\sigma}\sqrt{\frac{1}{4 \sigma^2} + \frac{2}{\ell}}$. Each $\lambda_k$ has multiplicity
\begin{align}
	 \ n_d(k) = \sum_{j=1}^k {d \choose j} {k-1 \choose j-1}
\end{align}
for $k \geq 1$. In particular, we have $n_d(0) = 1, n_d(1) = {d \choose 1}, n_d(2) = {d \choose 2} + d, \ldots$. 
In general, $n_d(k)$ is the number of ways to partition $k$ into $d$ non-negative integers.

The true SCT is therefore approximated solving the following equation numerically
\begin{align}\label{eq:SCT-approx-formula-num}
		\ \vartheta = \lambda + \frac{\vartheta}{N} \sum_{i=1}^{k} \frac{n_d(k) \lambda_k}{\lambda_k + \vartheta}.
\end{align}

\begin{figure}[!h]
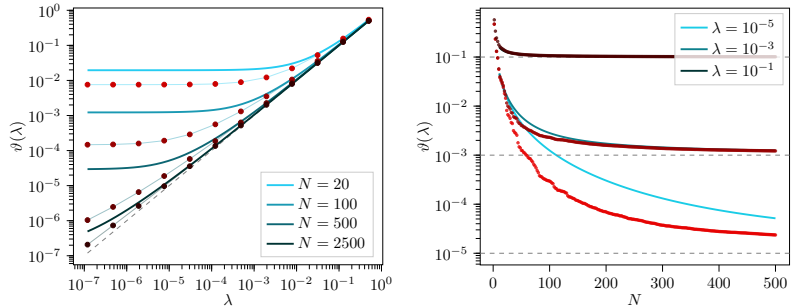

	\centering
  \input{SCT-d5.tex} \input{SCT-d5-N.tex} 
  \caption{\textit{Behavior of SCT as a function of $\lambda$ and $N$.} True SCT is calculated on the $k=50$ biggest distinct eigenvalues using the formula \ref{eq:SCT-approx-formula-num} for $\ell = d = 5$ and $\sigma = 1$. Red dots are the approximations obtained using Proposition 5 in the main text, i.e. $\sct\approx 1 / \Tr[(\frac{1}{N}K(X,X) -\lambda I)^{-1}]$.}
\end{figure}

Note that in the Figure 2 in the main text, we limit the approximation to $k=10$ for $d=20$ because the multiplicity $n_d(k)$ grows polynomially with $d^k$.

\clearpage

\section{Proofs}\label{sec:proofs}

\subsection*{Preliminary: Big-P notation}

Throughout our proofs, we will frequently rely on a polynomial analogue of the big-O notation, which we call big-P:
\begin{definition} \label{def:big-p}
    For two functions $ f $ and $ g $ (of one or several variables, defined on an arbitrary common domain $ \mathcal D $),
    we write $ f = \mathcal{P} (g) $ if $ g $ is nonnegative over $ \mathcal D $ and there exists a polynomial $ \mathbf P$
    with nonnegative coefficients and $ \mathbf P (0) = 0 $ such that $ | f | \leq \mathbf P (g) $ over $ \mathcal D $.
\end{definition}
Note that the big-O notation corresponds to the case when the polynomial $ \mathbf P $ is of degree at most one.

\subsection{Objects of Interest and general strategy}

The central object of our analysis is the $N\times N$ Gram matrix
$\mathcal{O}K\mathcal{O}^{T}$, in particular the related Stieltjes
transform:
\[
m(z)=\frac{1}{N}\mathrm{Tr}\left[B(z)^{-1}\right]
\]
where $B(z)=\frac{1}{N}\mathcal{O}K\mathcal{O}^{T}-zI_{N}$ and $z\in\mathbb{C}\setminus\mathbb{R}_{+}$.

From now on, we consider only $z\in\mathbb{H}_{<0}=\left\{z : \Re (z) < 0 \right\}$. Note that $m(z)=\frac{1}{N}\sum_{\ell}\frac{1}{\lambda_{\ell}-z}$ where $\lambda_{\ell}\geq 0$ are the real eigenvalues of $\frac{1}{N}\mathcal{O}K\mathcal{O}^{T}$, hence $m(z)$ lies in the cone $\Gamma$ spanned by $1$ and $-\nicefrac{1}{z}$, i.e.   $\Gamma = \{a - b \frac 1 z | a,b \geq 0 \}$.
We will first show that for $z\in\mathbb{H}_{<0}$, the Stieltjes transform concentrates around the unique solution $\tilde{m}(z)$ to the equation
\begin{equation}
\tilde{m}(z)=-\frac{1}{z}\left(1-\frac{1}{N}\mathrm{Tr}\left[\tilde{m}(z)T_{K}\left(I_{\mathcal{C}}+\tilde{m}(z)T_{K}\right)^{-1}\right]\right),\label{eq:tilde_m}
\end{equation}
and then show that the linear map
\[
A(z)=\frac{1}{N}K\mathcal{O}^{T}\left(\frac{1}{N}\mathcal{O}K\mathcal{O}^{T}-zI_{N}\right)^{-1}\mathcal{O}=\frac{1}{N}K\mathcal{O}^{T}B(z)^{-1}\mathcal{O}
\]
concentrates around the map $\tilde{A}_{\vartheta(-z)} = T_{K}\left(T_{K}+\vartheta(-z)I_{\mathcal{C}}\right)^{-1}$, where $(T_{K}f)(x)=\mathbb{E}_{o\sim\pi}\left[o(K(x,\cdot))o(f)\right]=\left\langle K(x,\cdot),f\right\rangle _{S}$ and $\vartheta(-z)=\frac{1}{\tilde{m}(z)}$ is the Signal Capture Threshold.
From Equation (\ref{eq:tilde_m}), the SCT can be also defined as the solution to the equation
\begin{equation}
\vartheta(-z)=-z+\frac{\vartheta(-z)}{N}\mathrm{Tr}\left[T_{K}\left(T_{K}+\vartheta(-z)I_{\mathcal{C}}\right)^{-1}\right].\label{eq:t_lambda}
\end{equation}
From now on, we denote $\vartheta(-z)$ by $\vartheta$. Note that here, in the Appendix, we use the resolvent notation: in particular the KRR reconstruction operator $A_\lambda$ is equal to $A(-\lambda)$.

\subsubsection{Spectral decomposition and generalized matrix representation}
Throughout this paper it is assumed that there exists an orthonormal basis of continuous functions $\left(f^{(k)}\right)_{k}$ for the scalar product $\bra \cdot, \cdot \ket_{S}$ such that $K=\sum_{k\in\mathbb{N}}d_{k}f^{(k)}\otimes f^{(k)}$ and $\sum_{k\in\mathbb{N}}d_{k}<\infty$. For a linear map $M:\mathcal{C}\to\mathcal{C}$, we define the $(k,\ell)-$entry of $M$ as:
\[
M_{k\ell}=\left\langle f^{(k)},Mf^{(\ell)}\right\rangle _{S}.
\]
With this notation, the trace of a linear map $M$ becomes $\mathrm{Tr}\left(M\right)=\sum_{k\in\mathbb{N}}M_{kk}$.

Similarly, using the canonical basis $(b_{i})_{i=1,\ldots,N}$ of $\mathbb{R}^{N}$, we define the entries of $\mathcal{O}:\mathcal{C}\to\mathbb{R}^{N}$ and $\mathcal{O}^{T}:\mathbb{R}^{N}\to\mathcal{C}^{*}$ by
\[
\mathcal{O}_{ik}=b_{i}\cdot\mathcal{O}f^{(k)}=o_{i}(f^{(k)}),\quad\mathcal{O}_{ik}^{T}=\mathcal{O}^{T}b_{k}(f^{(i)})=b_{k}\cdot\mathcal{O}f^{(i)}=o_{k}(f^{(i)}).
\]
Since the observations $o_{i}$ are i.i.d. Gaussians with zero mean and covariance $\mathbb{E}\left[o_{i}(f)o_{i}(g)\right]=\left\langle f,g\right\rangle _{S}$ and since $\left(f^{(k)}\right)_k$ is an orthonormal basis for the scalar product $\left\langle \cdot,\cdot\right\rangle _{S}$, the entries $\mathcal{O}_{ik}$ are i.i.d standard Gaussians.

Using the spectral decomposition of $K$, the entries of $\mathcal{O}K\mathcal{O}^{T}$ are given by:
\[
\left(\mathcal{O}K\mathcal{O}^{T}\right)_{i,j}=\sum_{\ell}d_{\ell}o_{i}(f^{(\ell)})o_{j}(f^{(\ell)}),
\]
where the sum converges absolutely (thanks to the trace assumption on $K$) and the entries of $A$ are then given by:
\begin{equation}
A_{k\ell}(z)=\frac{d_{k}}{N}\left(\mathcal{O}_{\cdot k}\right)^{T}\left(\frac{1}{N}\mathcal{O}K\mathcal{O}^{T}-zI_{N}\right)^{-1}\mathcal{O}_{\cdot \ell}\label{eq:def_A}
\end{equation}
 where $\mathcal{O}_{\cdot k}=\left(o_{i}(f^{(k)})\right)_{i=1,\ldots,N}$.

\subsubsection{Shermann-Morrison Formula \label{subsec:Shermann-Morrison-Formula}}
The Shermann-Morrison formula allows one to study how the inverse of a matrix is modified by a rank one perturbation of the matrix. The matrix $\mathcal{O}K\mathcal{O}^{T}$
can be seen as a perturbation of $\mathcal{O}K_{(k)}\mathcal{O}^{T}$ by the rank one matrix $d_{k}\mathcal{O}_{\cdot k}\mathcal{O}_{\cdot k}^{T}$, where $K_{(k)}:=\sum_{\ell\neq k}d_{\ell}f^{(\ell)}\otimes f^{(\ell)}$. By doing so, one isolates the contribution of the $k$-th eigenvalue of $K$.
Thus, one can compute $B(z)^{-1}=\left(\frac{1}{N}\mathcal{O}K\mathcal{O}^{T}-zI_{N}\right)^{-1}$ using the Shermann-Morrison formula:
\begin{equation}
B(z)^{-1}=B_{(k)}(z)^{-1}-\frac{1}{N}\frac{d_{k}}{1+d_{k}g_{k}(z)}B_{(k)}(z)^{-1}\mathcal{O}_{\cdot k}\mathcal{O}_{\cdot k}^{T}B_{(k)}(z)^{-1}\label{eq:Shermann}
\end{equation}
where $B_{(k)}(z)=\frac{1}{N}\mathcal{O}K_{(k)}\mathcal{O}^{T}-zI_{N}$ and $g_{k}(z)=\frac{1}{N}\mathcal{O}_{\cdot k}^{T}B_{(k)}(z)^{-1}\mathcal{O}_{\cdot k}.$ A crucial property is that, since $o_{i}\left(f^{(k)}\right)$ does not appear anymore in $\mathcal{O}K_{(k)}\mathcal{O}^{T}$ and, since for any $\ell\neq k$ and any $i,j$,  we have that $o_{i}\left(f^{(k)}\right)$ is independent from $o_{j}(f^{(\ell)})$, we obtain that the matrix $B_{(k)}(z)^{-1}$ is independent of $\mathcal{O}_{\cdot k}$.
\begin{rem}\label{rem:lying-in-the-cone}
Using the diagonalization of $B_{(k)}(z)^{-1}=U^{T}\mathrm{diag}\left(\frac{1}{\nu_{\ell}-z}\right)U$ with $U$ orthogonal and $\nu_{\ell}\geq0$, we have that $g_{k}(z)=\frac{1}{N}\sum_{\ell}\frac{\left[\sum_{i}U_{\ell,i}o_{i}(f^{(k)})\right]^{2}}{\nu_{\ell}-z}$ lies in the cone spanned by $1$ and $-\nicefrac{1}{z}$, in particular, $\Re(g_k)\geq 0$ on $\mathbb H _{<0}$.
\end{rem}
As a result of Equations (\ref{eq:def_A}) and (\ref{eq:Shermann}), the diagonal entries of the operator $A(z)=\frac{1}{N}K\mathcal{O}^{T}B(z)^{-1}\mathcal{O}$ are equal to
\begin{equation}\label{eq:diagonal_term}
A_{kk}(z)=\frac{d_{k}g_{k}(z)}{1+d_{k}g_{k}(z)}.
\end{equation}

\begin{rem}
For any $z\in\mathbb{H}_{<0},$ the sum $\sum_{k}\left|A_{kk}(z)\right|$ is almost surely finite. Indeed, notice that
\[
\left|\frac{d_{k}g_{k}(z)}{1+d_{k}g_{k}(z)}\right|\leq\left|d_{k}g_{k}(z)\right|\leq\frac{1}{N}d_{k}\left\Vert \mathcal{O}_{\cdot k}\right\Vert ^{2}\left\Vert B_{(k)}(z)^{-1}\right\Vert _{\mathrm{op}}.
\]
For any $z\in\mathbb{H}_{<0}$, $\left\Vert B_{(k)}(z)^{-1}\right\Vert_{\mathrm{op}} \leq\frac{1}{\left|z\right|}$
and thus
\[
\left|\frac{d_{k}g_{k}(z)}{1+d_{k}g_{k}(z)}\right|\leq\frac{1}{N\left|z\right|}d_{k}\left\Vert \mathcal{O}_{\cdot k}\right\Vert ^{2}.
\]
 Since $\mathbb{E}\left[\sum_{k}d_{k}\left\Vert \mathcal{O}_{\cdot k}\right\Vert ^{2}\right]=N\Tr[T_K]<\infty$, we have that
$\sum_{k}\left|A_{kk}(z)\right|$ is almost surely finite.

The operator $A$ is therefore a.s. trace-class and $\mathrm{Tr}(A)=\sum_{k}\frac{d_{k}g_{k}(z)}{1+d_{k}g_{k}(z)}$,
where the sum is absolutely convergent.
\end{rem}
Another important observation is that the Stieltjes transform $m(z)$ and the $g_{k}(z)$ are closely related.
\begin{lem}
For any $z\in\mathbb{H}_{<0}$, a.s. we have
\begin{equation}
m(z)=-\frac{1}{z}\left(1-\frac{1}{N}\sum_{k=1}^{\infty}\frac{d_{k}g_{k}(z)}{1+d_{k}g_{k}(z)}\right).\label{eq:relation_mP_gk_fixpoint}
\end{equation}
\end{lem}
\begin{proof}
Indeed, using the trivial relation $\mathrm{Tr}\left[B(z)B(z)^{-1}\right]=N$, expanding $B(z)$, we obtain $\mathrm{Tr}\left[\frac{1}{N}\mathcal{O}K\mathcal{O}^{T}B(z)^{-1}\right]-z\mathrm{Tr}\left[B(z)^{-1}\right]=N$.
Since  $\mathcal O $ is an operator from $ \mathcal C $ to $ \mathbb{R}^N$, which is a finite dimensional space, we can apply the cyclic property of the trace and obtain $\mathrm{Tr}\left[\frac{1}{N}\mathcal{O}K\mathcal{O}^{T}B(z)^{-1}\right]=\mathrm{Tr}\left[A(z)\right]$.
Thus, $$\mathrm{Tr}\left[A(z)\right]-z\mathrm{Tr}\left[B(z)^{-1}\right]=N.$$
Dividing both sides by $N$ and using Equation (\ref{eq:diagonal_term}), we obtain
\[
1=\frac{1}{N}\sum_{k=1}^{\infty}\frac{d_{k}g_{k}(z)}{1+d_{k}g_{k}(z)}-zm(z),
\]
hence the result.
\end{proof}

\subsection{Concentration of the Stieltjes Transform\label{subsec:Appendix-Concentration-Stieltjes}}

We will now show that $g_{k}(z)=\frac{1}{N}\mathcal{O}_{\cdot k}^{T}B_{(k)}(z)^{-1}\mathcal{O}_{\cdot k}$ is close to $\frac{1}{N}\mathrm{Tr}\left(B_{(k)}(z)^{-1}\right)$, as suggested by the fact that by Wick's formula $\mathbb E [g_k]=\frac{1}{N}\mathrm{Tr}\left(\mathbb{E}\left[B_{(k)}(z)^{-1}\right]\right)$.
Since $B(z)$ is obtained using a rank one permutation of $B_{(k)}(z)$, $\frac{1}{N}\mathrm{Tr}\left(B_{(k)}(z)^{-1}\right)$ is close to the Stieltjes transform $m$. As a result, all the $g_{k}$'s are close to the Stieltjes transform $m$: it is natural to think that for $z\in\mathbb{H}_{<0}$, both $g_{k}(z)$'s and $m(z)$ should concentrate around the unique solution $\tilde{m}(z)$ in the cone spanned by $1$ and $-\nicefrac{1}{z}$ of the equation
\begin{equation}
\tilde{m}(z)=-\frac{1}{z}\left(1-\frac{1}{N}\sum_{k=1}^{\infty}\frac{d_{k}\tilde{m}(z)}{1+d_{k}\tilde{m}(z)}\right).\label{eq:fixpoint_def}
\end{equation}
\begin{rem} The existence and the uniqueness of the solution in the cone spanned by $1$ and $-\nicefrac{1}{z}$ of the equation can be argued as follows. If in Equation (\ref{eq:fixpoint_def}) we truncate the series and consider the sum of the first $M$ terms, one can show that there exists a unique fixed point $\tilde m_M(z)$ in the region $R$ given by intersection between the cone spanned by $1$ and $-\nicefrac{1}{z}$ and the cone spanned by $z$ and $\nicefrac{1}{z}$ translated by $+{} 1$ and multiplied by $-\nicefrac{1}{z}$ (see Lemma C.6 in the Supplementary Material of \cite{jacot-2020}). Since $R$ is a compact region, we can extract a converging subsequence that solves Equation (\ref{eq:fixpoint_def}), the limit of which can be showed to be unique, again using the same arguments of Lemma C.6 in the Supplementary Material of \cite{jacot-2020}.
\end{rem}
From now on we omit the $z$ dependence and we set $m=m(z)$, $\tilde m=\tilde m(z)$ and $g_k(z)=g_k$.

\subsubsection{Concentration bounds}
Using Equation \ref{eq:relation_mP_gk_fixpoint} and the definition of the fixed point $\tilde{m}$ (Equation \ref{eq:fixpoint_def}), we obtain the following formula for the difference between the Stieltjes transform $m$ and $\tilde{m}$:
\begin{align*}
\tilde{m}-m & =\frac{1}{z}\frac{1}{N}\sum_{k=1}^{\infty}\frac{d_{k}\left(\tilde{m}-g_{k}\right)}{(1+d_{k}\tilde{m})(1+d_{k}g_{k})}\\
 & =\frac{\tilde{m}-m}{z}\frac{1}{N}\sum_{k=1}^{\infty}\frac{d_{k}}{(1+d_{k}\tilde{m})(1+d_{k}g_{k})}+\frac{1}{z}\frac{1}{N}\sum_{k=1}^{\infty}\frac{d_{k}\left(m-g_{k}\right)}{(1+d_{k}\tilde{m})(1+d_{k}g_{k})},
\end{align*}
where the well-posedness of the two infinite sums of the r.h.s is granted by the fact that:
\begin{enumerate}[leftmargin=*]
\item $\left|\frac{d_{k}}{(1+d_{k}\tilde{m})(1+d_{k}g_{k})}\right|\leq d_{k}$ since $\Re\left(\tilde{m}\right),\Re(g_{k})$ are positive, thus the first sum is absolutely convergent,
\item being the difference of two absolutely convergent series, the second sum is also absolutely convergent.
\end{enumerate}
As a consequence, the difference $\tilde{m}-m$ can be expressed as
\begin{equation}
\tilde{m}-m=\frac{\frac{1}{N}\sum_{k=1}^{\infty}\frac{d_{k}\left(m-g_{k}\right)}{(1+d_{k}\tilde{m})(1+d_{k}g_{k})}}{z-\frac{1}{N}\sum_{k=1}^{\infty}\frac{d_{k}}{(1+d_{k}\tilde{m})(1+d_{k}g_{k})}},\label{eq:error_mN_mtilde}
\end{equation}
which allows us to show the concentration of $m$ around $\tilde{m}$ from the concentration of $g_{k}$ around $m$.

Regarding the concentration of the $g_{k}$'s around $m$, we have the following result:
\begin{lem}
\label{lem:loose_bound_mN_gk} For any $N,s\in\mathbb{N}$ and any $z\in\mathbb{H}_{<0}$, we have
\begin{eqnarray*}
\mathbb{E}\left[\left|m-g_{k}\right|^{2s}\right] & \leq & \frac{\boldsymbol{c}_{s}}{\left|z\right|^{2s}N^{s}},\\
\mathbb{E}\left[\left|m-m_{(k)}\right|^{2s}\right] & \leq & \frac{1}{\left|z\right|^{2s}N^{2s}}.
\end{eqnarray*}
where $\boldsymbol{c}_{s}$ only depends on $s$.
\end{lem}
\begin{proof}
The second inequality will be proven while proving the first one.
Let $m_{(k)}=\frac{1}{N}\mathrm{Tr}\left[B_{(k)}^{-1}\right]$ where $B_{(k)}$ was defined in Section \ref{subsec:Shermann-Morrison-Formula}.
By convexity:
\begin{align}
\mathbb{E}\left[\left|m-g_{k}\right|^{2s}\right] & \leq2^{2s-1}\mathbb{E}\left[\left|m-m_{(k)}\right|^{2s}\right]+2^{2s-1}\mathbb{E}\left[\left|m_{(k)}-g_{k}\right|^{2s}\right].\label{eq:convex_in}
\end{align}
\textbf{Bound on $\mathbb{E}[|m-m_{(k)}|^{2s}]$:}
We obtain the bound on the expectation by showing that a deterministic bound holds for the random variable $|m-m_{(k)}|^{2s}$. Using the Sherman-Morrison formula (Equation (\ref{eq:Shermann})), and using the cyclic property of the trace,
\[
m=m_{(k)}-\frac{1}{N}\frac{d_{k}g'_{k}}{1+d_{k}g_{k}}
\]
since the derivative $g'_{k}(z)$ of $g_{k}(z)$ is equal to $\frac{1}{N}\mathcal{O}_{\cdot k}^{T}B(z)^{-2}\mathcal{O}_{\cdot k}$.
As a result, we obtain $\left|m-m_{(k)}\right|^{2s}=\frac{1}{N^{2s}}\frac{d_{k}^{2s}\left|g'_{k}\right|^{2s}}{\left|1+d_{k}g_{k}\right|^{2s}}$.
Using the fact that $\left|1+d_{k}g_{k}\right|\geq\left|d_{k}g_{k}\right|$
since $\Re\left(g_{k}\right)\geq0$,
\[
\left|m-m_{(k)}\right|^{2s}\leq\frac{1}{N^{2s}}\frac{\left|g'_{k}\right|^{2s}}{\left|g_{k}\right|^{2s}}.
\]
Notice now that
\[
\left|\frac{g'_{k}}{g_{k}}\right|=\left|\frac{\mathcal{O}_{\cdot k}^{T}B_{(k)}(z)^{-2}\mathcal{O}_{\cdot k}}{\mathcal{O}_{\cdot k}^{T}B_{(k)}(z)^{-1}\mathcal{O}_{\cdot k}}\right|\leq\max_{w\in \mathbb R ^N}\left|\frac{w^T B_{(k)}(z)^{-2}w}{w^T B_{(k)}(z)^{-1}w}\right|\leq\left\Vert B_{(k)}(z)^{-1}\right\Vert _{\mathrm{op}}.
\]
The eigenvalues of $B_{(k)}(z)^{-1}$ are given by $\frac{1}{\lambda_{i}-z}$ where the $\lambda_{i}>0$ are the eigenvalues of the symmetric matrix $\frac{1}{N}\mathcal{O}K_{(k)}\mathcal{O}^{T}$: $\left\Vert B_{(k)}(z)^{-1}\right\Vert _{\mathrm{op}}\leq\max_{i}\frac{1}{\left|\lambda_{i}-z\right|}$ is also bounded by $\frac{1}{\left|z\right|}$ if $z\in\mathbb{H}_{<0}$. Thus we get
\[
\left|m-m_{(k)}\right|^{2s}\leq\frac{1}{\left|z\right|^{2s}N^{2s}}.
\]
\textbf{Bound on $\mathbb{E}[|m_{(k)}-g_{k}|^{2s}]$:}
The term $\mathbb{E}\left[\left(\left(m_{(k)}-g_{k}\right)\overline{\left(m_{(k)}-g_{k}\right)}\right)^{s}\right]$ is equal to
\[
\mathbb{E}\left[\left(\left(\frac{1}{N}\mathrm{Tr}\left[B_{(k)}^{-1}\right]-\frac{1}{N}\mathcal{O}_{\cdot k}^{T}B_{(k)}^{-1}\mathcal{O}_{\cdot k}\right)\left(\frac{1}{N}\mathrm{Tr}\left[\overline{B_{(k)}^{-1}}\right]-\frac{1}{N}\mathcal{O}_{\cdot k}^{T}\overline{B_{(k)}^{-1}}\mathcal{O}_{\cdot k}\right)\right)^{s}\right].
\]
Let $\boldsymbol{B}=\left(B_{(k)},\overline{B_{(k)}},\ldots,B_{(k)},\overline{B_{(k)}}\right)$ and let us denote by $\boldsymbol{B}(i)$ the $i^{th}$ element of $\boldsymbol{B}$. Using Wick's formula (Lemma \ref{lem:Wicks-formula-matrix}), we have
\[
  \mathbb{E}\left[\left|m_{(k)}-g_{k}\right|^{2s}\right]=\frac{1}{N^{s}}\sum_{\sigma\in\mathfrak{S}_{2s}^{\dagger}}\frac{1}{N^{s-\boldsymbol{c}(\sigma)}}2^{2s-\boldsymbol{c}(\sigma)}\mathbb{E}\left[\prod_{\stackrel{}{}c\text{ cycle of }\sigma}\frac{1}{N}\mathrm{Tr}\left[\prod_{i\in c}\boldsymbol{B}(i)\right]\right],
\]
where we recall that $\mathfrak{S}_{2s}^{\dagger}$ is the set of permutations with no fixed points and the product over $i$ is taken according to the order given by the cycle $c$ and does not depend on the starting point. Using the fact that the eigenvalues of $B_{(k)}$ are of the form $\nicefrac{1}{\left(\lambda_{i}-z\right)}$ with $\lambda_{i}\geq0$,
\[
\left|\frac{1}{N}\mathrm{Tr}\left[\prod_{i\in c}\boldsymbol{B}(i)\right]\right|\leq\frac{1}{\left|z\right|^{\#c}}.
\]
Hence,
\[
\mathbb{E}\left[\left|m_{(k)}-g_{k}\right|^{2s}\right]\leq\frac{1}{N^{s}}\frac{1}{\left|z\right|^{2s}}\sum_{\sigma\in\mathfrak{S}_{2s}^{\dagger}}\frac{2^{2s-\boldsymbol{c}(\sigma)}}{N^{s-\boldsymbol{c}(\sigma)}}.
\]
Note that, since $\sigma\in\mathfrak{S}_{2s}^{\dagger},$ it has no
fixed point, hence $\boldsymbol{c}(\sigma)\leq s$ and thus $K_{s}:=\sup_{N}\sum_{\sigma\in\mathfrak{S}_{2s}^{\dagger}}\frac{2^{2s-\boldsymbol{c}(\sigma)}}{N^{s-\boldsymbol{c}(\sigma)}}$ is finite. This yields the inequality
\[
\mathbb{E}\left[\left|m_{(k)}-g_{k}\right|^{2s}\right]\leq\frac{K_{s}}{\left|z\right|^{2s}N^{s}}.
\]

Using the two bounds on $\mathbb{E}\left[\left|m-m_{(k)}\right|^{2s}\right]$ and $\mathbb{E}\left[\left|m_{(k)}-g_{k}\right|^{2s}\right]$ in Equation (\ref{eq:convex_in}), we get
\[
\mathbb{E}\left[\left|m-g_{k}\right|^{2s}\right]\leq\frac{\boldsymbol{c}_{s}}{\left|z\right|^{2s}N^{s}},
\]
where $\boldsymbol{c}_{s}=2^{2s-1}\left[1+K_{s}\right]$.
\end{proof}
As a result, we can show the concentration of the Stieltjes transform $m$ and of the $g_{k}$'s around the fixed point $\tilde{m}$:
\begin{prop}
\label{prop:loose_concentration_Stieltjes_moments} For any $N,s\in\mathbb{N}$, and any $z\in\mathbb{H}_{<0}$, we have
\begin{align*}
\mathbb{E}\left[\left|\tilde{m}-m\right|^{2s}\right] & \leq\frac{\boldsymbol{c}_{s}\left(\Tr[T_K]\right)^{2s}}{\left|z\right|^{4s}N^{3s}},\\
\mathbb{E}\left[\left|\tilde{m}-g_{k}\right|^{2s}\right] & \leq\frac{2^{2s-1}\boldsymbol{c}_{s}\left(\Tr[T_K]\right)^{2s}}{\left|z\right|^{4s}N^{3s}}+\frac{2^{2s-1}\boldsymbol{c}_{s}}{\left|z\right|^{2s}N^{s}}.
\end{align*}
where $\boldsymbol{c}_{s}$ is the same constant as in Lemma \ref{lem:loose_bound_mN_gk}.
\end{prop}
\begin{proof}
The second bound is a direct consequence of the first one, Lemma \ref{lem:loose_bound_mN_gk} and convexity. It remains to prove the first bound. Recall Equation (\ref{eq:error_mN_mtilde})
\[
\tilde{m}-m=\frac{\frac{1}{N}\sum_{k=1}^{\infty}\frac{d_{k}\left(m-g_{k}\right)}{(1+d_{k}\tilde{m})(1+d_{k}g_{k})}}{z-\frac{1}{N}\sum_{k=1}^{\infty}\frac{d_{k}}{(1+d_{k}\tilde{m})(1+d_{k}g_{k})}}.
\]
We first bound from below the norm of the denominator using Lemma \ref{lem:distance_cone}: since $\tilde{m}$ and $g_{k}$ all lie in the cone spanned by $1$ and $-\nicefrac{1}{z}$ we have
\[
\left|z-\frac{1}{N}\sum_{k=1}^{\infty}\frac{d_{k}}{(1+d_{k}\tilde{m})(1+d_{k}g_{k})}\right|\geq\left|z\right|.
\]
Using this bound, we can bound from below $\mathbb{E}\left[\left|\tilde{m}-m\right|^{2s}\right]$ by:
\[
\frac{1}{\left|z\right|^{2s}N^{2s}}\sum_{k_{1},...,k_{2s}=1}^{\infty}\frac{d_{k_{1}}\cdots d_{k_{2s}}}{\left|1+d_{k_{1}}\tilde{m}\right|\cdots\left|1+d_{k_{2s}}\tilde{m}\right|}\mathbb{E}\left[\left|m-g_{k_{1}}\right|\cdots\left|m-g_{k_{2s}}\right|\right],
\]
and hence, using a generalization of Cauchy-Schwarz inequality (Lemma \ref{lem:multiple-Cauchy-Schwarz}), by:
\[
\frac{1}{\left|z\right|^{2s}N^{2s}}\sum_{k_{1},...,k_{2s}=1}^{\infty}\frac{d_{k_{1}}\cdots d_{k_{2s}}}{\left|1+d_{k_{1}}\tilde{m}\right|\cdots\left|1+d_{k_{2s}}\tilde{m}\right|}\left(\mathbb{E}\left[\left|m-g_{k_{1}}\right|^{2s}\right]\cdots\mathbb{E}\left[\left|m-g_{k_{2s}}\right|^{2s}\right]\right)^{\frac{1}{2s}}.
\]
Using the fact that $\Re\left(\tilde{m}\right)\geq0$ and hence $\left|1+d_{k_{1}}\tilde{m}\right|\geq1$, and using Lemma \ref{lem:loose_bound_mN_gk}, this gives the following upper bound:
\[
\mathbb{E}\left[\left|m-g_{k}\right|^{2s}\right]\leq\frac{\boldsymbol{c}_{s}}{\left|z\right|^{4s}N^{3s}}\left(\Tr[T_K]\right)^{2s}.
\]
\end{proof}

We now give tighter bounds for $\left|\tilde{m}-\mathbb{E}\left[m\right]\right|$ and $\left|\tilde{m}-\mathbb{E}\left[g_{k}\right]\right|$:

\begin{prop}
\label{cor:loose_concentration-Stieltjes-expectation} For any $N\in\mathbb{N}$
and any $z\in\mathbb{H}_{<0}$, we have
\begin{eqnarray*}
\left|\tilde{m}-\mathbb{E}\left[m\right]\right| & \leq & \frac{\Tr[T_K]}{\left|z\right|^{2}N^{2}}+\frac{2\boldsymbol{c}_{1}\left(\Tr[T_K]\right)^{2}}{\left|z\right|^{3}N^{2}}+\frac{2\boldsymbol{c}_{1}\left(\Tr[T_K]\right)^{4}}{\left|z\right|^{5}N^{4}},\\
\left|\tilde{m}-\mathbb{E}\left[g_{k}\right]\right| & \leq & \frac{1}{\left|z\right|N}+\frac{\Tr[T_K]}{\left|z\right|^{2}N^{2}}+\frac{2\boldsymbol{c}_{1}\left(\Tr[T_K]\right)^{2}}{\left|z\right|^{3}N^{2}}+\frac{2\boldsymbol{c}_{1}\left(\Tr[T_K]\right)^{4}}{\left|z\right|^{5}N^{4}},
\end{eqnarray*}
where $\boldsymbol{c}_{1}$ is the constant in Lemma\textcolor{red}{{}
}\ref{lem:loose_bound_mN_gk}.
\end{prop}
\begin{proof}
\textbf{First bound: }Following similar ideas to the one which provided Equation
(\ref{eq:error_mN_mtilde}), notice that
\begin{align*}
\tilde{m}-m & =\frac{1}{z}\frac{1}{N}\sum_{k=1}^{\infty}\frac{d_{k}\left(\tilde{m}-g_{k}\right)}{(1+d_{k}\tilde{m})(1+d_{k}g_{k})}\\
 & =\frac{1}{z}\frac{1}{N}\sum_{k=1}^{\infty}\frac{d_{k}\left(\tilde{m}-g_{k}\right)}{(1+d_{k}\tilde{m})^{2}}+\frac{1}{z}\frac{1}{N}\sum_{k=1}^{\infty}\frac{d_{k}^{2}\left(\tilde{m}-g_{k}\right)^{2}}{(1+d_{k}\tilde{m})^{2}(1+d_{k}g_{k})}\\
 & =\frac{\tilde{m}-m}{z}\frac{1}{N}\sum_{k=1}^{\infty}\frac{d_{k}}{(1+d_{k}\tilde{m})^{2}}+\frac{1}{z}\frac{1}{N}\sum_{k=1}^{\infty}\frac{d_{k}\left(m-g_{k}\right)}{(1+d_{k}\tilde{m})^{2}}+\frac{1}{z}\frac{1}{N}\sum_{k=1}^{\infty}\frac{d_{k}^{2}\left(\tilde{m}-g_{k}\right)^{2}}{(1+d_{k}\tilde{m})^{2}(1+d_{k}g_{k})},
\end{align*}
hence the new identity:
\begin{align*}
\tilde{m}-m & =\frac{\frac{1}{N}\sum_{k=1}^{\infty}\frac{d_{k}\left(m-g_{k}\right)}{(1+d_{k}\tilde{m})^{2}}+\frac{1}{N}\sum_{k=1}^{\infty}\frac{d_{k}^{2}\left(\tilde{m}-g_{k}\right)^{2}}{(1+d_{k}\tilde{m})^{2}(1+d_{k}g_{k})}}{z-\frac{1}{N}\sum_{k=1}^{\infty}\frac{d_{k}}{(1+d_{k}\tilde{m})^{2}}}.
\end{align*}
Again, using Lemma \ref{lem:distance_cone}, the norm of the denominator is bounded from below by $\left|z\right|.$ From Lemma \ref{lem:Wicks-formula-matrix}, $\mathbb{E}\left[g_{k}\right]=\mathbb{E}\left[m_{(k)}\right]$, and thus from Lemma \ref{lem:loose_bound_mN_gk}, $\left|\mathbb{E}\left[m-g_{k}\right]\right|\leq\mathbb{E}\left[\left|m-m_{(k)}\right|\right]\leq\frac{1}{\left|z\right|N}$.
Furthermore, from Proposition \ref{prop:loose_concentration_Stieltjes_moments}, $\mathbb{E}\left[\left|g_{k}-\tilde{m}\right|^{2}\right]\leq\frac{2\boldsymbol{c}_{1}\left(\Tr[T_K]\right)^{2}}{\left|z\right|^{4}N^{3}}+\frac{2\boldsymbol{c}_{1}}{\left|z\right|^{2}N}.$ Thus, the expectation of the numerator is bounded by
\[
\frac{1}{\left|z\right|N^{2}}\sum_{k=1}^{\infty}\frac{d_{k}}{\left|1+d_{k}\tilde{m}\right|^{2}}+\left(\frac{2\boldsymbol{c}_{1}\left(\Tr[T_K]\right)^{2}}{\left|z\right|^{4}N^{4}}+\frac{2\boldsymbol{c}_{1}}{\left|z\right|^{2}N^{2}}\right)\sum_{k=1}^{\infty}\frac{d_{k}^{2}}{\left|1+d_{k}\tilde{m}\right|^{2}}.
\]
Hence, using again the inequality $\left|1+d_{k}\tilde{m}\right|\geq1$, it is bounded by
\[
\frac{\Tr[T_K]}{\left|z\right|N^{2}}+\frac{2\boldsymbol{c}_{1}\left(\Tr[T_K]\right)^{2}}{\left|z\right|^{2}N^{2}}+\frac{2\boldsymbol{c}_{1}\left(\Tr[T_K]\right)^{4}}{\left|z\right|^{4}N^{4}}.
\]
This allows us to conclude that
\[
\left|\tilde{m}-\mathbb{E}\left[m\right]\right|\leq\frac{\Tr[T_K]}{\left|z\right|^{2}N^{2}}+\frac{2\boldsymbol{c}_{1}\left(\Tr[T_K]\right)^{2}}{\left|z\right|^{3}N^{2}}+\frac{2\boldsymbol{c}_{1}\left(\Tr[T_K]\right)^{4}}{\left|z\right|^{5}N^{4}}.
\]

\textbf{Second bound:} Since $\mathbb{E}\left[g_{k}\right]=\mathbb{E}\left[m_{(k)}\right]$, one has
\begin{align*}
\left|\tilde{m}-\mathbb{E}\left[g_{k}\right]\right| & \leq\left|\tilde{m}-\tilde{m}_{(k)}\right|+\left|\tilde{m}_{(k)}-\mathbb{E}\left[m_{(k)}\right]\right|,
\end{align*}
where $\tilde{m}_{(k)}$ is the unique solution in the cone spanned by $1$ and $-\nicefrac{1}{z}$ to the equation
\[
\tilde{m}_{(k)}=-\frac{1}{z}\left(1-\frac{1}{N}\sum_{m\neq k}^{\infty}\frac{d_{m}\tilde{m}_{(k)}}{1+d_{m}\tilde{m}_{(k)}}\right).
\]
From Lemma \ref{lem:distance_tilde_m_without_k}, $\left|\tilde{m}-\tilde{m}_{(k)}\right|\leq\frac{1}{\left|z\right|N}$. The second term $\left|\tilde{m}_{(k)}-\mathbb{E}\left[m_{(k)}\right]\right|$ is bounded by applying the first bound of this proposition to the Stieltjes transform $m_{(k)}$. As a result, we obtain
\[
  \left|\tilde{m}-\mathbb{E}\left[g_{k}\right]\right|\leq\frac{1}{\left|z\right|N}+\frac{\Tr[T_K]}{\left|z\right|^{2}N^{2}}+\frac{2\boldsymbol{c}_{1}\left(\Tr[T_K]\right)^{2}}{\left|z\right|^{3}N^{2}}+\frac{2\boldsymbol{c}_{1}\left(\Tr[T_K]\right)^{4}}{\left|z\right|^{5}N^{4}}.
\]
\end{proof}

\subsection{Properties of the effective dimension and SCT }
\subsubsection{General properties}
We begin with general properties on the Signal Capture Threshold $\vartheta$ (which depends on $\lambda,N$ and on the eigenvalues $d_{k}$ of $T_{K}$), valid for any kernel $K$.
\begin{prop}\label{prop:appendix-bound-sct-universal} For any $\lambda > 0$, we have
\begin{align*}
  \lambda < \sctln \leq \lambda + \oon \Tr[T_K], \quad \quad 1 \leq \psctln \leq \frac{1}{\lambda} \sctln,
\end{align*}
moreover $\sctln$ is decreasing as a function of $N$ and $\partial_{\lambda}\vartheta(\lambda,N)$ is decreasing as a function of $\lambda$.
\end{prop}
\begin{proof}
Let $\lambda>0$.
\begin{enumerate}[leftmargin=*]
\item Recall that $\sctl$ is the unique positive real number such that
\[
\sctl=\lambda+\frac{\sctl}{N}\mathrm{Tr}\left[T_{K}\left(T_{K}+\sctl I_{\mathcal{C}}\right)^{-1}\right].
\]
Since $T_{K}$ is a positive operator, $\mathrm{Tr}\left[T_{K}\left(T_{K}+\sctl I_{\mathcal{C}}\right)^{-1}\right]\geq0$ and thus $\sctl \geq\lambda$. Moreover, $T_{K}+\sctl I_{\mathcal{C}}\geq \sctl I_{\mathcal{C}}$, thus
\[
  T_{K}\left(T_{K}+\sctl I_{\mathcal{C}}\right)^{-1}\leq\frac{T_{K}}{\sctl}
\]
and thus $\sctl\leq\lambda+\frac{1}{N}\mathrm{Tr}\left[T_{K}\right]$, which gives the desired inequality.
\item Differentiating Equation (\ref{eq:t_lambda}), the derivative $\partial_{\lambda}\sctl$
is given by:
\begin{equation}\label{eq:derivative-sctl}
\partial_{\lambda}\sctl=\frac{1}{\left(1-\frac{1}{N}\mathrm{Tr}\left[\left(T_{K}\left(T_{K}+\sctl I_{\mathcal{C}}\right)^{-1}\right)^{2}\right]\right)}.
\end{equation}
Using the fact that $T_{K}\left(T_{K}+\sctl I_{\mathcal{C}}\right)^{-1}\leq I_{\mathcal{C}}$, one has
\[
\left(T_{K}\left(T_{K}+\sctl I_{\mathcal{C}}\right)^{-1}\right)^{2}\leq T_{K}\left(T_{K}+\sctl I_{\mathcal{C}}\right)^{-1},
\]
thus $0\leq\frac{1}{N}\mathrm{Tr}\left[\left(T_{K}\left(T_{K}+\sctl I_{\mathcal{C}}\right)^{-1}\right)^{2}\right]\leq\frac{1}{N}\mathrm{Tr}\left[T_{K}\left(T_{K}+\sctl I_{\mathcal{C}}\right)^{-1}\right]$.
Using Equation (\ref{eq:t_lambda}), $\frac{1}{N}\mathrm{Tr}\left[T_{K}\left(T_{K}+\sctl I_{\mathcal{C}}\right)^{-1}\right]=1-\frac{\lambda}{\sctl}$.
This yields
\[
0\leq\frac{\lambda}{\sctl}\leq1-\frac{1}{N}\mathrm{Tr}\left[\left(T_{K}\left(T_{K}+\sctl I_{\mathcal{C}}\right)^{-1}\right)^{2}\right]\leq1.
\]
 Inverting this inequality yields the desired inequalities.
 \item In order to study the variation of $\sctln$ as a function of $N$, we take the derivatives of Equation (\ref{eq:t_lambda}) w.r.t $\lambda$ and $N$, and notice that
 \[
 \partial_{N}\sctln=\frac{1}{N}(\lambda-\vartheta)\partial_{\lambda}\sctln.
 \]
 In particular, since $\vartheta > \lambda$ and $\partial_{\lambda}\vartheta\geq 1$, we get that $\partial_{N}\sctln < 0$ hence $\sctln$ is decreasing as a function of $N$.
 \item Finally, we conclude by noting that since $\partial_{\lambda}\sctln > 0$, $ \sctln$ is an increasing function of $\lambda$ and thus, from the Equation (\ref{eq:derivative-sctl}) we have that $\partial_{\lambda}\sctln$ is decreasing as a function of $\vartheta$ and thus as a function of $\lambda$.
\end{enumerate}
\end{proof}
\subsubsection{Bounds under polynomial decay hypothesis}
In this subsection only, we assume that $d_{k}=\Theta(k^{-\beta})$ with $\beta>1$, i.e, there exist $c_{\ell}$ and $c_{h}$ positive such that for any $k\geq1$, $c_{\ell}k^{-\beta}\leq d_{k}\leq c_{h}k^{-\beta}$.
We first study the asymptotic behavior of $\vartheta(0,N)$ and $\partial_{\lambda}\vartheta(0,N) $ as $N$ goes to infinity, then using these results, we investigate the asymptotic behavior of $\vartheta(\lambda ,N)$ and $\partial_{\lambda}\vartheta(\lambda ,N) $ as $N$ goes to infinity.

For any $t\in\mathbb{R}^+$, let $\mathcal N(t)$ denote the $t$-effective dimension \cite{zhang-03, caponnetto-07} defined by
\[
\mathcal{N}(t):=\sum_{k=1}^{\infty}\frac{d_{k}}{t+d_{k}}.
\]
For any $\lambda>0$, the SCT is the unique solution of $\vartheta\left(\lambda,N\right)=\lambda+\frac{\vartheta\left(\lambda,N\right)}{N}\mathcal{N}(\vartheta(\lambda,N)).$
In particular, $\vartheta(0,N)$ is the unique solution of $\mathcal{N}(\vartheta(0,N))=N$.

Since $\mathcal{N}(t)$ is decreasing from $\infty$ to $0$, in order to study the asymptotic behavior of $\vartheta(0,N)$ as $N$ goes to infinity, one has to understand the rate of explosion of $\mathcal{N}(t)$ as $t$ goes to zero.
\begin{lem}
\label{lem:eff-dim}If $d_{k}=\Theta(k^{-\beta})$ with $\beta>1$,
then $\mathcal{N}(t)=\Theta(t^{-\frac{1}{\beta}})$ when
$t\to0$.
\end{lem}
\begin{proof}
For any $m\in\mathbb{R}_{+}$, $\mathcal{N}(t)=\sum_{k\leq m}\frac{d_{k}}{t+d_{k}}+\sum_{k>m}^{\infty}\frac{d_{k}}{t+d_{k}}\leq m+t^{-1}\sum_{k>m}d_{k}$.
Then there exists $c,d>0$ such that $\sum_{k>m}d_{k}\leq c\sum_{k>m}k^{-\beta}\leq dm^{1-\beta}$.
Thus $\mathcal{N}(t)$ is bounded by $m+dt^{-1}m^{1-\beta}$ for any
$m$. Taking $m=t^{-1}m^{1-\beta}$, i.e. $m=t^{-\frac{1}{\beta}}$,
one gets that $\mathcal{N}(t)\leq Ct^{-\frac{1}{\beta}}.$

For the lower bound, notice that $\mathcal{N}(t)\geq\sum_{k\mid d_{k}\geq t}\frac{d_{k}}{t+d_{k}}\geq\frac{1}{2}\#\left\{ k\mid d_{k}\geq t\right\} $.
Using the fact that there exists $c_{\ell}>0$ such that $d_{k}\geq c_{\ell}k^{-\beta}$,
$\#\left\{ k\mid d_{k}\geq t\right\} \geq\#\left\{ k\mid c_{\ell}k^{-\beta}\geq t\right\} =\left\lfloor \left(\nicefrac{t}{c_{\ell}}\right)^{-\frac{1}{\beta}}\right\rfloor $.
This yields the lower bound on $\mathcal{N}(t)$.
\end{proof}

\begin{lem}
\label{lem:SCT-equiv}If $d_{k}=\Theta(k^{-\beta})$ with $\beta>1$,
then $\vartheta(0,N)=\Theta\left(N^{-\beta}\right)$.
\end{lem}
\begin{proof}
From the previous lemma, there exist $b_{\ell}, b_{h}>0$ such that $b_{\ell}\vartheta(0,N)^{-\frac{1}{\beta}}\leq\mathcal{N}(\vartheta(0,N))\leq b_{h}\vartheta(0,N)^{-\frac{1}{\beta}}$.
From the definition of $\vartheta(0,N),$ $\mathcal{N}(\vartheta(0,N))=N$, thus we
get $\left(\nicefrac{N}{b_{\ell}}\right)^{-\beta}\leq \vartheta(0,N)\leq\left(\nicefrac{N}{b_{h}}\right)^{-\beta}.$
\end{proof}
With no assumption on the spectrum of $T_{K}$, the upper bound for the derivative of the SCT $\partial_{\lambda}\vartheta$ obtained in Proposition \ref{prop:bound-sct-universal}, becomes useless in the ridgeless limit $\lambda\to0$. Yet, with the assumption of power-law decay of the eigenvalues of $T_{K}$ we can refine the bound with a meaningful one. In order to obtain this we first prove a technical lemma.
\begin{lem}\label{lem:polynomial-derivative-finite}
If $d_{k}=\Theta(k^{-\beta})$ with $\beta>1$, then $\sup_{N}\partial_{\lambda}\vartheta(0,N)<\infty$.
\end{lem}

\begin{proof}
The derivative of the SCT with respect to $\lambda$ at $\lambda=0$
is given by:
\begin{align*}
\partial_{\lambda}\vartheta(0,N)=\frac{N}{\vartheta(0,N)\sum_{k=1}^{\infty}\frac{d_{k}}{\left(\vartheta(0,N)+d_{k}\right)^{2}}}.
\end{align*}
Set $\alpha>1$, then for all $d_{k}\in[\alpha^{-1}t, \alpha t]$, we have that $\frac{d_{k}}{(t+d_{k})^{2}} \geq \frac{\alpha t }{(t+\alpha t)^{2}}=\frac{\alpha}{(1+\alpha)^{2}}\frac{1}{t}$.
Thus,
\[
t\sum_{k=1}^{\infty}\frac{d_{k}}{(t+d_{k})^{2}}\geq t\sum_{\alpha^{-1}t<d_{k}< \alpha t}\frac{d_{k}}{(t+d_{k})^{2}} \geq \frac{\alpha}{(1+\alpha)^{2}} \#\{k\mid \alpha^{-1}t<d_{k}< \alpha t\}.
\]
It follows that
\[
\partial_{\lambda}\vartheta(0,N) \leq N \frac{(1+\alpha)^{2}}{\alpha}\frac{1}{\#\{k\mid \alpha^{-1}\vartheta(0,N)  <d_{k}< \alpha \vartheta(0,N) \}}
\]
Now, using Lemma \ref{lem:SCT-equiv}, we are going to find a value of $\alpha$ such that $\#\{k\mid\alpha^{-1}\vartheta(0,N)  <d_{k}< \alpha \vartheta(0,N) \} \geq  c N$ for some universal constant $c$: this will conclude the proof.

By using the assumption that there exist $c_{\ell},c_{h}>0$ such that $c_{\ell}k^{-\beta} \leq d_{k} \leq c_{h}k^{-\beta}$, in Lemma \ref{lem:SCT-equiv} we saw that there exist $c'_{\ell},c'_{h}>0$ such that $c'_{\ell}N^{-\beta} \leq \vartheta(0,N) \leq c'_{h}N^{-\beta}$. For sake of simplicity, let us assume that the ratios $\frac{c_{\ell}}{c'_{\ell}}$ and  $\frac{c_{h}}{c'_{h}}$ are not integer. Hence we have
\begin{align*}
\#\{k\mid \alpha^{-1}\vartheta(0,N)  \leq d_{k}\leq \alpha \vartheta(0,N) \}&\geq \#\left\{k\mid\frac{1} {\alpha c_{\ell}}\vartheta(0,N) \leq k^{-\beta}\leq \frac{\alpha}{c_{h}} \vartheta(0,N)\right \} \\
&\geq \#\left\{k\mid\frac{1} {\alpha c_{\ell}}c'_{h}N^{-\beta} \leq k^{-\beta}\leq \frac{\alpha}{c_{h}} c'_{\ell}N^{-\beta} \right \} \\
&= \left(\left\lfloor\left(\frac{\alpha c_{\ell}}{c'_{h}}\right)^{\frac{1}{\beta}}\right\rfloor-\left\lfloor \left( \frac{c_{h}}{\alpha c'_{\ell}}\right)^{\frac{1}{\beta}}\right\rfloor \right)N
\end{align*}
For one of the two values $\alpha\in\{\frac{c_{h}}{c_{\ell}},\alpha=\frac{c'_{h}}{c'_{\ell}}\}$, we have a meaningful (positive) bound:
\[
\#\left\{k\mid \alpha^{-1}\vartheta(0,N)  \leq d_{k}\leq \alpha \vartheta(0,N) \right\}\geq\left\vert\left\lfloor\left( \frac{c_{h}}{c'_{h}}\right)^{\frac{1}{\beta}}\right\rfloor-\left\lfloor \left( \frac{c_{\ell}}{c'_{\ell}}\right)^{\frac{1}{\beta}}\right\rfloor \right\vert N.
\]
This allows us to conclude.
\end{proof}

\begin{prop} \label{prop:appendix-bound-sct-polynomial}
If there exist $\beta>1$ and $c_{\ell},c_{h}>0$ s.t. for any $k\in \mathbb N$, $c_{\ell}k^{-\beta}\leq d_{k}\leq c_{h}k^{-\beta}$, then for any integer $N$,
\begin{enumerate}[leftmargin=*]
\item $\lambda+a_{\ell}N^{-\beta}\leq \vartheta(\lambda,N)\leq c\lambda+a_{h}N^{-\beta}$,
\item $1\leq\partial_{\lambda}\vartheta(\lambda,N)\leq c$,
\end{enumerate}
where $a_{\ell},a_{h}\geq0$ and $c\geq1$ depend only on $c_{\ell},c_{h},\beta$.
\end{prop}
\begin{proof}
We start by proving the inequalities for the derivative of the SCT $\partial_{\lambda}\vartheta(\lambda,N)$. The left side of the inequality has already been proven in Proposition \ref{prop:bound-sct-universal}. For the right side, from Proposition \ref{prop:bound-sct-universal}, the derivative $\partial_\lambda \vartheta(\lambda,N)$ is decreasing in $\lambda$. In particular, by Lemma \ref{lem:polynomial-derivative-finite},  $\partial_{\lambda}\vartheta(\lambda,N) \leq \sup_{N} \partial_{\lambda}\vartheta(0,N) < \infty. $
Thus, the right side holds with $c:= \sup_{N}\partial_{\lambda}\vartheta(0,N)$.

The inequality for the SCT $\vartheta(\lambda,N)$ is then obtained by integrating the second inequality and by using the initial value condition $a_{\ell}N^{-\beta}\leq \vartheta(0,N) \leq a_{h}N^{-\beta}$ provided by Lemma \ref{lem:SCT-equiv}.
\end{proof}

\subsection{The Operator $A(z)$}

We have now the tools to describe the moments of the operator $A(z)$
which allow us to describe the moments of the predictor $\hat{f}_{\lambda}$.

\subsubsection{Expectation}

Writing $\tilde{A}_{\vartheta(-z)} = T_{K}\left(T_{K}+\vartheta(-z) I_{\mathcal{C}}\right)^{-1}$ and for any diagonalizable operator $A$ writing $|A|$ for the operator with the same eigenfunctions but with eigenvalues replaced by their absolute values, we have:

\begin{thm}
\label{thm:approx_expectation_A}  For any $z \in \mathbb{H}_{<0}$, for any $f, g \in \mathcal{C}$, we have
\begin{align}
  \label{eq:expectation_A}
  \left|\left\langle f,\left(\mathbb{E}\left[A(z)\right]-\tilde A_{\vartheta(-z)}\right)g\right\rangle _{S}\right| \leq \left| \bra f, |\tilde{A}_{\vartheta(-z)}| |I_\mathcal{C} - \tilde{A}_{\vartheta(-z)}| g\ket_S \right| \left(\oon+ \mathcal{P}\left(\frac{\Tr[K]}{|z|N}\right)\right)
  \end{align}
using the big-P notation of Definition \ref{def:big-p}.
\end{thm}
\begin{rem}
Note that in particular since the polynomial implicitly embedded in $ \mathcal P $ vanishes at $ 0 $, the right hand side tends to $ 0 $ as $ N\to\infty$.
\end{rem}

\begin{proof}
As before, let $(f^{(k)})_{k\in\mathbb N}$ be the orthonormal basis of $\mathcal C$ defined above and $A_{k\ell}(z)=\langle f^{(k)},A(z) f^{(\ell)}\rangle_S$. Using a symmetry argument, we first show that for any $\ell \neq k$, $\mathbb{E}\left[A_{\ell k}(z)\right]=0$: this implies that $\mathbb{E}\left[A(z)\right]$ and $\tilde A_{\vartheta(-z)}$ have the same eigenfunctions $f^{(k)}$. Thus, to conclude the proof, we only need to prove Equation \ref{eq:expectation_A} for $f=g=f^{(k)}$.

\begin{itemize}[leftmargin=*]
\item \textbf{Off-Diagonal terms:} By a symmetry argument, we show that the off-diagonal terms are null. Consider the map $s_k:\mathcal C \to \mathcal C$ defined by $s_k: f\mapsto f-2 \bra f,f^{(k)}\ket_S f^{(k)}$, and note that $s_k(f^{(m)})=f^{(m)}$ if $m\neq k$ and $s_k(f^{(k)})=-f^{(k)}$. The map $s_k$ is a symmetry for the observations, i.e. for any observations $o_1,\ldots,o_N$, and any functions $f_1, \ldots, f_N$, the vector $(o_i(s_k(f_i))_{i=1,\ldots,N}$ and $(o_i(f_i))_{i=1,\ldots,N}$ have the same law. Thus, the sampling operator $\OO$ and the operator $\OO s_k$ have the same law, hence so do $A(z)$ and $A^{s_{k}}(z)$, where
\[
A^{s_{k}}(z):=\frac{1}{N}Ks_{k}^{T}\mathcal{O}^{T}(\frac{1}{N}\mathcal{O}s_{k}Ks_{k}^{T}\mathcal{O}^{T}-zI_{N})^{-1}\mathcal{O}s_{k}.
\]
Note that $K s_{k}^{T} = s_{k} K$ and since $s_{k}^2=\mathrm{Id}$, $s_k K s_{k}^{T} = K$. This implies that $A^{s_{k}}(z) = s_{k} A(z) s_{k}$. For any $\ell \neq k$, $A_{\ell k}^{s_{k}}(z) = - A_{\ell k}(z)$, hence $\mathbb{E}[ A_{\ell k}(z)]=0$.

\item \textbf{Diagonal terms:} Using Equation \ref{eq:diagonal_term}, we have
\begin{align*}
	A_{kk}(z)=\frac{d_{k}g_{k}}{1+d_{k}g_{k}}&=\frac{d_{k}\tilde{m}}{1+d_{k}\tilde{m}}+\frac{d_{k}(g_{k}-\tilde{m})}{\left(1+d_{k}\tilde{m}\right)\left(1+d_{k}g_{k}\right)}\\
	&=\frac{d_{k}\tilde{m}}{1+d_{k}\tilde{m}}+\frac{d_{k}(g_{k}-\tilde{m})}{\left(1+d_{k}\tilde{m}\right)^{2}}-\frac{d_{k}^{2}(g_{k}-\tilde{m})^{2}}{\left(1+d_{k}\tilde{m}\right)^{2}\left(1+d_{k}g_{k}\right)}.
\end{align*}
From this, using the fact that $\Re(g_k)>0$, we obtain
\begin{align*}
	\left|\mathbb{E}\left[A_{kk}(z)\right]-\frac{d_{k}\tilde{m}}{1+d_{k}\tilde{m}}\right| \leq \frac{d_{k}\left|\mathbb{E}\left[g_{k}\right]-\tilde{m}\right|}{\left|1+d_{k}\tilde{m}\right|^{2}}+\frac{d_{k}^{2}\mathbb{E}\left[\left|g_{k}-\tilde{m}\right|^{2}\right]}{\left|1+d_{k}\tilde{m}\right|^{2}}.
\end{align*}
Using Proposition \ref{cor:loose_concentration-Stieltjes-expectation}, we can bound the first fraction by
\begin{align*}
\frac{d_{k}\left|\mathbb{E}\left[g_{k}\right]-\tilde{m}\right|}{\left|1+d_{k}\tilde{m}\right|^{2}}&\leq\frac{d_{k}}{\left|1+d_{k}\tilde{m}\right|^{2}}\left(\frac{1}{\left|z\right|N}+\frac{\Tr[T_K]}{\left|z\right|^{2}N^{2}}+\frac{2\boldsymbol{c}_{1}\left(\Tr[T_K]\right)^{2}}{\left|z\right|^{3}N^{2}}+\frac{2\boldsymbol{c}_{1}\left(\Tr[T_K]\right)^{4}}{\left|z\right|^{5}N^{4}}\right) \\
	&\leq \frac{d_{k}|\vartheta(-z)|^2}{\left|\vartheta(-z)+d_{k}\right|^2} \left(\frac{1}{\left|z\right|N}+\frac{\Tr[T_K]}{\left|z\right|^{2}N}+\frac{2\boldsymbol{c}_{1}\left(\Tr[T_K]\right)^{2}}{\left|z\right|^{3}N^{2}}+\frac{2\boldsymbol{c}_{1}\left(\Tr[T_K]\right)^{4}}{\left|z\right|^{5}N^{4}}\right) \\
	&\leq \frac{d_{k}}{\left|\vartheta(-z)+d_{k}\right|}\left|1-\frac{d_{k}}{\vartheta(-z)+d_{k}}\right| \left(\frac{1}{N} + \mathcal{P}\left(\frac{\Tr[T_K]}{\left|z\right|N}\right) \right),
	\end{align*}
by substituting  $\vartheta(-z)=\frac{1}{\tilde{m}(z)}$, using the bound $ \vert \vartheta(-z)\vert \leq |z| + \frac{\Tr[T_K]}{N}$ (see Proposition \ref{prop:bound-sct-universal}).

Using Proposition \ref{prop:loose_concentration_Stieltjes_moments}, the inequality $d_k^2 \leq d_k \Tr[T_K]$ and similar arguments as above, we can bound the second fraction by
\begin{align*}
	\frac{d_{k}^{2}\mathbb{E}\left[\left|g_{k}-\tilde{m}\right|^{2}\right]}{\left|1+d_{k}\tilde{m}\right|^{2}} &\leq\frac{d_{k}^{2}}{\left|1+d_{k}\tilde{m}\right|^{2}}\left(\frac{2\boldsymbol{c}_{1}\left(\Tr[T_K]\right)^{2}}{\left|z\right|^{4}N^{3}}+\frac{2\boldsymbol{c}_{1}}{\left|z\right|^{2}N}\right) \\
	&\leq \frac{d_{k}|\vartheta(-z)|^2}{\left|\vartheta(-z)+d_{k}\right|^2} \left(\frac{2\boldsymbol{c}_{1}\left(\Tr[T_K]\right)^{3}}{\left|z\right|^{4}N^{3}}+\frac{2\boldsymbol{c}_{1}\Tr[T_K]}{\left|z\right|^{2}N}\right) \\
	&\leq  \frac{d_{k}}{\left|\vartheta(-z)+d_{k}\right|}\left|1-\frac{d_{k}}{\vartheta(-z)+d_{k}}\right| \mathcal{P} \left(\frac{\Tr[T_K]}{\left|z\right|N}\right)
  \end{align*}
Finally, putting everything together, we get:
\begin{equation}\label{eq:bound-expectation-diagonal-A}
\left|\mathbb{E}\left[A_{kk}(z)\right]-\frac{d_{k}\tilde{m}}{1+d_{k}\tilde{m}}\right| \leq\frac{d_{k}}{\left|\vartheta(-z)+d_{k}\right|}\left|1-\frac{d_{k}}{\vartheta(-z)+d_{k}}\right| \left(\frac{1}{N} + \mathcal{P}\left(\frac{\Tr[T_K]}{\left|z\right|N}\right)\right)
\end{equation}
\end{itemize}
\end{proof}

\subsubsection{Variance}
To study the variance of $A(z)$ we will need to apply the Shermann-Morrison formula twice, to isolate the contribution of the two eigenfunctions $f^{(k)}$ and $f^{(\ell)}$. Similarly to above, we set $K_{(k\ell)}=\sum_{n\notin \{k,\ell\}}d_{n}f^{(n)}\otimes f^{(n)}$ and
we define
\begin{align*}
B_{(k\ell)}(z) = \frac{1}{N}\mathcal{O}K_{(k\ell)}\mathcal{O}^{T}-zI_{N}, \quad \quad
m_{(k\ell)}(z) =\frac{1}{N}\mathrm{Tr}\left[B_{(k\ell)}(z)^{-1}\right].
\end{align*}
Note that the concentration results of Section \ref{subsec:Appendix-Concentration-Stieltjes} apply to $m_{(k\ell)}$: it concentrates around $\tilde{m}_{(k\ell)}$, the unique solution, in the cone spanned by $1$ and $-\nicefrac{1}{z}$, to the equation
\[
\tilde{m}_{(k\ell)}=-\frac{1}{z}\left(1-\frac{\tilde{m}_{(k\ell)}}{N}\mathrm{Tr}\left[T_{K_{(k\ell)}}\left(T_{K_{(k\ell)}}+\tilde{m}_{(k\ell)}I_{\mathcal{C}}\right)^{-1}\right]\right).
\]


In order to compute the off-diagonal entry $A_{k\ell}(z)=\frac{1}{N}d_{k}\mathcal{O}_{.k}^{T}B(z)^{-1}\mathcal{O}_{.\ell}$, we use the Shermann-Morrison formula twice: when applied to $B(z)=B_{(k)}(z)+\frac{d_{k}}{N}\mathcal{O}_{.k}\mathcal{O}_{.k}^{T}$ we get
\[
  B(z)^{-1}=B_{(k)}(z)^{-1}-\frac{d_{k}}{N}\frac{B_{(k)}(z)^{-1}\mathcal{O}_{.k}\mathcal{O}_{.k}^{T}B_{(k)}(z)^{-1}}{1+\frac{d_{k}}{N}\mathcal{O}_{.k}^{T}B_{(k)}(z)^{-1}\mathcal{O}_{.k}};
\]
thus, recalling that $g_{(k)}=\frac{1}{N}\mathcal{O}_{.,k}^{T}B_{(k)}(z)^{-1}\mathcal{O}_{.,k}$, we have
\[
  A_{k\ell}(z) = \frac{d_k}{1+d_k g_k}\frac{1}{N}\mathcal{O}_{.k}^{T}B_{(k)}(z)^{-1}\mathcal{O}_{.\ell}.
\]
We then apply the Shermann-Morrison formula to $B_{(k)}(z)=B_{(k\ell)}(z)+\frac{d_{\ell}}{N}\mathcal{O}_{.\ell}\mathcal{O}_{.\ell}^{T}$ and obtain
\[
  B_{(k)}(z)^{-1}=B_{(k\ell)}(z)^{-1}-\frac{d_{\ell}}{N}\frac{B_{(k\ell)}(z)^{-1}\mathcal{O}_{.\ell}\mathcal{O}_{.\ell}^{T}B_{(k\ell)}(z)^{-1}}{1+\frac{d_{\ell}}{N}\mathcal{O}_{.\ell}^{T}B_{(k\ell)}(z)^{-1}\mathcal{O}_{.\ell}}.
  \]
Thus, we obtain the following formula for the off-diagonal entry:
\begin{align}
\label{eq:off-diagonal-formula}
A_{k\ell}(z)=\frac{d_k}{1+d_{k}g_{k}}\frac{h_{k\ell}}{1+d_{\ell}h_{\ell}}
\end{align}
where $h_{\ell}=\frac{1}{N}\left(\mathcal{O}_{.\ell}\right)^{T}B_{(k\ell)}^{-1}(z)\mathcal{O}_{. \ell}$
and $h_{k\ell}=\frac{1}{N}\left(\mathcal{O}_{. k}\right)^{T}B_{(k\ell)}^{-1}(z)\mathcal{O}_{. \ell}$.

We can apply the results of Section \ref{subsec:Appendix-Concentration-Stieltjes} showing the concentration of $g_{k}$ around $\tilde{m}_{(k)}$: $h_{\ell}$ concentrates around $\tilde{m}_{(k)}$ which itself is close to $\tilde{m}$:
\begin{lem}
\label{cor:concentration_moments_h_K_m_tilde}For $z\in\mathbb{H}_{<0}$, and $s\in \mathbb{N}$,
we have
\[
\mathbb{E}\left[\left|h_{\ell}-\tilde{m}\right|^{2s}\right]\leq\frac{\boldsymbol{a}_s\left(\mathrm{Tr}[T_{K}]\right)^{2s}}{\left|z\right|^{4s}N^{3s}}+\frac{\boldsymbol{b}_s}{\left|z\right|^{2s}N^{s}},
\]
where $\boldsymbol{a}_s, \boldsymbol{b}_s$ only depend on $s$.
\end{lem}
\begin{proof}
By convexity, for $k\neq \ell$,
\begin{align*}
\mathbb{E}\left[\left|h_{\ell}-\tilde{m}\right|^{2s}\right] & \leq2^{2s-1}\mathbb{E}\left[\left|h_{\ell}-\tilde{m}_{(k)}\right|^{2s}\right]+2^{2s-1}\left|\tilde{m}_{(k)}-\tilde{m}\right|^{2s}\\
 & \leq2^{2s-1}\left(\frac{2^{2s-1}\boldsymbol{c}_{s}\left(\Tr[T_K]\right)^{2s}}{\left|z\right|^{4s}N^{3s}}+\frac{2^{2s-1}\boldsymbol{c}_{s}}{\left|z\right|^{2s}N^{s}}\right)+\frac{2^{2s-1}}{|z|^{2s}N^{2s}}
\end{align*}
where for the first term, we applied Proposition \ref{prop:loose_concentration_Stieltjes_moments} to the matrix $B_{(k)}$ instead of $B$ and the second term is bounded by $\left|\tilde{m}_{(k)}-\tilde{m}\right|\leq\frac{1}{\vert z \vert N}$ by Lemma \ref{lem:distance_tilde_m_without_k}. Finally, letting $\boldsymbol{a}_s=4^{2s-1}\boldsymbol{c}_s$ and $\boldsymbol{b}_s=4^{2s-1}\boldsymbol{c}_s+2^{2s-1}$, we obtain the result.
\end{proof}

The scalar $h_{k\ell}$ on the other hand has $0$ expectation and, using Wick's formula (Lemma \ref{lem:Wicks-formula-matrix}), its variance $\mathbb{E}\left[h_{k\ell}^{2}\right]$ is equal to $\frac{1}{N^2} \mathbb{E}[\Tr[B_{(k\ell)}^{-2}]]=\frac{1}{N}\mathbb{E}\left[\partial_{z}m_{(k\ell)}(z)\right]$. Since $\mathbb{E}\left[m_{(k\ell)}(z)\right]$ is close to $\tilde{m}(z)$, from Lemma \ref{lem:cauchy_inequality}, its derivative, and hence the variance of $h_{k\ell}$, is close to $\frac{1}{N}\partial_{z}\tilde{m}$:
\begin{lem}
\label{cor:concentration_m_km_and_derivative}For $z\in\mathbb{H}_{<0},$ we
have:
\begin{align*}
\left|\mathbb{E}\left[m_{(k\ell)}(z)\right]-\tilde{m}(z)\right| & \leq \frac{\Tr[T_K]}{\left|z\right|^{2}N^{2}}+\frac{2\boldsymbol{c}_{1}\left(\Tr[T_K]\right)^{2}}{\left|z\right|^{3}N^{2}}+\frac{2\boldsymbol{c}_{1}\left(\Tr[T_K]\right)^{4}}{\left|z\right|^{5}N^{4}}+\frac{2}{ |z| N} ,
\end{align*}
where $\boldsymbol{c}_1$ is as in Proposition \ref{cor:loose_concentration-Stieltjes-expectation}.
\end{lem}
\begin{proof}
We use Proposition \ref{cor:loose_concentration-Stieltjes-expectation} and Lemma \ref{lem:distance_tilde_m_without_k} twice to obtain
\begin{align*}
\left|\mathbb{E}\left[m_{(k\ell)}(z)\right]-\tilde{m}(z)\right| & \leq\left|\mathbb{E}\left[m_{(k\ell)}(z)\right]-\tilde{m}_{(k\ell)}(z)\right|+\left|\tilde{m}_{(k\ell)}(z)-\tilde{m}_{(k)}(z)\right|+\left|\tilde{m}_{(k)}(z)-\tilde{m}(z)\right|\\
 & \leq \frac{\Tr[T_K]}{\left|z\right|^{2}N^{2}}+\frac{2\boldsymbol{c}_{1}\left(\Tr[T_K]\right)^{2}}{\left|z\right|^{3}N^{2}}+\frac{2\boldsymbol{c}_{1}\left(\Tr[T_K]\right)^{4}}{\left|z\right|^{5}N^{4}}+\frac{2}{ |z| N},
\end{align*}
which yields the desired result.
\end{proof}
To approximate the variance $\mathrm{Var}\left(\left\langle f^{(k)},A_{\lambda}f^{*}\right\rangle _{S}\right)$
of the coordinate of the noiseless predictor, we need the following results regarding the covariance of the entries
of $A(z)$.
\begin{prop}
\label{prop:variance_entries_A}For $z\in\mathbb{H}_{<0}$, any $k, \ell \in \mathbb{N}$, we have
\begin{align*}
\left|\mathrm{Var}\left(A_{kk}(z)\right)-\frac{2}{N}\frac{d_{k}^{2}\partial_{z}\tilde{m}}{\left(1+d_{k}\tilde{m}\right)^{4}}\right| & \leq \frac{1}{N}\frac{d_{k}^{2}|\partial_{z}\tilde{m}|}{\left|1+d_{k}\tilde{m}\right|^{4}} \left(\frac 1 N + \frac{|z|}{-\Re(z)}\mathcal{P}\left(\frac{\Tr[T_K]}{|z|N^{\frac 1 2}}\right)\right)\\
\left|\mathrm{Var}\left(A_{k\ell}(z)\right)-\frac{1}{N}\frac{d_{k}^{2}\partial_{z}\tilde{m}}{\left(1+d_{k}\tilde{m}\right)^{2}\left(1+d_{\ell}\tilde{m}\right)^{2}}\right| & \leq \frac{1}{N}\frac{d_{k}^{2} | \partial_{z}\tilde{m}|}{\left|1+d_{k}\tilde{m}\right|^{2}\left|1+d_{\ell}\tilde{m}\right|^{2}}\frac{|z|}{-\Re(z)}\mathcal{P}\left(\frac{\Tr[T_K]}{|z|N^{\frac 1 2}}\right) \\
\left|\mathrm{Cov}\left(A_{k\ell}(z),A_{\ell k}(z)\right)-\frac{1}{N}\frac{d_{k}d_{\ell}\partial_{z}\tilde{m}}{\left(1+d_{k}\tilde{m}\right)^{2}\left(1+d_{\ell}\tilde{m}\right)^{2}}\right| & \leq \frac{1}{N}\frac{d_{k}d_{\ell} | \partial_{z}\tilde{m}|}{\left|1+d_{k}\tilde{m}\right|^{2}\left|1+d_{\ell}\tilde{m}\right|^{2}}\frac{|z|}{-\Re(z)}\mathcal{P}\left(\frac{\Tr[T_K]}{|z|N^{\frac 1 2}}\right)
\end{align*}
where we use the big-P notation of Definition \ref{def:big-p}.
Whenever a value in the quadruple $(k,h,n,\ell)$ appears an odd number of times, we have
\[
\mathrm{Cov}\left(A_{kh}(z),A_{n\ell}(z)\right)=0.
\]
\end{prop}

\begin{proof}
Let $s_{k}$ be the symmetry map in the proof of Theorem \ref{thm:approx_expectation_A}: the matrices $A(z)$ and $A^{s_{k}}(z)$ have the same law. Since $A_{\ell n}^{s_{k}}(z)=-A_{\ell n}(z)$ whenever exactly one of $\ell, n$ is equal to $k$, we have for $h,n,\ell$ distinct from $k$:
\[
\mathrm{Cov}\left(A_{kh}(z),A_{n\ell}(z)\right)=\mathrm{Cov}\left(A_{kh}^{s_{k}}(z),A_{n\ell}^{s_{k}}(z)\right)=\mathrm{Cov}\left(-A_{kh}(z),A_{n\ell}(z)\right)
\]
which implies that $\mathrm{Cov}\left(A_{kh}(z),A_{n\ell}(z)\right)=0$ when $h,n,\ell$ are distinct from $k$. More generally, it is easy to see that $\mathrm{Cov}\left(A_{kh}(z),A_{n\ell}(z)\right)=0$ whenever a value in the quadruple $(k,h,n,\ell)$ appears an odd number of times.

\textbf{Approximation of $\mathrm{Var}\left(A_{kk}(z)\right)$:} Since $\mathbb E [A_{kk}(z)]\approx \frac{d_k \tilde m}{1+d_k\tilde m}$ (Theorem \ref{thm:approx_expectation_A}), we decompose the variance of $A_{kk}(z)$ as follows:
\begin{align*}
  \mathrm{Var}\left(A_{kk}\right)=\mathbb{E}\left[\left(A_{kk}-\frac{d_{k}\tilde{m}}{1+d_{k}\tilde{m}}\right)^{2}\right]-\left[\mathbb{E}\left[A_{kk}\right]-\frac{d_{k}\tilde{m}}{1+d_{k}\tilde{m}}\right]^{2}.
\end{align*}
This gives us an approximation $\mathrm{Var}\left(A_{kk}\right)\approx\mathbb{E}\left[\left(A_{kk}-\frac{d_{k}\tilde{m}}{1+d_{k}\tilde{m}}\right)^{2}\right]$ since the term $\left|\mathbb{E}\left[A_{kk}\right]-\frac{d_{k}\tilde{m}}{1+d_{k}\tilde{m}}\right|^{2}$, by using Theorem \ref{thm:approx_expectation_A}, we get the following bound :
\begin{align*}
\left|\mathbb{E}\left[A_{kk}\right]-\frac{d_{k}\tilde{m}}{1+d_{k}\tilde{m}}\right|^{2} &\leq \left|\frac{d_k \tilde m}{(1+d_k\tilde m)^2}\left(\frac{1}{N}+ \mathcal{P}\left(\frac{\Tr[T_K]}{|z|N}\right)\right)\right|^2 \\
&= \frac{1}{N} \frac{d_k^2 |\tilde m|^2}{|1+d_k\tilde m|^4}\left(\frac{1}{N} + 2 \mathcal{P}\left(\frac{\Tr[T_K]}{|z|N}\right) + N \mathcal{P}\left(\frac{\Tr[T_K]}{|z|N}\right)^2 \right)
\end{align*}
Since $ \mathcal{P}\left(\frac{\Tr[T_K]}{|z|N}\right) = \mathcal{P}\left(\frac{\Tr[T_K]}{|z|N^{\nicefrac{1}{2}}}\right)$ and $N \mathcal{P}\left(\frac{\Tr[T_K]}{|z|N}\right)^2 = \mathcal{P}\left(\frac{(\Tr[T_K])^2}{|z|^2 N}\right)$, we can bound $\left|\mathbb{E}\left[A_{kk}\right]-\frac{d_{k}\tilde{m}}{1+d_{k}\tilde{m}}\right|^{2} $ by
\begin{align*}
\frac{1}{N} \frac{d^2_k |\tilde m|^2}{|1+d_k\tilde m|^4} \left(\frac{1}{N} + \mathcal{P}\left(\frac{\Tr[T_K]}{|z|N^{\frac 1 2}}\right) \right).
\end{align*}

Using Formula (\ref{eq:diagonal_term}) for the diagonal entries of $A$, we have:
\[
  \left(A_{kk}-\frac{d_{k}\tilde{m}}{1+d_{k}\tilde{m}}\right)^{2}=\frac{d_{k}^{2}\left[g_{k}-\tilde{m}\right]^{2}}{(1+d_{k}g_{k})^{2}(1+d_{k}\tilde{m})^{2}}.
\]
which can be also expressed as:
\[
\left(\frac{d_{k}\left[g_{k}-\tilde{m}\right]}{(1+d_{k}g_{k})(1+d_{k}\tilde{m})}\right)^{2}=\left(\frac{d_{k}\left[g_{k}-\tilde{m}\right]}{(1+d_{k}\tilde{m})(1+d_{k}\tilde{m})}-\frac{d_{k}^2\left[g_{k}-\tilde{m}\right]^{2}}{(1+d_{k}g_{k})(1+d_{k}\tilde{m})^{2}}\right)^{2}.
\]
This yields
\begin{align*}
  \mathbb{E}\left[\left(A_{kk}-\frac{d_{k}\tilde{m}}{1+d_{k}\tilde{m}}\right)^{2}\right]&-\mathbb{E}\left[\left(\frac{d_{k}\left[g_{k}-\tilde{m}\right]}{(1+d_{k}\tilde{m})^{2}}\right)^{2}\right] \\
  & =-\mathbb{E}\left[\frac{d^2_{k}\left[g_{k}-\tilde{m}\right]^{2}}{(1+d_{k}g_{k})(1+d_{k}\tilde{m})^{2}}\left(\frac{2d_{k}\left[g_{k}-\tilde{m}\right]}{(1+d_{k}\tilde{m})(1+d_{k}\tilde{m})}-\frac{d_{k}^2\left[g_{k}-\tilde{m}\right]^{2}}{(1+d_{k}g_{k})(1+d_{k}\tilde{m})^{2}}\right)\right].
\end{align*}
Using Proposition \ref{prop:loose_concentration_Stieltjes_moments}, the absolute value of the r.h.s. can now be bounded by
\begin{align*}
  \frac{d_{k}^{3}\left(2\mathbb{E}\left[\left|g_{k}-\tilde{m}\right|^{3}\right]+d_k\mathbb{E}\left[\left|g_{k}-\tilde{m}\right|^{4}\right]\right)}{\left|1+d_{k}\tilde{m}\right|^{4}}\leq& \frac{d^3_k}{|1+d_k\tilde m|^4}2\left(\frac{2^{3}\boldsymbol{c}_{2}\left(\Tr[T_K]\right)^{4}}{\left|z\right|^{8}N^{6}}+\frac{2^{3}\boldsymbol{c}_{2}}{\left|z\right|^{4}N^{2}}\right)^{\frac{3}{4}}
  \\ &+ \frac{d_k^4}{|1+d_k\tilde m|^4}\left(\frac{2^{3}\boldsymbol{c}_{2}\left(\Tr[T_K]\right)^{4}}{\left|z\right|^{8}N^{6}}+\frac{2^{3}\boldsymbol{c}_{2}}{\left|z\right|^{4}N^{2}}\right)
 \\ \leq& \frac 2 N \frac{d^2_k |\tilde m|^2}{|1+d_k\tilde m|^4}\frac{\Tr[T_K]}{|\tilde m|^2}\left(\frac{2^{\frac 9 4}\boldsymbol{c}_{2}^{\frac 3 4}\left(\Tr[T_K]\right)^{3}}{\left|z\right|^{6}N^{\frac 7 2}}+\frac{2^{\frac 9 4}\boldsymbol{c}_{2}^{\frac 3 4}}{\left|z\right|^{3}N^{\frac 1 2}}\right)
 \\ &+ \frac{1}{N} \frac{d_k^2 |\tilde m|^2}{|1+d_k\tilde m|^4}\frac{(\Tr[T_K])^2}{|\tilde m|^2}\left(\frac{2^{3}\boldsymbol{c}_{2}\left(\Tr[T_K]\right)^{4}}{|z|^{8}N^{5}}+\frac{2^{3}\boldsymbol{c}_{2}}{|z|^{4}N}\right),
\end{align*}
using the inequality $(a + b)^{\frac 3 4}\leq a^{\frac 3 4} + b^{\frac 3 4}$ and the fact that $d_k\leq\Tr[T_K]$. From Proposition \ref{prop:bound-sct-universal}, we have $\frac{1}{\tilde m^2} \leq \left(|z|+\frac{\Tr[T_K]}{N}\right)^2$, so that
\begin{align*}
  \left| \mathbb{E}\left[\left(A_{kk}-\frac{d_{k}\tilde{m}}{1+d_{k}\tilde{m}}\right)^{2}\right] -\mathbb{E}\left[\left(\frac{d_{k}\left[g_{k}-\tilde{m}\right]}{(1+d_{k}\tilde{m})^{2}}\right)^{2}\right] \right| \leq \frac 1 N \frac{d_k^2 |\tilde m|^2}{|1+d_k\tilde m|^4} \mathcal{P}\left(\frac{\Tr[T_K]}{|z|N^{\frac 1 2}}\right).
\end{align*}
This yields the approximation $\mathrm{Var}\left(A_{kk}\right)\approx\frac{d_{k}^{2}\mathbb{E}\left[\left(g_{k}-\tilde{m}\right)^{2}\right]}{(1+d_{k}\tilde{m})^{4}}$.

Using Wick's formula (Lemma \ref{lem:Wicks-formula-matrix}),
\[
  \mathbb{E}\left[\left(g_{k}-\tilde{m}\right)^{2}\right]=\mathbb{E}\left[\left(m_{(k)}-\tilde{m}\right)^{2}\right]+\frac{2}{N}\mathbb{E}[\partial_{z}m_{(k)}(z)],
\]
hence we get:
\[
\frac{d_{k}^{2}\mathbb{E}\left[\left(g_{k}-\tilde{m}\right)^{2}\right]}{(1+d_{k}\tilde{m})^{4}}=\frac{\frac{2}{N}d_{k}^{2}\partial_z\mathbb{E}[ m_{(k)}(z)]}{(1+d_{k}\tilde{m})^{4}}+\frac{d_{k}^{2}\mathbb{E}\left[\left(m_{(k)}-\tilde{m}\right)^{2}\right]}{(1+d_{k}\tilde{m})^{4}}.
\]
Using Proposition \ref{prop:loose_concentration_Stieltjes_moments},
\begin{align*}
  \frac{d_{k}^{2} \mathbb{E}\left[\left|m_{(k)}-\tilde{m}\right|^{2}\right]}{|1+d_{k}\tilde{m}|^{4}}&\leq\frac{d_{k}^{2}}{|1+d_{k}\tilde{m}|^{4}}\left|\frac{\boldsymbol{c}_1(\Tr[T_K])^2}{|z|^4 N^3}\right|
  \\ &\leq \frac 1 N \frac{d_{k}^{2} |\tilde m|^2}{|1+d_{k}\tilde{m}|^{4}}\mathcal{P}\left(\frac{\Tr[T_K]}{|z|N}\right),
\end{align*}
hence the approximation $\mathrm{Var}\left(A_{kk}\right)\approx\frac{\frac{2}{N}d_{k}^{2}\partial_z \mathbb{E}[m_{(k)}(z)]}{(1+d_{k}\tilde{m})^{4}}$.

At last, by using the approximation $\mathbb E [\partial_z m_{(k)}(z)] = \mathbb E [\partial_z g_{k}(z)] \approx \partial_z \tilde m(z)$ (Proposition \ref{cor:loose_concentration-Stieltjes-expectation} and Lemma \ref{lem:cauchy_inequality}), we obtain
\begin{align*}
  &\left|\frac{\frac{2}{N}d_{k}^{2}\partial_z \mathbb{E}[m_{(k)}(z)]}{(1+d_{k}\tilde{m})^{4}}  -\frac{\frac{2}{N}d_{k}^{2}\partial_z\tilde{m}(z)}{(1+d_{k}\tilde{m}(z))^{4}} \right| \\
  &\leq  \frac{2}{N}\frac{d_{k}^{2}}{|1+d_{k}\tilde{m}|^{4}}\frac{2}{-\Re(z)}\left(\frac{2^2\Tr[T_K]}{\left|z\right|^{2}N^{2}}+\frac{2^4\boldsymbol{c}_{1}\left(\Tr[T_K]\right)^{2}}{\left|z\right|^{3}N^{2}}+\frac{2^6\boldsymbol{c}_{1}\left(\Tr[T_K]\right)^{4}}{\left|z\right|^{5}N^{4}}+\frac{2^2}{ |z| N}\right) \\
  &\leq \frac{2}{N}\frac{d_{k}^{2} |\tilde m|^2}{|1+d_{k}\tilde{m}|^{4}}\frac{2|z|}{-\Re(z)}\mathcal{P}\left(\frac{\Tr[T_K]}{|z|N}\right).
\end{align*}
 Hence we get the approximation $\mathrm{Var}\left(A_{kk}\right)\approx\frac{\frac{2}{N}d_{k}^{2}\partial_z \tilde{m}(z)}{(1+d_{k}\tilde{m}(z))^{4}}$, more precisely $  \left|\mathrm{Var}\left(A_{kk}\right)-\frac{\frac{2}{N}d_{k}^{2}\partial_z\tilde{m}(z)}{(1+d_{k}\tilde{m}(z))^{4}} \right| $ is bounded by
\begin{align*}
  &\frac{2}{N}\frac{d_{k}^{2} |\tilde m|^2}{|1+d_{k}\tilde{m}|^{4}}\left(\frac 1 N + \mathcal{P}\left(\frac{\Tr[T_K]}{|z|N^{\frac 1 2}} \right) + \mathcal{P}\left(\frac{\Tr[T_K]}{|z|N}\right) +  \frac{|z|}{-\Re(z)}\mathcal{P}\left(\frac{\Tr[T_K]}{|z|N}\right)\right).
\end{align*}
Putting everything together, we get
\begin{align*}
  \left|\mathrm{Var}\left(A_{kk}\right)-\frac{\frac{2}{N}d_{k}^{2}\partial_z\tilde{m}(z)}{(1+d_{k}\tilde{m}(z))^{4}} \right|  &\leq \frac{2}{N}\frac{d_{k}^{2} |\tilde m|^2}{|1+d_{k}\tilde{m}|^{4}}\left(\frac 1 N + \frac{|z|}{-\Re(z)}\mathcal{P}\left(\frac{\Tr[T_K]}{|z|N^{\frac 1 2}}\right)\right).
\end{align*}

Since $\partial_{z}\vartheta=\frac{\partial_{z}\tilde{m}}{\tilde{m}^{2}}$, from Proposition \ref{prop:bound-sct-universal} we have $|\partial_\lambda \vartheta (\lambda ) | \geq 1 $, i.e. $|\tilde m|^2\leq|\partial_\lambda \tilde m|$ and thus we conclude.

\textbf{Approximation of $\Cov\left(A_{k\ell}(z),A_{\ell k}(z)\right)$:}
Note that $A_{k\ell}(z)=\frac{d_{k}}{N}\mathcal{O}_{.k}^{T}B(z)^{-1}\mathcal{O}_{.\ell}$, hence, since $B(z)$ is symmetric, $$A_{k\ell}(z)=\frac{d_{k}}{d_{\ell}}A_{\ell k}(z).$$ In particular, we have $\Cov\left(A_{k\ell}(z),A_{\ell k}(z)\right)=\frac{d_{\ell}}{d_{k}}\Var\left(A_{k\ell}(z)\right)$. Hence the approximation of $\Cov\left(A_{k\ell}(z),A_{\ell k}(z)\right)$ follows from the one of $\Var\left(A_{k\ell}(z)\right)$.

\textbf{Approximation of $\Var\left(A_{k\ell}(z)\right)$:} We have seen in Theorem \ref{thm:approx_expectation_A} that $\mathbb{E}\left(A_{k\ell}(z)\right)=0$: we need to bound $\mathbb{E}\left(A_{k\ell}(z)^{2}\right)$. Using Equation (\ref{eq:off-diagonal-formula}):
\[
\mathbb{E}\left[A_{k\ell}(z)^{2}\right]=\mathbb{E}\left[\left(\frac{d_{k}}{1+d_{k}g_{k}}\frac{h_{k\ell}}{1+d_{\ell}h_{\ell}}\right)^{2}\right],
\]
where we recall that $h_{\ell}=\frac{1}{N}\mathcal{O}_{.,\ell}^{T}B_{(k\ell)}(z)^{-1}\mathcal{O}_{.,\ell}$ and $h_{k\ell}=\frac{1}{N}\mathcal{O}_{.,k}^{T}B_{(k\ell)}(z)^{-1}\mathcal{O}_{.,\ell}$. Since
\begin{align}\label{eq:akl_with_gk_and_hl}
\frac{d_{k}}{1+d_{k}g_{k}}\frac{h_{k\ell}}{1+d_{\ell}h_{\ell}}=\frac{d_{k}}{1+d_{k}\tilde{m}}\frac{h_{k\ell}}{1+d_{\ell}\tilde{m}}-d_{k}h_{k\ell}\left(\frac{d_{k}\left(g_{k}-\tilde{m}\right)\left(1+d_{\ell}h_{\ell}\right)+d_{\ell}\left(1+d_{k}\tilde{m}\right)\left(h_{\ell}-\tilde{m}\right)}{\left(1+d_{k}\tilde{m}\right)\left(1+d_{\ell}\tilde{m}\right)\left(1+d_{k}g_{k}\right)\left(1+d_{\ell}h_{\ell}\right)}\right),
\end{align}
using Lemma \ref{lem:approx} below, we get the approximation $\mathbb{E}\left[A_{k\ell}(z)^{2}\right]\approx\mathbb{E}\left[\frac{d_{k}^{2}h_{k\ell}^{2}}{\left(1+d_{k}\tilde{m}\right)^{2}\left(1+d_{\ell}\tilde{m}\right)^{2}}\right]$. Using Wick's formula (Lemma \ref{lem:Wicks-formula-matrix} below):
\[
\mathbb{E}\left[h_{k\ell}^{2}\right]=\frac{1}{N}\partial_{z}\mathbb{E}\left[m_{(k\ell)}(z)\right].
\]
Hence the approximation $\mathbb{E}\left[A_{k\ell}(z)^{2}\right]\approx\frac{\frac{1}{N}d_k^2\partial_{z}\mathbb{E}\left[m_{(k\ell)}(z)\right]}{\left(1+d_{k}\tilde{m}\right)^{2}\left(1+d_{\ell}\tilde{m}\right)^{2}}$. At last, by using the approximation $\mathbb E [\partial_z m_{(kl)}(z)] \approx \partial_z \tilde m(z)$  (Lemma \ref{cor:concentration_m_km_and_derivative} above and the technical complex analysis Lemma \ref{lem:cauchy_inequality} below), we can bound the difference $\left|\frac{\frac{1}{N}d_k^2\partial_{z}\mathbb{E}\left[m_{(k\ell)}(z)\right]}{\left(1+d_{k}\tilde{m}\right)^{2}\left(1+d_{\ell}\tilde{m}\right)^{2}}  -\frac{\frac{1}{N}d_{k}^{2}\partial_z\tilde{m}(z)}{\left(1+d_{k}\tilde{m}\right)^{2}\left(1+d_{\ell}\tilde{m}\right)^{2}} \right|$ by
\begin{align*}
   &\frac{1}{N}\frac{d_{k}^{2}}{|1+d_{k}\tilde{m}|^{2}|1+d_{\ell}\tilde{m}|^{2}}\frac{2}{-\Re(z)}\left(\frac{2^2\Tr[T_K]}{\left|z\right|^{2}N^{2}}+\frac{2^4\boldsymbol{c}_{1}\left(\Tr[T_K]\right)^{2}}{\left|z\right|^{3}N^{2}}+\frac{2^6\boldsymbol{c}_{1}\left(\Tr[T_K]\right)^{4}}{\left|z\right|^{5}N^{4}}+\frac{2^2}{ |z| N}\right) \\
  &\leq \frac{1}{N}\frac{d_{k}^{2} |\tilde m|^2}{|1+d_{k}\tilde{m}|^{2}|1+d_{\ell}\tilde{m}|^{2}}\frac{2|z|}{-\Re(z)}\mathcal{P}\left(\frac{\Tr[T_K]}{|z|N}\right)
\end{align*}

Finally, we can bound the error $\left|\mathbb E\left[ \left(A_{k\ell}(z)\right)^2 \right]-\frac{1}{N}\frac{d_{k}^{2}\partial_{z}\tilde{m}}{\left(1+d_{k}\tilde{m}\right)^{2}\left(1+d_{\ell}\tilde{m}\right)^{2}}\right|$ by
\begin{align*}
   &\frac{1}{N}\frac{d_{k}^{2} | \partial_{z}\tilde{m}|}{\left|1+d_{k}\tilde{m}\right|^{2}\left|1+d_{\ell}\tilde{m}\right|^{2}} \left(\mathcal{P}\left(\frac{\Tr[ T_K]}{|z| N^{\frac 1 2}}\right) + \frac{2|z|}{-\Re(z)}\mathcal{P}\left(\frac{\Tr[T_K]}{|z|N^{\frac 1 2}}\right)\right) \\
  & \leq \frac{1}{N}\frac{d_{k}^{2} | \partial_{z}\tilde{m}|}{\left|1+d_{k}\tilde{m}\right|^{2}\left|1+d_{\ell}\tilde{m}\right|^{2}} \frac{|z|}{-\Re(z)}\mathcal{P}\left(\frac{\Tr[T_K]}{|z|N^{\frac 1 2}}\right).
\end{align*}
\end{proof}

\begin{lem}
  \label{lem:approx}
Using the same notation as in the proof of Proposition \ref{prop:variance_entries_A},
\[
\epsilon_{k\ell}  = \mathbb{E}\left[A_{k\ell}(z)^{2}\right] - \frac{d_{k}^2 h^2_{k\ell}}{(1+d_{k}\tilde{m})^2(1+d_{\ell}\tilde{m})^2}
  \]
is bounded by:
\begin{eqnarray*}
\left\vert\epsilon_{k\ell}\right\vert \leq  \frac{1}{N} \frac{d_{k}^{2}\partial_\lambda \tilde{m}}{\left|1+d_{\ell}\tilde{m}\right|^{2}\left|1+d_{k} \tilde{m}\right|^{2}} \mathcal{P}\left(\frac{\Tr[ T_K]}{|z| N^{\frac 1 2}}\right)
\end{eqnarray*}

\end{lem}

\begin{proof}
  Using Equation \ref{eq:akl_with_gk_and_hl}, by setting $c=2\frac{1}{1+d_{k}\tilde{m}}\frac{1}{1+d_{\ell}\tilde{m}}$, $X_{1}=d_{k}h_{k\ell}$, and
  $$X_{2}=\frac{d_{k}\left(g_{k}-\tilde{m}\right)}{\left(1+d_{k}\tilde{m}\right)\left(1+d_{\ell}\tilde{m}\right)\left(1+d_{k}g_{k}\right)}+\frac{d_{\ell}\left(h_{\ell}-\tilde{m}\right)}{\left(1+d_{\ell}\tilde{m}\right)\left(1+d_{k}g_{k}\right)\left(1+d_{\ell}h_{\ell}\right)},$$ we have that $\epsilon_{k\ell}$ is equal to:
\[
\epsilon_{k\ell} = \mathbb{E}\left[ -X_1^2 X_2 (c - X_2)\right]
\]
we can thus control $\epsilon_{k\ell}$ with the following bound
\begin{align*}
\left \vert \epsilon_{k\ell} \right \vert &\leq c \mathbb{E}\left[ \left \vert X_1\right \vert^2 \left \vert X_2\right \vert\right] + \mathbb{E}\left[  \left \vert X_1\right \vert^2 \vert X_2\vert ^2 \right]  \\
&\leq\mathbb{E}\left[\left|X_{1}\right|^{4}\right]^{\frac{1}{2}}\left(c\mathbb{E}\left[\left|X_{2}\right|^{2}\right]^{\frac{1}{2}}+\mathbb{E}\left[\left|X_{2}\right|^{4}\right]^{\frac{1}{2}}\right).
\end{align*}

\begin{itemize}[leftmargin=*]
\item Bound on $\mathbb{E}[\left|X_{1}\right|^{4}]$: using the same argument as for $\mathbb{E}\left[| m_{(k)}-g_{k}|^{2s}\right]$ and Wick's formula (Lemma \ref{lem:Wicks-formula-matrix}), there exists a constant $\boldsymbol{a}$ such that
    \[
  \mathbb{E}\left[\left|X_{1}\right|^{4}\right]^{\frac{1}{2}}=    \mathbb{E}\left[\left|d_{k}h_{k\ell}\right|^{4}\right]^{\frac{1}{2}}=d_{k}^{2}\mathbb{E}\left[\left|h_{k\ell}\right|^{4}\right]^{\frac{1}{2}}\leq\frac{\boldsymbol{a}d_{k}^{2}}{\left|z\right|^{2}N}
  \]
\item Bound on $\mathbb{E}[|X_{2}|^{2s}]$: in order to bound $\mathbb{E}[\left|X_{2}\right|^{2s}]$  we decompose $X_{2}$ as $X_{2}= Y_{1}+ Y_{2}+Y_{3}$ where
\begin{align*}
Y_{1}&=\frac{d_{k}\left(g_{k}-\tilde{m}\right)}{\left(1+d_{k}\tilde{m}\right)\left(1+d_{\ell}\tilde{m}\right)\left(1+d_{k}g_{k}\right)},\\
Y_{2}&=\frac{d_{\ell}\left(h_{\ell}-\tilde{m}\right)}{\left(1+d_{\ell}\tilde{m}\right)\left(1+d_{k}\tilde{m}\right)\left(1+d_{\ell}h_{\ell}\right)}, \\
Y_{3}&=\frac{d_{\ell}d_{k}\left(h_{\ell}-\tilde{m}\right)\left(\tilde{m}-g_{k}\right)}{\left(1+d_{\ell}\tilde{m}\right)\left(1+d_{k}\tilde{m}\right)\left(1+d_{k}g_{k}\right)\left(1+d_{\ell}h_{\ell}\right)},
\end{align*}
so that by Minkowski inequality,
\[
\mathbb{E}\left[\left|X_{2}\right|^{2s}\right]^{\frac{1}{2s}}\leq\mathbb{E}\left[\left| Y_{1}\right|^{2s}\right]^{\frac{1}{2s}}+\mathbb{E}\left[\left| Y_{2}\right|^{2s}\right]^{\frac{1}{2s}}+\mathbb{E}\left[\left| Y_{3}\right|^{2s}\right]^{\frac{1}{2s}},
\]
We can bound the terms in the r.h.s. of the above by applying Proposition \ref{prop:loose_concentration_Stieltjes_moments} and Lemma \ref{cor:concentration_moments_h_K_m_tilde}:
\begin{itemize}[leftmargin=*]
  \item Bound on $\mathbb{E}[\left|Y_{1}\right|^{2s}]$:
  \[
\mathbb{E}\left[\left|Y_{1}\right|^{2s}\right]^{\frac{1}{2s}}\leq \frac{d_{k}}{\left|1+d_{\ell}\tilde{m}\right|\left|1+d_{k}\tilde{m}\right|}\mathbb{E}\left[|\left(g_{k}-\tilde{m}\right)|^{2s}\right]^{\frac{1}{2s}}\leq\frac{d_{k}}{\left|1+d_{\ell}\tilde{m}\right|\left|1+d_{k}\tilde{m}\right|}\left[\frac{2^{2s-1}\boldsymbol{c}_{s}(\Tr[T_{K}])^{2s}}{|z|^{4s}N^{3s}}+\frac{2^{2s-1}\boldsymbol{c}_{s}}{|z|^{2s}N^{s}}\right]^{\frac{1}{2s}}
\]
  \item Bound on $\mathbb{E}[\left|Y_{2}\right|^{2s}]$:
    \[
\mathbb{E}\left[\left|Y_{2}\right|^{2s}\right]^{\frac{1}{2s}} \leq \frac{d_{\ell }}{\left|1+d_{\ell}\tilde{m}\right|\left|1+d_{k}\tilde{m}\right|}\mathbb{E}\left[|\left(h_{\ell}-\tilde{m}\right)|^{2s}\right]^{\frac{1}{2s}}\leq\frac{d_{k}}{\left|1+d_{\ell}\tilde{m}\right|\left|1+d_{k}\tilde{m}\right|}\left[\frac{\boldsymbol{a}_{s}(\Tr[T_{K}])^{2s}}{|z|^{4s}N^{3s}}+\frac{\boldsymbol{b}_{s}}{|z|^{2s}N^{s}}\right]^{\frac{1}{2s}}
\]
  \item Bound on $\mathbb{E}\left[\left|Y_{3}\right|^{2s}\right]^{\frac{1}{2s}}$:
  \begin{align*}
  \mathbb{E}\left[\left|Y_{3}\right|^{2s}\right]^{\frac{1}{2s}}&\leq \frac{d_{\ell}d_{k}}{\left|1+d_{\ell}\tilde{m}\right|\left|1+d_{k}\tilde{m}\right|}\mathbb{E}\left[\left|\left(h_{\ell}-\tilde{m}\right)\right|^{2s}\left|\left(\tilde{m}-g_{k}\right)\right|^{2s}\right]^{\frac{1}{2s}}\\
  &\leq\frac{d_{\ell}d_{k}}{\left|1+d_{\ell}\tilde{m}\right|\left|1+d_{k}\tilde{m}\right|}\mathbb{E}\left[\left|\left(h_{\ell}-\tilde{m}\right)\right|^{4s}\right]^{\frac{1}{4s}}\mathbb{E}\left[\left|\left(\tilde{m}-g_{k}\right)\right|^{4s}\right]^{\frac{1}{4s}}\\
  &\leq\frac{d_{\ell}d_{k}}{\left|1+d_{\ell}\tilde{m}\right|\left|1+d_{k}\tilde{m}\right|}\left[\frac{\boldsymbol{a}_{2s}(\Tr[T_{K}])^{4s}}{|z|^{8s}N^{6s}}+\frac{\boldsymbol{b}_{2s}}{|z|^{4s}N^{2s}}\right]^{\frac{1}{4s}} \left[\frac{2^{4s-1}\boldsymbol{c}_{2s}(\Tr[T_{K}])^{4s}}{|z|^{8s}N^{6s}}+\frac{2^{4s-1}\boldsymbol{c}_{2s}}{|z|^{4s}N^{2s}}\right]^{\frac{1}{4s}}
  \end{align*}
\end{itemize}
\end{itemize}
Let $\boldsymbol{r}_{s}=\max\{2^{2s-1}\boldsymbol{c}_{s},\boldsymbol{a}_{s}\}$ and
$\boldsymbol{t}_{s}=\max\{2^{2s-1}\boldsymbol{c}_{s},\boldsymbol{b}_{s}\} $; then putting the pieces together we have
\[
\mathbb{E}\left[\left|X_{2}\right|^{2s}\right]^{\frac{1}{2s}}\leq\frac{d_{\ell}+d_{k}}{\left|1+d_{\ell}\tilde{m}\right|\left|1+d_{k}\tilde{m}\right|}\left[\frac{\boldsymbol{r}_{s}(\Tr[T_{K}])^{2s}}{|z|^{4s}N^{3s}}+\frac{\boldsymbol{t}_{s}}{|z|^{2s}N^{s}}\right]^{\frac{1}{2s}} +\frac{d_{\ell}d_{k}}{\left|1+d_{\ell}\tilde{m}\right|\left|1+d_{k}\tilde{m}\right|}\left[\frac{\boldsymbol{r}_{2s}(\Tr[T_{K}])^{4s}}{|z|^{8s}N^{6s}}+\frac{\boldsymbol{t}_{2s}}{|z|^{4s}N^{2s}}\right]^{\frac{1}{2s}}
\]
and thus
\[
\mathbb{E}\left[\left|X_{2}\right|^{2}\right]^{\frac{1}{2}}\leq\frac{d_{\ell}+d_{k}}{\left|1+d_{\ell}\tilde{m}\right|\left|1+d_{k}\tilde{m}\right|}\left[\frac{\boldsymbol{r}^{\nicefrac{1}{2}}_{1}(\Tr[T_{K}])}{|z|^{2}N^{\nicefrac{3}{2}}}+\frac{\boldsymbol{t}^{\nicefrac{1}{2}}_{1}}{|z|\sqrt{N}}\right] +\frac{d_{\ell}d_{k}}{\left|1+d_{\ell}\tilde{m}\right|\left|1+d_{k}\tilde{m}\right|}\left[\frac{\boldsymbol{r}^{\nicefrac12}_{2}(\Tr[T_{K}])^{2}}{|z|^{4}N^{3}}+\frac{\boldsymbol{t}^{\nicefrac{1}{2}}_{2}}{|z|^{2}N^{1}}\right]
\]
\[
\mathbb{E}\left[\left|X_{4}\right|^{4}\right]^{\frac{1}{2}}\leq\frac{2 (d_{\ell}+d_{k})^{2}}{\left|1+d_{\ell}\tilde{m}\right|^{2}\left|1+d_{k}\tilde{m}\right|^{2}}\left[\frac{\boldsymbol{r}^{\nicefrac{1}{2}}_{2}(\Tr[T_{K}])^{2}}{|z|^{4}N^{3}}+\frac{\boldsymbol{t}^{\nicefrac{1}{2}}_{2}}{|z|^{2}N}\right] +\frac{2 d_{\ell}^{2}d_{k}^{2}}{\left|1+d_{\ell}\tilde{m}\right|^{2}\left|1+d_{k}\tilde{m}\right|^{2}}\left[\frac{\boldsymbol{r}^{\nicefrac12}_{4}(\Tr[T_{K}])^{4}}{|z|^{8}N^{6}}+\frac{\boldsymbol{t}^{\nicefrac{1}{2}}_{4}}{|z|^{4}N^{2}}\right]
\]
And finally, putting all the pieces together, we have
\begin{align*}
\left\vert\epsilon_{k\ell}\right\vert &\leq\mathbb{E}\left[\left|X_{1}\right|^{4}\right]^{\frac{1}{2}}\left(c\mathbb{E}\left[\left|X_{2}\right|^{2}\right]^{\frac{1}{2}}+\mathbb{E}\left[\left|X_{2}\right|^{4}\right]^{\frac{1}{2}}\right) \\
&\leq \frac{\boldsymbol{a}d_{k}^{2}}{\left|z\right|^{2}N}\frac{2(d_{\ell}+d_{k})}{\left|1+d_{\ell}\tilde{m}\right|^{2}\left|1+d_{k}\tilde{m}\right|^{2}}\left[\frac{\boldsymbol{r}^{\nicefrac{1}{2}}_{1}(\Tr[T_{K}])}{|z|^{2}N^{\nicefrac32}}+\frac{\boldsymbol{t}^{\nicefrac{1}{2}}_{1}}{|z|\sqrt{N}}\right] \\
&\ \ \ \ \  +\frac{\boldsymbol{a}d_{k}^{2}}{\left|z\right|^{2}N}\frac{2(d_{\ell}+d_{k})^{2}+ 2 d_{\ell}d_{k}}{\left|1+d_{\ell}\tilde{m}\right|^{2}\left|1+d_{k}\tilde{m}\right|^{2}}\left[\frac{\boldsymbol{r}^{\nicefrac12}_{2}(\Tr[T_{K}])^{2}}{|z|^{4}N^{3}}+\frac{\boldsymbol{t}^{\nicefrac{1}{2}}_{2}}{|z|^{2}N}\right] \\
&\ \ \ \ \  + \frac{\boldsymbol{a}d_{k}^{2}}{\left|z\right|^{2}N}\frac{2 d_{\ell}^{2}d_{k}^{2}}{\left|1+d_{\ell}\tilde{m}\right|^{2}\left|1+d_{k}\tilde{m}\right|^{2}}\left[\frac{\boldsymbol{r}^{\nicefrac12}_{4}(\Tr[T_{K}])^{4}}{|z|^{8}N^{6}}+\frac{\boldsymbol{t}^{\nicefrac{1}{2}}_{4}}{|z|^{4}N^{2}}\right].
\end{align*}
Using the fact that $| \partial_z \tilde m|\leq |\tilde m|^2$ and Proposition \ref{prop:bound-sct-universal}, we get:
\[
\frac{1}{|z|^2} \leq \frac{| \partial_z \tilde m|}{|z|^2 |\tilde m|^2}\leq | \partial_z \tilde m|\left (1 + 2\frac{\Tr [T_K]}{|z| N}+ \frac{(\Tr [T_K])^2}{|z|^2 N^2}\right ),
\]
we conclude saying that
\begin{eqnarray*}
\left\vert\epsilon_{k\ell}\right\vert \leq  \frac{1}{N} \frac{d_{k}^{2}\partial_z \tilde{m}}{\left|1+d_{\ell}\tilde{m}\right|^{2}\left|1+d_{k} \tilde{m}\right|^{2}} \mathcal{P}\left(\frac{\Tr[ T_K]}{|z| N^{\frac 1 2}}\right).
\end{eqnarray*}
\end{proof}

\begin{rem}
\label{rem:boundswithvartheta}
Since $\tilde{m}(z)=\frac{1}{\vartheta(-z)}$, the derivative $\partial_z\tilde{m}(z)$ can also be expressed in terms of the SCT: $\partial_z\tilde{m}(z) = \partial_z\vartheta(-z)\frac{1}{\vartheta(-z)^2}$, hence the previous approximations can also be written as:
\begin{align*}
\mathrm{Var}\left(A_{kk}(z)\right)\approx  \frac{2}{N}\frac{d_{k}^{2}\vartheta(-z)^{2}\partial_{z}\vartheta(-z)}{(\vartheta(-z)+d_{k})^{4}} \qquad
\mathrm{Var}\left(A_{k\ell}(z)\right)\approx  \frac{1}{N}\frac{d_{k}^{2}\vartheta(-z)^{2}\partial_{z}\vartheta(-z)}{(\vartheta(-z)+d_{k})^{2}(\vartheta(-z)+d_{\ell})^{2}}.
\end{align*}
\end{rem}

We can now describe the variance of the predictor. The variance of the predictor along the eigenfunction $f^{(k)}$ is estimated by $ V_k $, where
\[
V_k(f^*, \lambda, N, \epsilon) = \frac{\psctl}{N}\left(\left\Vert (I_{\mathcal{C}}-\tilde{A}_{\sct})f^{*}\right\Vert_{S}^2+\epsilon ^2 + \bra f^{(k)},f^*\ket_S^2\frac{\vartheta^2(\lambda)}{(\sctl+d_k)^2}\right)\frac{d_k^2}{(\sctl+d_k)^2}.
\]
\begin{thm}
  \label{thm:variance_A} There is a constant $\boldsymbol{C}_1>0$ such that, with the notation of Definition \ref{def:big-p}, we have
  \[
  \left|\mathrm{Var}\left(\bra f^{(k)},\krrf \ket _{S}\right)-V_k(f^*, \lambda, N, \epsilon) \right| \leq  \left(\frac{\boldsymbol{C}_1}{N} + \mathcal{P} \left(\frac{\Tr[T_K]}{\lambda N^{\frac 1 2}} \right) \right) V_k(f^*, \lambda, N, \epsilon).
  \]
  \end{thm}
\begin{proof}
Using the law of total variance, we decompose the variance with respect to the observations $\mathcal{O}$ and the vector of noise $E=(e_{1},\dots,e_{N})^{T}$
\begin{align*}
\mathrm{Var}\left(\left\langle f^{(k)},\hat{f}_{\lambda}^{\epsilon}\right\rangle _{S}\right) & =\mathrm{Var}_{\mathcal{O}}\left(\left\langle f^{(k)},\mathbb{E}_{E}\left[\hat{f}_{\lambda}^{\epsilon}\right]\right\rangle _{S}\right)+\epsilon^{2}\mathbb{E}_{\mathcal{O}}\left[\mathrm{Var}_{E}\left(\frac{d_{k}}{N}\left(\mathcal{O}_{\cdot k}\right)^{T}\left(\frac{1}{N}\mathcal{O}K\mathcal{O}^{T}+\lambda I_{N}\right)^{-1}E\right)\right]\\
 & =\mathrm{Var}_{\mathcal{O}}\left(\left\langle f^{(k)},A(-\lambda)f^{*}\right\rangle _{S}\right)+\epsilon^{2}\mathbb{E}_{\mathcal{O}}\left[\frac{d_{k}}{N}\partial_{\lambda}A_{ kk}(-\lambda)\right].
\end{align*}
Since the randomness is now only on $A$ through $\mathcal O$, from now on, we will lighten the notation by sometimes omitting the $\mathcal O$ dependence in the expectations.

We first show how the approximation $V_k(f^*, \lambda, N, \epsilon)$ appears, and then establish the bounds which allow one to study the quality of this approximation.

\textbf{Approximations:} Decomposing the true function along the principal
components $f^{*}=\sum_{k=1}^{\infty}b_{k}f^{(k)}$ with $b_{k}=\left\langle f^{(k)},f^{*}\right\rangle _{S}$, we have
\begin{align*}
\mathrm{Var}(\langle f^{(k)},A(-\lambda)f^{*}\rangle _{S}) = \sum_{\ell}b_{\ell}^{2}\mathrm{Var}\left(A_{k\ell}(-\lambda)\right).
\end{align*}
From Proposition \ref{prop:variance_entries_A} and the remark after, we have two different approximations for $\mathrm{Var}\left(A_{k\ell}(-\lambda)\right)$. For any $\ell\neq k$, we have
\begin{align*}
\mathrm{Var}\left(A_{kk}(-\lambda)\right)\approx  \frac{2}{N}\frac{d_{k}^{2}\vartheta(\lambda)^{2}\partial_{\lambda}\vartheta(\lambda)}{(\vartheta(\lambda)+d_{k})^{4}}, \qquad
\mathrm{Var}\left(A_{k\ell}(-\lambda)\right)\approx  \frac{1}{N}\frac{d_{k}^{2}\vartheta(\lambda)^{2}\partial_{\lambda}\vartheta(\lambda)}{(\vartheta(\lambda)+d_{k})^{2}(\vartheta(\lambda)+d_{\ell})^{2}}.
\end{align*}
Hence
\begin{align*}
\mathrm{Var}(\langle f^{(k)},A_{\lambda}f^{*}\rangle _{S}) & \approx \frac{b_k^2}{N}\frac{d_{k}^{2}\vartheta(\lambda)^{2}\partial_{\lambda}\vartheta(\lambda)}{(\vartheta(\lambda)+d_{k})^{4}} + \sum_\ell  \frac{b_\ell^2}{N}\frac{d_{k}^{2}\vartheta(\lambda)^{2}\partial_{\lambda}\vartheta(\lambda)}{(\vartheta(\lambda)+d_{k})^{2}(\vartheta(\lambda)+d_{\ell})^{2}}\\
& = \frac{\partial_\lambda \vartheta(\lambda)}{N} \left( \bra f^{(k)},f^*\ket_S^2\frac{\vartheta^2(\lambda)}{(\sctl+d_k)^2} + \sum_\ell  {b_\ell^2}\frac{\vartheta(\lambda)^{2}}{(\vartheta(\lambda)+d_{\ell})^{2}} \right) \frac{d_k^2}{(\sctl+d_k)^2}.
\end{align*}
Since $\sum_\ell  {b_\ell^2}\frac{\vartheta(\lambda)^{2}}{(\vartheta(\lambda)+d_{\ell})^{2}}= \Vert (I_{\mathcal{C}}-\tilde{A}_{\sct})f^{*}\Vert_{S}^2$, this provides the approximation:

\begin{align}
\label{eq:var_predic}
\mathrm{Var}(\langle f^{(k)},A_{\lambda}f^{*}\rangle _{S})\approx \frac{\psctl}{N}\left(\Vert
(I_{\mathcal{C}}-\tilde{A}_{\sct})f^{*}\Vert_{S}^2 + \langle f^{(k)},f^*\rangle_S^2\frac{\vartheta^2(\lambda)}{(\sctl+d_k)^2}\right)\frac{d_k^2}{(\sctl+d_k)^2}.
\end{align}

Now, using Lemma \ref{lem:cauchy_inequality} and Theorem \ref{thm:approx_expectation_A}:
\begin{align}
\label{eq:var_noise}
\epsilon^{2}\mathbb{E}_{\mathcal{O}}\left[\frac{d_{k}}{N}\partial_{\lambda}A_{kk}(-\lambda)\right] \approx \epsilon^2 \frac{\psctl}{N} \frac{d_k^2}{(\sctl+d_k)^2}.
\end{align}

Combining Equations \ref{eq:var_predic} and \ref{eq:var_noise}, we obtain the approximation $$\mathrm{Var}(\langle f^{(k)},\krrf \rangle _{S})\approx V_k(f^*, \lambda, N, \epsilon).$$

Now, we explain how to quantify the quality of the approximations, and thus how to get the bound stated in the theorem. Recall that we decomposed $\mathrm{Var}(\langle f^{(k)},\krrf \rangle _{S})$ into two terms using the law of total variance.

\textbf{First term:} We have seen that:
\begin{align*}
\mathrm{Var}\left(\left\langle f^{(k)},A_{\lambda}f^{*}\right\rangle _{S}\right) = b_{k}^{2}\mathrm{Var}\left(A_{kk}(-\lambda)\right)+\sum_{\ell\neq k}b_{\ell}^{2}\mathrm{Var}\left(A_{k\ell}(-\lambda)\right).
\end{align*}
By Proposition \ref{prop:variance_entries_A}, we have
\begin{align*}
\left| b_{k}^{2}\mathrm{Var}\left(A_{kk}(-\lambda)\right)-2 b_{k}^{2}\frac{\partial_{\lambda}\vartheta(\lambda)}{N}\frac{\vartheta(\lambda)^{2}d_{k}^{2}}{(\vartheta(\lambda)+d_{k})^{4}}\right| & =b_{k}^{2}\left|\mathrm{Var}\left(A_{kk}(-\lambda)\right)-\frac{2}{N}\frac{d_{k}^{2}\partial_{\lambda}\tilde{m}}{(1+d_{k}\tilde{m})^{4}}\right|\\
 & \leq b_{k}^{2}\frac{\left|\partial_{\lambda}\vartheta(\lambda)\right|}{N}\frac{\left|\vartheta(\lambda)\right|^{2}d_{k}^{2}}{\left|\vartheta(\lambda)+d_{k}\right|^{4}}\left(\frac 1 N + \mathcal{P}\left(\frac{\Tr[T_K]}{\lambda N^{\frac 1 2}}\right)\right)
\end{align*}
and
\begin{align*}
\left|b_{\ell}^{2}\mathrm{Var}\left(A_{k\ell}(-\lambda)\right)-\frac{1}{N}b_{\ell}^{2}\frac{d_{k}^{2}\partial_{\lambda}\tilde{m}}{(1+d_{k}\tilde{m})^{2}(1+d_{\ell}\tilde{m})^{2}}\right| & \leq b_{\ell}^{2}\frac{1}{N}\frac{d_{k}^{2}\left|\vartheta(\lambda )\right|^{2}\left|\partial_{\lambda}\vartheta(\lambda)\right|}{\left|\vartheta(\lambda)+d_{k}\right|^{2}\left|\vartheta(\lambda)+d_{\ell}\right|^{2}}\mathcal{P}\left(\frac{\mathrm{Tr}\left[T_{K}\right]}{\lambda N^{\frac{1}{2}}}\right).
\end{align*}
Thus we have
\begin{align*}
 & \left|\sum_{\ell}b_{\ell}^{2}\mathrm{Var}\left(A_{k\ell}(-\lambda)\right)-\frac{\partial_{\lambda}\vartheta(\lambda)}{N}\frac{d_{k}^{2}}{(\vartheta(\lambda)+d_{k})^{2}}\left(2b_{k}^{2}\frac{\vartheta(\lambda)^{2}}{(\vartheta(\lambda)+d_{k})^{2}}+\sum_{\ell\neq k} b_{\ell}^{2}\frac{\vartheta(\lambda)^{2}}{(\vartheta(\lambda)+d_{\ell})^{2}}\right)\right|\\
 & \leq b_{k}^{2}\left|\mathrm{Var}\left(A_{kk}(-\lambda)\right)-\frac{2}{N}\frac{d_{k}^{2}\partial_{\lambda}\tilde{m}}{(1+d_{k}\tilde{m})^{4}}\right|+\sum_{\ell\neq k}b_{\ell}^{2}\left|\mathrm{Var}\left(A_{k\ell}(-\lambda)\right)-\frac{1}{N}\frac{d_{k}^{2}\partial_{\lambda}\tilde{m}}{(1+d_{k}\tilde{m})^{2}(1+d_{\ell}\tilde{m})^{2}}\right|\\
 & \leq b_{k}^{2}\frac{1}{N}\frac{d_{k}^{2}\left|\vartheta(\lambda)\right|^{2}\left|\partial_{\lambda}\vartheta(\lambda)\right|}{\left|\vartheta(\lambda)+d_{k}\right|^{4}}\left(\frac 1 N + \mathcal{P}\left(\frac{\Tr[T_K]}{\lambda N^{\frac 1 2}}\right)\right)+\sum_{\ell \neq k}b_{\ell}^{2}\frac{1}{N}\frac{d_{k}^{2}\left|\vartheta(\lambda )\right|^{2}\left|\partial_{\lambda}\vartheta(\lambda)\right|}{\left|\vartheta(\lambda)+d_{k}\right|^{2}\left|\vartheta(\lambda)+d_{\ell}\right|^{2}}\mathcal{P}\left(\frac{\mathrm{Tr}\left[T_{K}\right]}{\lambda N^{\frac{1}{2}}}\right)\\
& \leq\frac{\left|\partial_{\lambda}\vartheta(\lambda)\right|}{N}\frac{d_{k}^{2}}{\left|\vartheta(\lambda)+d_{k}\right|^{2}}
\sum_{\ell}b_{\ell}^{2}\frac{|\vartheta(\lambda)|^{2}}{|\vartheta(\lambda)+d_{\ell}|^{2}}\left(\frac1N + \mathcal{P}\left(\frac{\mathrm{Tr}\left[T_{K}\right]}{\lambda N^{\frac{1}{2}}}\right)\right)\\
& \leq\frac{\left|\partial_{\lambda}\vartheta(\lambda)\right|}{N}\frac{d_{k}^{2}}{\left|\vartheta(\lambda)+d_{k}\right|^{2}}\left\Vert \left(I_{\mathcal{C}}-\tilde{A}_{\vartheta(\lambda)}\right)f^{*}\right\Vert _{S}^{2}\left(\frac1N+ \mathcal{P}\left(\frac{\mathrm{Tr}\left[T_{K}\right]}{\lambda N^{\frac{1}{2}}}\right)\right).
\end{align*}
We deduce:
\begin{align*}
& \left|\mathrm{Var}_{\mathcal{O}}\left(\left\langle f^{(k)},A_{\lambda}f^{*}\right\rangle _{S}\right)-\frac{\partial_{\lambda}\vartheta(\lambda)}{N}\left(\left\Vert \left(I_{\mathcal{C}}-\tilde{A}_{\lambda}\right)f^{*}\right\Vert _{S}^{2}+\left\langle f^{(k)},f^{*}\right\rangle _{S}^{2}\frac{\vartheta(\lambda)^{2}}{(\vartheta(\lambda)+d_{k})^{2}}\right)\frac{d_{k}^{2}}{(\vartheta(\lambda)+d_{k})^{2}}\right|\\
& \leq\frac{\partial_{\lambda}\vartheta(\lambda)}{N}\left(\left\Vert \left(I_{\mathcal{C}}-\tilde{A}_{\vartheta(\lambda)}\right)f^{*}\right\Vert _{S}^{2}\right)\frac{d_{k}^{2}}{(\vartheta(\lambda)+d_{k})^{2}}\left(\frac1N+ \mathcal{P}\left(\frac{\mathrm{Tr}\left[T_{K}\right]}{\lambda N^{\frac{1}{2}}}\right)\right)\\
& \leq\frac{\partial_{\lambda}\vartheta(\lambda)}{N}\left(\left\Vert \left(I_{\mathcal{C}}-\tilde{A}_{\vartheta(\lambda)}\right)f^{*}\right\Vert _{S}^{2}+\left\langle f^{(k)},f^{*}\right\rangle _{S}^{2}\frac{\vartheta(\lambda)^{2}}{(\vartheta(\lambda)+d_{k})^{2}}\right)\frac{d_{k}^{2}}{(\vartheta(\lambda)+d_{k})^{2}}\left(\frac1N+ \mathcal{P}\left(\frac{\mathrm{Tr}\left[T_{K}\right]}{\lambda N^{\frac{1}{2}}}\right)\right).
\end{align*}
\textbf{Second term: } To approximate, we apply Cauchy's inequality
to Equation (\ref{eq:bound-expectation-diagonal-A}) of Theorem \ref{thm:approx_expectation_A}:
\begin{align*}
\left|\mathbb{E}\left[\partial_{z}A_{kk}(z)\right]-\partial_{z}\vartheta(-z)\frac{d_{k}}{(\vartheta(-z)+d_{k})^{2}}\right| & \leq\frac{2}{-\Re(z)}\sup_{\left|w-z\right|=-\frac{1}{2}\Re(z)}\left|\mathbb{E}[A_{kk}(w)]-\frac{d_{k}}{\vartheta(-w)+d_{k}}\right|\\
& \leq \frac{2}{-\Re(z)}\sup_{\left|w-z\right|=-\frac{1}{2}\Re(z)}\frac{d_{k}\left|\vartheta(-w)\right|}{\left|\vartheta(-w)+d_{k}\right|^{2}}\left(\frac{1}{N}+\mathcal{P}\left(\frac{\mathrm{Tr}\left[T_{K}\right]}{\left|w\right|N}\right)\right).
\end{align*}
By choosing $z=-\lambda$, in the region $\{w\in \mathbb C \mid |w + \lambda|=\frac \lambda2\}$ the polynomial $\mathcal{P}\left(\frac{\mathrm{Tr}\left[T_{K}\right]}{\left|w\right|N}\right)$ is uniformly bounded by $\mathcal{P}\left(\frac{2\mathrm{Tr}\left[T_{K}\right]}{\lambda N}\right)$ and $\frac{d_{k}\left|\vartheta(-w)\right|}{\left|\vartheta(-w)+d_{k}\right|^{2}}\leq \frac{d_{k}\left|\vartheta(\lambda)\right|}{\left|\vartheta(\lambda)+d_{k}\right|^{2}}$. Thus we get
\begin{align*}
\left|\mathbb{E}\left[\partial_{\lambda}A_{kk}(-\lambda)\right]-\partial_{\lambda}\vartheta(\lambda)\frac{d_{k}}{(\vartheta(\lambda)+d_{k})^{2}}\right| & \leq 2 \frac{d_{k}}{\left|\vartheta(\lambda)+d_{k}\right|^{2}}\frac{\vartheta(\lambda)}{\lambda}\left(\frac{1}{N}+\mathcal{P}\left(\frac{\mathrm{Tr}\left[T_{K}\right]}{\lambda N}\right)\right) \\
& \leq 2 \frac{d_{k}}{\left|\vartheta(\lambda)+d_{k}\right|^{2}}\left(1+\frac{\Tr[T_{K}]}{\lambda N}\right)\left(\frac{1}{N}+\mathcal{P}\left(\frac{\mathrm{Tr}\left[T_{K}\right]}{\lambda N}\right)\right) \\
& \leq \frac{d_{k}}{\left|\vartheta(\lambda)+d_{k}\right|^{2}}\left(\frac{2}{N}+\mathcal{P}\left(\frac{\mathrm{Tr}\left[T_{K}\right]}{\lambda N}\right)\right).
\end{align*}
By using the fact that $1\leq\left|\partial_{\lambda}\vartheta(\lambda)\right|$
(see Proposition \ref{prop:bound-sct-universal}), we have that
\[
\left|\frac{d_{k}}{N}\mathbb{E}\left[\partial_{\lambda}A_{kk}(-\lambda)\right]-\frac{\partial_{\lambda}\vartheta(\lambda)}{N}\frac{d_{k}^{2}}{(\vartheta(\lambda)+d_{k})^{2}}\right|\leq\frac{\left|\partial_{\lambda}\vartheta(\lambda)\right|}{N}\frac{d_{k}^{2}}{\left|\vartheta(\lambda)+d_{k}\right|^{2}}\left(\frac{2}{N}+\mathcal{P}\left(\frac{\mathrm{Tr}\left[T_{K}\right]}{\left|z\right|N}\right)\right).
\]
Finally, by putting the bounds for the two terms together we have
\begin{align*}
&\left|\mathrm{Var}\left(\left\langle f^{(k)},\hat{f}_{\lambda}^{\epsilon}\right\rangle _{S}\right) -\frac{\partial_{\lambda}\vartheta(\lambda)}{N}\frac{d_{k}^{2}}{(\vartheta(\lambda)+d_{k})^{2}}\left(2b_{k}^{2}\frac{\vartheta(\lambda)^{2}}{(\vartheta(\lambda)+d_{k})^{2}}+\sum_{\ell\neq k} b_{\ell}^{2}\frac{\vartheta(\lambda)^{2}}{(\vartheta(\lambda)+d_{\ell})^{2}}+\epsilon^{2}\right)\right| \\
& \leq \left|\mathrm{Var}\left(\left\langle f^{(k)},A_{\lambda}f^{*}\right\rangle _{S}\right) -\frac{\partial_{\lambda}\vartheta(\lambda)}{N}\frac{d_{k}^{2}}{(\vartheta(\lambda)+d_{k})^{2}}\left(2b_{k}^{2}\frac{\vartheta(\lambda)^{2}}{(\vartheta(\lambda)+d_{k})^{2}}+\sum_{\ell\neq k} b_{\ell}^{2}\frac{\vartheta(\lambda)^{2}}{(\vartheta(\lambda)+d_{\ell})^{2}}\right)\right|\\
& \qquad\ \ \ \ \ \ \qquad\qquad\qquad\qquad\qquad\qquad\qquad\qquad+ \epsilon^{2}\frac{d_{k}}{N}\left|\partial_{\lambda}\mathbb{E}[A_{kk}(-\lambda)] -\partial_{\lambda}\vartheta(\lambda)\frac{d_{k}}{(\vartheta(\lambda)+d_{k})^{2}}\right| \\
&\leq \frac{\partial_{\lambda}\vartheta(\lambda)}{N}\left(\left\Vert \left(I_{\mathcal{C}}-\tilde{A}_{\vartheta(\lambda)}\right)f^{*}\right\Vert _{S}^{2}+\epsilon^{2}\right)\frac{d_{k}^{2}}{(\vartheta(\lambda)+d_{k})^{2}}\left(\frac2N + \mathcal{P}\left(\frac{\mathrm{Tr}\left[T_{K}\right]}{\lambda N^{\frac{1}{2}}}\right)\right).
\end{align*}
This concludes the proof.
\end{proof}

\subsection{Expected Risk}

We now have all the tools required to describe the expected risk and
empirical risk. In particular, we now show that the distance between the expected risk $\mathbb{E}[R^{\epsilon}(\hat{f}_{\lambda}^{\epsilon})]$ and
\[
\tilde R^{\epsilon}(f^{*},\lambda)=\partial_{\lambda}\vartheta(\lambda)(\|(I_{\mathcal{C}}-\tilde A_{\vartheta(\lambda)})f^{*}\|_{S}^{2}+\epsilon^{2})
\]
is relatively small:
\begin{thm}
\label{thm:appendix-expected-risk}We have
\begin{align*}
\left|\mathbb{E}\left[R^{\epsilon}\left(\hat{f}_{\lambda}^{\epsilon}\right)\right]-\tilde{R}^{\epsilon}\left(f^{*},\lambda\right)\right| & \leq\tilde{R}^{\epsilon}\left(f^{*},\lambda\right)\left(\frac{1}{N}+\mathcal{P}\left(\frac{\mathrm{Tr}\left[T_{K}\right]}{\lambda N^{\frac{1}{2}}}\right)\right).
\end{align*}
\end{thm}
\begin{proof}
The expected risk can be written as $\mathbb{E}[R^\epsilon (\hat{f}_{\lambda}^{\epsilon})]=\mathbb{E}[\Vert\hat{f}_{\lambda}^{\epsilon} - f^*\Vert_S^2]+ \epsilon^2 = \sum_k \mathbb{E}[(a_k - b_k)^2] + \epsilon^2$, where $a_k = \langle f^{(k)},\hat{f}_{\lambda}^{\epsilon}  \rangle_S$ and $b_k = \langle f^{(k)},f^* \rangle_S$. Hence, using the classical bias-variance decomposition for each summand, we get that the expected risk is equal to:
\[
\mathbb{E}[R^{\epsilon}(\hat{f}_{\lambda}^{\epsilon})]=R^{\epsilon}(\mathbb{E}[\hat{f}_{\lambda}^{\epsilon}])+\sum_{k=1}^{\infty}\mathrm{Var}(\langle f^{(k)},\hat{f_{\lambda}^{\epsilon}}\rangle_{S}).
\]

Similarly to the proof of Theorem \ref{thm:variance_A}, we explain how the approximation of the expected arises, then we establish the bounds which allow one to study the quality of this approximation.

\textbf{Approximations:} The bias term $R^{\epsilon}(\mathbb{E}[\hat{f}_{\lambda}^{\epsilon}])$ is equal to $ \Vert \mathbb{E}[\hat{f}_{\lambda}^{\epsilon}]- f^*\Vert_S^2 + \epsilon^2  =  \Vert (I_{\mathcal{C}} - \mathbb{E}[A_\lambda]) f^*\Vert_S^2 + \epsilon^2$. Using Theorem \ref{thm:approx_expectation_A}, one gets the approximation of the bias term:
\begin{align*}
R^{\epsilon}(\mathbb{E}[\hat{f}_{\lambda}^{\epsilon}]) \approx \Vert (I_{\mathcal{C}} - \tilde{A}_{\vartheta(\lambda)}) f^*\Vert_S^2 + \epsilon^2.
\end{align*}
As for the variance term $\sum_{k=1}^{\infty}\mathrm{Var}(\langle f^{(k)},\hat{f_{\lambda}^{\epsilon}}\rangle _{S})$, we use Theorem \ref{thm:variance_A}.

 \[
\sum_{k=1}^{\infty}\mathrm{Var}(\langle f^{(k)},\hat{f_{\lambda}^{\epsilon}}\rangle _{S})\approx \sum_{k=1}^{\infty} V_k(f^*, \lambda, N, \epsilon),
\]
where
\[
V_k(f^*, \lambda, N, \epsilon) =  \frac{\psctl}{N}\left(\left\Vert (I_{\mathcal{C}}-\tilde{A}_{\sct})f^{*}\right\Vert_{S}^2+\epsilon ^2 + \bra f^{(k)},f^*\ket_S^2\frac{\vartheta^2(\lambda)}{(\sctl+d_k)^2}\right)\frac{d_k^2}{(\sctl+d_k)^2}.
\]
Thus the variance term is approximately equal to:
\[
(\Vert (I_{\mathcal{C}}-\tilde{A}_{\sct})f^{*}\Vert_{S}^2+\epsilon ^2 ) \frac{\psctl}{N} \sum_{k=1}^\infty \frac{d_k^2}{(\sctl+d_k)^2} +  \frac{\psctl}{N} \sum_{k=1}^\infty \langle f^{(k)},f^*\rangle_S^2\frac{\vartheta^2(\lambda)d_k^2}{(\sctl+d_k)^4}.
\]
Noting that from Equation \ref{eq:derivative-sctl}, we have
$(\partial_{\lambda}\vartheta(\lambda)-1)=\frac{\partial_{\lambda}\vartheta(\lambda)}{N}\sum_{k=1}^{\infty}\frac{d_{k}^{2}}{(\vartheta(\lambda)+d_{k})^{2}},$ we get:
\[
\sum_{k=1}^{\infty}\mathrm{Var}(\langle f^{(k)},\hat{f_{\lambda}^{\epsilon}}\rangle _{S})\approx (\partial_\lambda \vartheta(\lambda) - 1 ) (\Vert (I_{\mathcal{C}}-\tilde{A}_{\sct})f^{*}\Vert_{S}^2+\epsilon ^2 ) +  \frac{\psctl}{N} \sum_{k=1}^\infty \langle f^{(k)},f^*\rangle_S^2\frac{\vartheta^2(\lambda)d_k^2}{(\sctl+d_k)^4}.
\]
The second term in the r.h.s. is a residual term: using the fact that $\frac{d_k^2}{(\vartheta(\lambda)+d_k)^2}\leq 1$, this term is bounded by $\frac{\partial_{\lambda}\vartheta(\lambda)}{N}\Vert(I_{\mathcal C}-\tilde A_{\vartheta(\lambda)})f^{*}\Vert^{2}_{S}$.

Hence, we get the following approximation of the variance term:
\[
\sum_{k=1}^{\infty}\mathrm{Var}(\langle f^{(k)},\hat{f_{\lambda}^{\epsilon}}\rangle _{S})\approx (\partial_\lambda \vartheta(\lambda) - 1 ) (\Vert (I_{\mathcal{C}}-\tilde{A}_{\sct})f^{*}\Vert_{S}^2+\epsilon ^2 ).
\]

Putting the approximations of the bias and variance terms together, we obtain:
$$\mathbb{E}\left[R^{\epsilon}\left(\hat{f}_{\lambda}^{\epsilon}\right)\right]\approx\tilde{R}^{\epsilon}\left(f^{*},\lambda\right).$$

Now, we explain how to quantify the quality of the approximations, and thus how to get the bound stated in the theorem. Recall that, using the bias-variance decomposition, we split the expected risk into two terms, the bias term and the variance term. We show now that:

\begin{align*}
\left|R^{\epsilon}(\mathbb{E}_{\mathcal{O},E}[\hat{f}_{\lambda}^{\epsilon}])-\left(\Vert (I_{\mathcal{C}}-\tilde{A}_{\vartheta(\lambda)})f^{*}\Vert _{S}^{2}+\epsilon^{2}\right)\right| & \leq\Vert (I_{\mathcal{C}}-\tilde{A}_{\vartheta(\lambda)})f^{*}\Vert _{S}^{2}\left(\frac{1}{N}+\mathcal{P}\left(\frac{\mathrm{Tr}\left[T_{K}\right]}{\lambda N}\right)\right)
\end{align*}
and
\begin{align*}
\Biggl|\sum_{k=1}^{\infty}\mathrm{Var}(\langle f^{(k)},\hat{f_{\lambda}^{\epsilon}}\rangle _{S})-(\partial_{\lambda}\vartheta(\lambda)&-1)\left(\Vert (I_{\mathcal{C}}-\tilde{A}_{\vartheta(\lambda)})f^{*}\Vert _{S}^{2}+\epsilon^{2}\right)\Biggr| \leq \\
&\partial_{\lambda}\vartheta(\lambda)\left(\Vert (I_{\mathcal{C}}-\tilde{A}_{\vartheta(\lambda)})f^{*}\Vert _{S}^{2}+\epsilon^{2}\right)\left(\frac{2}{N}+\mathcal{P}\left(\frac{\mathrm{Tr}\left[T_{K}\right]}{\lambda N^{\frac{1}{2}}}\right)\right).
\end{align*}
Combining the two inequations, and using the fact that $1\leq \partial_\lambda \vartheta(\lambda)$, we then get the desired inequality.

\textbf{Bias term:} Since $|\tilde{A}_{\vartheta(\lambda),kk}|\leq1$, Equation (\ref{eq:bound-expectation-diagonal-A}) of Theorem \ref{thm:approx_expectation_A} implies that
\[
\left|\tilde{A}_{\vartheta(\lambda),kk}-\mathbb{E}\left[A_{kk}(-\lambda)\right]\right|\leq |1-\tilde{A}_{\vartheta(\lambda),kk}|\left(\frac{1}{N} + \mathcal{P}\left(\frac{\Tr[T_K]}{\lambda N}\right)\right).
\]

We then get
\[
    1-\mathbb{E}\left[A_{\lambda,kk}\right]\leq1-\tilde{A}_{\lambda,kk}+\frac{c}{\lambda^{2}N}\left(1-\tilde{A}_{\lambda,kk}\right)\tilde{A}_{\lambda,kk}\leq\left(1-\tilde{A}_{\lambda,kk}\right)\left(1+\frac{c}{\lambda^{2}N}\right).
\]
We decompose the true function $f^{*}$ into
$f^{*}=\sum_{k=1}^{\infty}b_{k}f^{(k)}$ for $b_{k}=\left\langle f^{*},f^{(k)}\right\rangle _{S}$,
and obtain
\begin{align*}
\left|R^{\epsilon}\left(\mathbb{E}_{\mathcal{O},E}\left[\hat{f}_{\lambda}^{\epsilon}\right]\right)-\left(\left\Vert \left(I_{\mathcal{C}}-\tilde{A}_{\vartheta(\lambda)}\right)f^{*}\right\Vert _{S}^{2}+\epsilon^{2}\right)\right| & =\left|\left\Vert \left(I_{\mathcal{C}}-\mathbb{E}\left[A_{\lambda}\right]\right)f^{*}\right\Vert _{S}^{2}-\left\Vert \left(I_{\mathcal{C}}-\tilde{A}_{\vartheta(\lambda)}\right)f^{*}\right\Vert _{S}^{2}\right|\\
 & =\left|\sum_{k=1}^{\infty}b_{k}^{2}\left(\left(1-\mathbb{E}\left[A_{\lambda,kk}\right]\right)^{2}-\left(1-\tilde{A}_{\vartheta(\lambda),kk}\right)^{2}\right)\right|\\
 & \leq\sum_{k=1}^{\infty}b_{k}^{2}\left|\tilde{A}_{\vartheta(\lambda),kk}-\mathbb{E}\left[A_{\lambda,kk}\right]\right|\left|2-\tilde{A}_{\vartheta(\lambda),kk}-\mathbb{E}\left[A_{\vartheta(\lambda),kk}\right]\right|.
\end{align*}
By the triangular inequality, we get that
\[
\left|2-\tilde{A}_{\vartheta(\lambda),kk}-\mathbb{E}\left[A_{\vartheta(\lambda),kk}\right]\right| \leq \left|1-\tilde{A}_{\vartheta(\lambda),kk}\right|\left(2+\left(\frac{1}{N} + \mathcal{P}\left(\frac{\Tr[T_K]}{\lambda N}\right)\right)\right)
\]
and thus
\begin{align*}
& \left|R^{\epsilon}\left(\mathbb{E}_{\mathcal{O},E}\left[\hat{f}_{\lambda}^{\epsilon}\right]\right)-\left(\left\Vert \left(I_{\mathcal{C}}-\tilde{A}_{\vartheta(\lambda)}\right)f^{*}\right\Vert _{S}^{2}+\epsilon^{2}\right)\right| \\ & \leq \sum_{k=1}^{\infty}b_{k}^{2}\left|1-\tilde{A}_{\vartheta(\lambda),kk}\right|^{2} \left(2+\frac{1}{N} + \mathcal{P}\left(\frac{\Tr[T_K]}{\lambda N}\right)\right)\left(\frac{1}{N} + \mathcal{P}\left(\frac{\Tr[T_K]}{\lambda N}\right)\right)\\
& \leq \sum_{k=1}^{\infty}b_{k}^{2}\left|1-\tilde{A}_{\vartheta(\lambda),kk}\right|^{2} \left(\frac{\boldsymbol C_{2}}{N} + \mathcal{P}\left(\frac{\Tr[T_K]}{\lambda N}\right)\right).
\end{align*}
\textbf{Variance term:} For the second term, recall that $
(\partial_{\lambda}\vartheta(\lambda)-1)=\frac{\partial_{\lambda}\vartheta(\lambda)}{N}\sum_{k=1}^{\infty}\frac{d_{k}^{2}}{(\vartheta(\lambda)+d_{k})^{2}},$
and that
\begin{align*}
 & \left|\sum_{k=1}^{\infty}\mathrm{Var}\left(\left\langle f^{(k)},\hat{f_{\lambda}^{\epsilon}}\right\rangle _{S}\right)-(\partial_{\lambda}\vartheta(\lambda)-1)\left(\left\Vert \left(I_{\mathcal{C}}-\tilde{A}_{\vartheta(\lambda)}\right)f^{*}\right\Vert _{S}^{2}+\epsilon^{2}\right)\right|\\
 & \leq\sum_{k=1}^{\infty}\left|\mathrm{Var}_{\mathcal{O}}\left(\left\langle f^{(k)},A(-\lambda)f^{*}\right\rangle _{S}\right)-\frac{\partial_{\lambda}\vartheta(\lambda)}{N}\left(\left\Vert \left(I_{\mathcal{C}}-\tilde{A}_{\vartheta(\lambda)}\right)f^{*}\right\Vert _{S}^{2}+\epsilon^{2}+\left\langle f^{(k)},f^{*}\right\rangle _{S}^{2}\frac{\vartheta(\lambda)^{2}}{(\vartheta(\lambda)+d_{k})^{2}}\right)\frac{d_{k}^{2}}{(\vartheta(\lambda)+d_{k})^{2}}\right|\\
 & +\sum_{k=1}^{\infty}\frac{\partial_{\lambda}\vartheta(\lambda)}{N}\left\langle f^{(k)},f^{*}\right\rangle _{S}^{2}\frac{\vartheta(\lambda)^{2}d_{k}^{2}}{(\vartheta(\lambda)+d_{k})^{4}}.
 \end{align*}
Using Theorem \ref{thm:variance_A}, we can control the terms in the first series: there is a constant $\boldsymbol{C}_1 > 0$ such that
\begin{align*}
&\sum_{k=1}^{\infty}\left|\mathrm{Var}_{\mathcal{O}}\left(\left\langle f^{(k)},A(-\lambda)f^{*}\right\rangle _{S}\right)-\frac{\partial_{\lambda}\vartheta(\lambda)}{N}\left(\left\Vert \left(I_{\mathcal{C}}-\tilde{A}_{\vartheta(\lambda)}\right)f^{*}\right\Vert _{S}^{2}+\epsilon^{2}+\left\langle f^{(k)},f^{*}\right\rangle _{S}^{2}\frac{\vartheta(\lambda)^{2}}{(\vartheta(\lambda)+d_{k})^{2}}\right)\frac{d_{k}^{2}}{(\vartheta(\lambda)+d_{k})^{2}}\right|\\
&\leq \left(\frac{\boldsymbol{C}_1}{N} + \mathcal{P}\left(\frac{\Tr[T_K]}{\lambda N^{\frac 1 2}} \right) \right) \frac{\psctl}{N}\sum_{k=1}^{\infty} \left(\left\Vert (I_{\mathcal{C}}-\tilde{A}_{\sctl})f^{*}\right\Vert_{S}^2+\epsilon ^2 + \bra f^{(k)},f^*\ket_S^2\frac{\vartheta^2(\lambda)}{(\sctl+d_k)^2}\right)\frac{d_k^2}{(\sctl+d_k)^2}\\
&\leq\left(\frac{\boldsymbol{C}_1}{N} + \mathcal{P}\left(\frac{\Tr[T_K]}{\lambda N^{\frac 1 2}} \right) \right)\frac{\psctl}{N} \left(\left\|\left(I_{\mathcal C}-\tilde A_{\vartheta(\lambda)}\right)\tilde A_{\vartheta(\lambda)}f^{*}\right\|^{2}_{S} +\left(\left\Vert (I_{\mathcal{C}}-\tilde{A}_{\sctl})f^{*}\right\Vert_{S}^2+\epsilon ^2 \right)\sum_{k=1}^{\infty}\frac{d_k^2}{(\sctl+d_k)^2}\right)\\
&\leq \left(\frac{\boldsymbol{C}_1}{N} + \mathcal{P} \left(\frac{\Tr[T_K]}{\lambda N^{\frac 1 2}} \right)\right)\left(\frac{\psctl}{N}\left\|\left(I_{\mathcal C}-\tilde A_{\vartheta(\lambda)}\right)\tilde A_{\vartheta(\lambda)}f^{*}\right\|^{2}_{S}+ (\psctl -1)\left(\left\Vert (I_{\mathcal{C}}-\tilde{A}_{\sctl})f^{*}\right\Vert_{S}^2+\epsilon ^2 \right)\right)\\
&\leq \left(\frac{\boldsymbol{C}_1}{N} + \mathcal{P} \left(\frac{\Tr[T_K]}{\lambda N^{\frac 1 2}} \right)\right)\left(\frac{\psctl}{N}\left\|\left(I_{\mathcal C}-\tilde A_{\vartheta(\lambda)}\right)f^{*}\right\|^{2}_{S}+ (\psctl -1)\left(\left\Vert (I_{\mathcal{C}}-\tilde{A}_{\sctl})f^{*}\right\Vert_{S}^2+\epsilon ^2 \right)\right),
\end{align*}
whereas for the second series, as explained already above, we have
\[
\sum_{k=1}^{\infty}\frac{\partial_{\lambda}\vartheta(\lambda)}{N}\left\langle f^{(k)},f^{*}\right\rangle _{S}^{2}\frac{\vartheta(\lambda)^{2}d_{k}^{2}}{(\vartheta(\lambda)+d_{k})^{4}}= \frac{\partial_{\lambda}\vartheta(\lambda)}{N}\left\|\left(I_{\mathcal C}-\tilde A_{\vartheta(\lambda)}\right)\tilde A_{\vartheta(\lambda)}f^{*}\right\|^{2}_{S}\leq \frac{\partial_{\lambda}\vartheta(\lambda)}{N}\left\|\left(I_{\mathcal C}-\tilde A_{\vartheta(\lambda)}\right)f^{*}\right\|^{2}_{S}.
\]
Finally, putting the pieces together, we conclude.
\end{proof}
\subsection{Expected Empirical Risk}

The expected empirical risk can be approximated as follows:
\begin{thm}\label{thm:appendix-expected-empirical-risk}
We have
\[
\left|\mathbb{E}\left[\hat{R}^{\epsilon}\left(\hat{f}_{\lambda,E}^{\epsilon}\right)\right]-\frac{\lambda^{2}}{\vartheta(\lambda)^{2}}\tilde{R}^{\epsilon}\left(f^{*},\lambda\right)\right|\leq\tilde{R}^{\epsilon}\left(f^{*},\lambda\right)\mathcal{P}\left(\frac{\mathrm{Tr}\left[T_{K}\right]}{\lambda N}\right).
\]
\end{thm}
\begin{proof}
A small computation allows one to show that:
\[
\hat{R}^{\epsilon}\left(\hat{f}_{\lambda,E}^{\epsilon}\right)=\frac{\lambda^{2}}{N}\left(y^{\epsilon}\right)^{T}\left(\frac{1}{N}G+\lambda I_{N}\right)^{-2}y^{\epsilon}.
\]
Using the definition of $y^\epsilon$ and the fact that the noise on the labels is centered and independent from the observations, this yields:
\begin{align*}
\mathbb{E}\left[\hat{R}^{\epsilon}\left(\hat{f}_{\lambda,E}^{\epsilon}\right)\right] & =\frac{\lambda^{2}}{N}f^{*}\mathbb{E}\left[\mathcal{O}^{T}\left(\frac{1}{N}G+\lambda I_{N}\right)^{-2}\mathcal{O}\right]f^{*}+\lambda^{2}\epsilon^{2}\mathbb{E}\left[\frac{1}{N}\mathrm{Tr}\left(\frac{1}{N}G+\lambda I_{N}\right)^{-2}\right]\\
 & =\lambda^{2}\sum_{k=1}^{N}\langle f^{(k)},f^{*}\rangle _{S}^{2}\frac{\mathbb{E}\left[\partial_{\lambda}A_{kk}(-\lambda)\right]}{d_{k}}+\lambda^{2}\epsilon^{2}\mathbb{E}\left[\partial_{z}m(-\lambda)\right].
\end{align*}

Similarly to the proof of Theorem \ref{thm:variance_A}, we explain how the approximation of the expected empirical risk appears, then we establish the bounds which allow one to study the quality of this approximation.

\textbf{Approximations:}
Using Equation \ref{eq:var_noise}, $\mathbb{E}\left[\partial_{\lambda}A_{ kk}(-\lambda)\right] \approx \psctl \frac{d_k}{(\sctl+d_k)^2}$ hence
\begin{align*}
\lambda^{2}\sum_{k=1}^{N}\langle f^{(k)},f^{*}\rangle _{S}^{2}\frac{\mathbb{E}\left[\partial_{\lambda}A_{kk}(-\lambda)\right]}{d_{k}} &\approx \frac{\psctl\lambda^{2}}{\sctl^2}\sum_{k=1}^{N}\langle f^{(k)},f^{*}\rangle _{S}^{2}\frac{\sctl^2}{(\sctl+d_k)^2}\\&=\frac{\partial_{\lambda}\vartheta(\lambda)\lambda^{2}}{\vartheta(\lambda)^{2}}\Vert (I_{\mathcal{C}}-\tilde{A}_{\vartheta(\lambda)})f^{*}\Vert _{S}^{2}.
\end{align*}
The second term can be approximated using Proposition \ref{cor:loose_concentration-Stieltjes-expectation} and Lemma \ref{lem:cauchy_inequality}: this yields $$\mathbb{E}\left[\partial_{\lambda}m(-\lambda)\right] \approx \partial_{\lambda}\tilde{m}(-\lambda) =\frac{\partial_{\lambda}\vartheta(\lambda)}{\vartheta(\lambda)^{2}}.$$
Hence, putting the two approximations together, the expected empirical risk is approximated by:
\begin{align*}
\mathbb{E}\left[\hat{R}^{\epsilon}\left(\hat{f}_{\lambda,E}^{\epsilon}\right)\right]\approx \frac{\partial_\lambda \vartheta (\lambda) \lambda^2}{\vartheta(\lambda)^2} \left(\Vert (I_{\mathcal{C}}-\tilde{A}_{\vartheta(\lambda)})f^{*}\Vert _{S}^{2}+\epsilon^2\right)= \frac{ \lambda^2}{\vartheta(\lambda)^2}R^{\epsilon}(f^{*},\lambda).
\end{align*}

Now, we explain how to quantify the quality of the approximations, and thus how to get the bound stated in the theorem. Recall that, we split the expected empirical risk into two terms.

\textbf{First term}: We have already seen in Theorem \ref{thm:variance_A} that by applying Lemma \ref{lem:cauchy_inequality} to Equation (\ref{eq:bound-expectation-diagonal-A}) of Theorem \ref{thm:approx_expectation_A} we get
\begin{align*}
\left|\mathbb{E}\left[\partial_{z}A_{kk}(-\lambda)\right]-\partial_{\lambda}\vartheta(\lambda)\frac{d_{k}}{(\vartheta(\lambda)+d_{k})^{2}}\right| \leq \frac{d_{k}}{|\vartheta(\lambda)+d_{k}|^{2}}\left(\frac2N + \mathcal{P}\left(\frac{\Tr[T_{K}]}{\lambda N}\right)\right)
\end{align*}
and thus
\begin{align*}
 & \left|\lambda^{2}\sum_{k=1}^{N}\left\langle f^{(k)},f^{*}\right\rangle _{S}^{2}\frac{\mathbb{E}\left[\partial_{\lambda}A_{kk}(-\lambda)\right]}{d_{k}}-\frac{\partial_{\lambda}\vartheta(\lambda)\lambda^{2}}{\vartheta(\lambda)^{2}}\left\Vert \left(I_{\mathcal{C}}-\tilde{A}_{\vartheta(\lambda)}\right)f^{*}\right\Vert _{S}^{2}\right|\\
 & \leq\lambda^{2}\sum_{k=1}^{N}\left\langle f^{(k)},f^{*}\right\rangle _{S}^{2}\left|\frac{\mathbb{E}\left[\partial_{\lambda}A_{kk}(-\lambda)\right]}{d_{k}}-\frac{\partial_{\lambda}\vartheta(\lambda)}{\vartheta(\lambda)^{2}}\frac{\vartheta(\lambda)^{2}}{(\vartheta(\lambda)+d_{k})^{2}}\right|\\
 & \leq\lambda^{2}\sum_{k=1}^{N}\left\langle f^{(k)},f^{*}\right\rangle _{S}^{2}\frac{1}{\left|\vartheta(\lambda)+d_{k}\right|^{2}}\left(\frac2N + \mathcal{P}\left(\frac{\Tr[T_{K}]}{\lambda N}\right)\right)\\
 & =\frac{\lambda^{2}}{\vartheta(\lambda)^2}\left\Vert \left(I_{\mathcal{C}}-\tilde{A}_{\vartheta(\lambda)}\right)f^{*}\right\Vert _{S}^{2}\left(\frac2N + \mathcal{P}\left(\frac{\Tr[T_{K}]}{\lambda N}\right)\right)\\
 &\leq\frac{\partial_{\lambda}\vartheta(\lambda)\lambda^{2}}{\vartheta(\lambda)^{2}}\left\Vert \left(I_{\mathcal{C}}-\tilde{A}_{\vartheta(\lambda)}\right)f^{*}\right\Vert _{S}^{2}\left(\frac2N + \mathcal{P}\left(\frac{\Tr[T_{K}]}{\lambda N}\right)\right).
\end{align*}

\textbf{Second Term: } Using Proposition \ref{cor:loose_concentration-Stieltjes-expectation} and Lemma \ref{lem:cauchy_inequality}: 
\begin{align*}
\left|\mathbb{E}\left[\partial_{z}m(z)\right]-\partial_{z}\tilde{m}(z)\right| \leq\frac{\left|z\right|}{-\Re(z)}\left(\frac{2^{3}\mathrm{Tr}\left[T_{K}\right]}{\left|z\right|^{3}N^{2}}+\frac{2^{4}\boldsymbol{c}_{1}\left(\mathrm{Tr}\left[T_{K}\right]\right)^{2}}{\left|z\right|^{4}N^{2}}+\frac{2^{6}\boldsymbol{c}_{1}\left(\mathrm{Tr}\left[T_{K}\right]\right)^{4}}{\left|z\right|^{6}N^{4}}\right).
\end{align*}
Thus, since $\partial_{\lambda}\tilde{m}(-\lambda)=\frac{\partial_{\lambda}\vartheta(\lambda)}{\vartheta(\lambda)^{2}}$,
\[
\left|\lambda^{2}\epsilon^{2}\mathbb{E}\left[\partial_{\lambda}m(-\lambda)\right]-\frac{\lambda^{2}\epsilon^{2}}{\vartheta(\lambda)^{2}}\partial_{\lambda}\vartheta(\lambda)\right|\leq\epsilon^{2}\mathcal{P}\left(\frac{\Tr[T_{K}]}{\lambda N}\right).
\]
\end{proof}

\subsection{Bayesian Setting}\label{subsec:Bayesian-setting}
In this section, we consider the following Bayesian setting: let the true function $f^*$ be random with zero mean and covariance kernel $\Sigma(x,y)=\mathbb E _{f^*}[f^*(x)f^*(y)]$. We will first show that in this setting the KRR predictor with kernel $K=\Sigma$ and ridge $\lambda=\frac {\epsilon^2}{N}$ is optimal amongst all predictors which depend linearly on the noisy labels $\ye$. Second, given a kernel $K$ and a ridge $\lambda$, we provide a simple formula for the expected risk.

Let us consider predictors $\hat f$ that depend linearly on the labels $\ye$, i.e. for all $x$, there is a $M_x \in \mathbb R^N$ such that $\hat f(x)=M_x^T \ye$. Clearly, the KRR predictor belongs to this family of predictors. The pointwise expected squared error can be expressed for any such predictors in terms of the Gram matrix $\OO \Sigma \OO^T+\epsilon^2 I_N$  and the vector $\OO \Sigma(\cdot, x)$
\[
\mathbb E [(M_x^T \ye - f^*(x))^2] = M_x^T (\OO \Sigma \OO^T+\epsilon^2 I_N) M_x - 2 M_x^T \OO \Sigma(\cdot, x) + \Sigma(x,x).
\]
Differentiating w.r.t. $M_x$, we obtain that the above error is minimized when
\[
 M_x = \Sigma(x, \cdot) \OO^T (\OO \Sigma \OO^T+\epsilon^2 I_N)^{-1}.
\]
In other terms, in this Bayesian setting, the KRR predictor with kernel $K=\Sigma$ and ridge $\lambda=\frac{\epsilon^2}{N}$ minimizes the  expected squared error at all points $x$.

Using Theorem \ref{thm:expected-risk}, we obtain the following approximation of the expected risk for a general kernel $K$ and ridge $\lambda$:
\begin{cor} \label{cor:Bayesian-setting} For a random true function of zero mean and covariance kernel $\Sigma$ the expected risk is approximated by
\[
B(\lambda,K;\epsilon^2,\Sigma) = N \vartheta(\lambda,K) + N \partial_\lambda \vartheta(\lambda,K) (\frac{\epsilon^2}{N}-\lambda) +\partial_\tau \vartheta(\lambda,K+\tau (\Sigma - K)) \big|_{\tau=0},
\]
in the sense that
\[
| \mathbb E [R^\epsilon(\krrf)] - B(\lambda,K;\epsilon^2,\Sigma) | \leq B(\lambda,K;\epsilon^2,\Sigma)  \left(\frac{1}{N}+\mathcal{P}\left(\frac{\mathrm{Tr}\left[T_{K}\right]}{\left|z\right|N^{\frac{1}{2}}}\right)\right).
\]

\end{cor}
\begin{proof}
Denoting by $ \mathbb E $ the expectation taken with respect to the data points and the noise, and by $ \mathbb E_{f^*} $ the expectation taken with respect to the random true function $f^*$, from Theorem \ref{thm:expected-risk} we obtain
\begin{align*}
\left|\mathbb E_{f^*} \left[\mathbb{E}\left[R^{\epsilon}\left(\hat{f}_{\lambda}^{\epsilon}\right)\right]\right]-\mathbb E_{f^*} \left[\tilde{R}^{\epsilon}\left(f^{*},\lambda\right)\right]\right| &\leq  \mathbb E_{f^*} \left[ \left|\mathbb{E}\left[R^{\epsilon}\left(\hat{f}_{\lambda}^{\epsilon}\right)\right]-\tilde{R}^{\epsilon}\left(f^{*},\lambda\right)\right| \right] \\
&\leq \mathbb E_{f^*} \left[ \tilde{R}^{\epsilon}\left(f^{*},\lambda\right) \right] \left(\frac{1}{N}+\mathcal{P}\left(\frac{\mathrm{Tr}\left[T_{K}\right]}{\left|z\right|N^{\frac{1}{2}}}\right)\right)
\end{align*}
it therefore suffices to show that $\mathbb E_{f^*} \left[ \tilde{R}^{\epsilon}\left(f^{*},\lambda\right) \right]=B(\lambda,K;\epsilon^2,\Sigma)$.
\begin{align*}
\mathbb E_{f^*} [\tilde R ^\epsilon (f^*,\lambda,N)]&=\partial_\lambda \vartheta(\lambda) \left(\mathbb E_{f^*} \left[ \| (I_\mathcal{C} - \tilde A_\vartheta) f^* \|^2_S\right]  +\epsilon^2 \right) \\
&= \partial_\lambda \vartheta(\lambda,K) \left(\Tr \left[ T_\Sigma (I_\mathcal{C} - \tilde A_\vartheta)^2 \right]  +\epsilon^2 \right) \\
&= \partial_\lambda \vartheta(\lambda,K) \left(\vartheta^{2} \Tr \left[ T_K (T_K + \vartheta(\lambda,K)I_\mathcal{C})^{-2} \right]  +\epsilon^2 \right) \\
&+ \partial_\lambda \vartheta(\lambda,K) \Tr \left[ (T_\Sigma - T_K) (I_\mathcal{C} - \tilde A_\vartheta)^2 \right].
\end{align*}
This formula can be further simplified. First note that differentiating both sides of Equation \ref{eq:t_lambda} w.r.t. to $\lambda$, we obtain that
\[
\frac{\vartheta^{2}}{N}\Tr \left[ T_K (T_K + \vartheta(\lambda,K)I_\mathcal{C})^{-2} \right]= \frac{\vartheta}{\partial_{\lambda}\vartheta}-\lambda.
\]

Secondly, differentiating both sides of Equation \ref{eq:t_lambda}, we obtain, writing $K(\tau) = K + \tau(\Sigma - K)$
\begin{align*}
\partial_{\tau}\vartheta(\lambda,K(\tau)) & =\frac{\partial_{\tau}\vartheta}{N}\mathrm{Tr}\left[\tilde{A}_{\vartheta}\right]+\frac{\vartheta}{N}\mathrm{Tr}\left[\partial_{\tau}\tilde{A}_{\vartheta}\right]\\
 & =\frac{\partial_{\tau}\vartheta}{N}\mathrm{Tr}\left[T_{K}(T_{K}+\vartheta I_{\mathcal{C}})^{-1}\right]+\frac{\vartheta^{2}}{N}\mathrm{Tr}\left[(T_{K}+\vartheta I_{\mathcal{C}})^{-1}T_{(\Sigma-K)}(T_{K}+\vartheta I_{\mathcal{C}})^{-1}\right]\\
 &-\frac{\partial_{\tau}\vartheta\vartheta}{N}\mathrm{Tr}\left[T_{K}(T_{K}+\vartheta I_{\mathcal{C}})^{-2}\right]\\
 & =\frac{\partial_{\tau}\vartheta}{N}\mathrm{Tr}\left[T_{K}(T_{K}+\vartheta I_{\mathcal{C}})^{-1}-\vartheta T_{K}(T_{K}+\vartheta I_{\mathcal{C}})^{-2}\right]+\frac{\vartheta^{2}}{N}\mathrm{Tr}\left[T_{(\Sigma-K)}(T_{K}+\vartheta I_{\mathcal{C}})^{-2}\right]\\
 & =\frac{\partial_{\tau}\vartheta}{N}\mathrm{Tr}\left[T_{K}^{2}(T_{K}+\vartheta I_{\mathcal{C}})^{-2}\right]+\frac{\vartheta^{2}}{N}\mathrm{Tr}\left[T_{(\Sigma-K)}(T_{K}+\vartheta I_{\mathcal{C}})^{-2}\right]\\
 & =\partial_{\tau}\vartheta-\frac{\partial_{\tau}\vartheta}{\partial_{\lambda}\vartheta}+\frac{\vartheta^{2}}{N}\mathrm{Tr}\left[T_{(\Sigma-K)}(T_{K}+\vartheta I_{\mathcal{C}})^{-2}\right],
\end{align*}
where we used the fact that $\frac 1 N \mathrm{Tr}\left[T_{K}^{2}(T_{K}+\vartheta I_{\mathcal{C}})^{-2}\right]=1-1/\partial_\lambda \vartheta$. This implies that
\begin{align*}
\partial_{\tau}\vartheta &=\partial_{\lambda}\vartheta\frac{\vartheta^{2}}{N}\mathrm{Tr}\left[T_{(\Sigma-K)}(T_{K}+\vartheta I_{\mathcal{C}})^{-2}\right] \\
&= \partial_\lambda \vartheta(\lambda,K) \Tr \left[ (T_\Sigma - T_K) (I_\mathcal{C} - \tilde A_\vartheta)^2 \right].
\end{align*}

Putting everything together, we obtain that
\begin{align*}
\mathbb E_{f^*} [\tilde R ^\epsilon (f^*,\lambda,N)]&= N \vartheta(\lambda,K) + N \partial_\lambda \vartheta(\lambda,K) (\frac{\epsilon^2}{N}-\lambda) +\partial_\tau \vartheta(\lambda,K+\tau (\Sigma - K)) \big|_{\tau=0}.
\end{align*}
\end{proof}

\subsection{Technical Lemmas}

\subsubsection{Matricial observations and Wick formula}

For any family $\boldsymbol{A}=\left(A^{(1)},\ldots,A^{(k)}\right)$
of $k$ square matrices of same size, any permutation $\sigma\in\mathfrak{S}_{k}$,
we define:
\[
\sigma\left(\boldsymbol{A}\right)=\prod_{c\text{ cycle of }\sigma}\mathrm{Tr}\left[\prod_{i\in c}A^{(i)}\right],
\]
where the product inside the trace is taken following the order given by
the cycle and, by the cyclic property, does not depend on the starting
point (see \cite{gabriel-15}). For example if $k=4$ and $\sigma$
is the product of transpositions $(1,3)(2,4),$
\[
\sigma\left(\boldsymbol{A}\right)=\mathrm{Tr}(A^{(1)}A^{(3)})\mathrm{Tr}(A^{(2)}A^{(4)}).
\]
 The number of cycles of $\sigma$ is denoted by $\mathrm{c}(\sigma)$.
The set of permutations without fixed points, i.e. such that $\sigma(i)\neq i$
for any $i\in\left[1,\ldots,k\right]$ is denoted by $\mathfrak{S}_{k}^{\dagger}$ and the set of permutations
with cycles of even size is denoted by $\mathfrak{S}_{k}^{\mathrm{even}}$.

The following lemma, which is reminiscent of Lemma 4.5 in \cite{au-18} and which is a
rephrasing of Lemma C.3 of \cite{jacot-2020}, is
a consequence of Wick\textquoteright s formula for Gaussian random
variables and is key to study the $g_{k}$ and $h_{k,\ell}$.
\begin{lem}
\label{lem:Wicks-formula-matrix} If $\boldsymbol{A}=\left(A^{(1)},\ldots,A^{(k)}\right)$
is a family of $k$ square symmetric random matrices of size $P$
independent from a standard Gaussian vector $w$ of size $P$, we have
\begin{align}
  \label{eq:Wick-lem-1}
\mathbb{E}\left[\prod_{i=1}^{k}w^{T}A^{(i)}w\right] & =\sum_{\sigma\in\mathfrak{S}_{k}}2^{k-\mathrm{c}(\sigma)}\mathbb{E}\left[\mathrm{\sigma}\left(\boldsymbol{A}\right)\right],
\end{align}
and,
\begin{align}
  \label{eq:Wick-lem-2}
\mathbb{E}\left[\prod_{i=1}^{k}\left(w^{T}A^{(i)}w-\mathrm{Tr}\left(A^{(i)}\right)\right)\right] & =\sum_{\sigma\in\mathfrak{S}_{k}^{\dagger}}2^{k-\mathrm{c}(\sigma)}\mathbb{E}\left[\mathrm{\sigma}\left(\boldsymbol{A}\right)\right].
\end{align}
Furthermore, if $w$ and $v$ are independent Gaussian vectors of size $P$ and independent from $\boldsymbol{A}$, then
\begin{align}
  \label{eq:Wick-lem-3}
  \mathbb{E}\left[\prod_{i=1}^{k}w^{T}A^{(i)}v\right] & =\sum_{\sigma\in\mathfrak{S}_{k}^{\mathrm{even}}}\mathbb{E}\left[\mathrm{\sigma}\left(\boldsymbol{A}\right)\right].
  \end{align}
\end{lem}

\begin{proof}
The only differences with Lemma C.3 of \cite{jacot-2020} are in the r.h.s.
and the combinatorial sets used to express the left side. We only prove Equation (\ref{eq:Wick-lem-1}); Equations (\ref{eq:Wick-lem-2}) and (\ref{eq:Wick-lem-3}) can be proven similarly. Let $\boldsymbol{P}_{2}(2k)$ be the set of pair partitions of $\left\{ 1,\ldots,2k\right\} $ and let $p\in\boldsymbol{P}_{2}(2k)$. Let $p\left[\boldsymbol{A}\right]=\sum_{\stackrel{p\leq{\rm Ker}(i_{1},\ldots,i_{2k})}{i_{1},\ldots,i_{2k}\in\{1,\ldots,P\}}{}}\mathbb{E}\left[A_{i_{1}i_{2}}^{(1)}\ldots A_{i_{2k-1}i_{2k}}^{(k)}\right]$ where $\leq$ is the coarsed order (i.e. $p\leq q$ if $q$ is coarser than $p$) and where for any $i_{1},\ldots,i_{2k}$ in ${1,...,P}$, $\mathrm{Ker}(i_{1},\ldots,i_{2k})$
is the partition of $\left\{ 1,\ldots,2k\right\} $ such that two
elements $u$ and $v$ in $\left\{ 1,...,2k\right\} $ are in the
same block (i.e. pair) of $\mathrm{Ker}(i_{1},\ldots,i_{2k})$ if
and only if $i_{u}=i_{v}$. By Wick's formula, we have
\[
\mathbb{E}\left[\prod_{i=1}^{k}w^{T}A^{(i)}w\right]=\sum_{p\in \boldsymbol{P}_2(2k)}p\left[\boldsymbol{A}\right];
\]
therefore, it is sufficient to prove that
\[
\sum_{p\in\boldsymbol{P}_{2}(2k)}p\left[\boldsymbol{A}\right]=\sum_{\sigma\in\mathfrak{S}_{k}}2^{k-\mathrm{c}(\sigma)}\mathrm{\sigma}\left(\boldsymbol{A}\right),
\]

Let $\mathrm{Po}$ be the set of polygons on $\{1,\ldots,k\}$, i.e. the set of collections of non-crossing loops (disjoint unoriented cycles) which cover $\{1,\ldots,k\}$. Consider the two maps $F:\boldsymbol{P}_{2}(2k)\to\mathrm{Po}$ and $G:\mathfrak{S}_{k}\mapsto\mathrm{Po}$ obtained by forgetting the underlying structure: for any partition $p\in\boldsymbol{P}_{2}(2k)$, $F(p)$ is the collection of edges $(\ell,m)$ (viewed as collection of non-crossing loops) such that there exists $u\in\left\{ 2\ell-1,2\ell\right\} $ and $v\in\left\{ 2m-1,2m\right\} $ with $\left\{ u,v\right\} \in p$; for any permutation $\sigma\in\mathfrak{S}_{k}$, $G(\sigma)$ is the set of its loops (unoriented cycles).

One can check that for any $\pi\in\mathrm{Po}$,
\[
\#\left\{ p\in\boldsymbol{P}_{2}(2k)\mid F(p)=\pi\right\} =2^{k-\mathrm{c}_{\leq2}(\pi)},\quad\#\left\{ \sigma\in\mathfrak{S}_{k}\mid G(\sigma)=\pi\right\} =2^{\mathrm{c}(\pi)-\mathrm{c}_{\leq2}(\pi)},
\]
where $\mathrm{c}(\pi)$, resp. $\mathrm{c}_{\leq2}(\pi)$, is the number of unoriented cycles, resp. unoriented cycles of size smaller than or equal to $2$, of $\pi$. Note that $\mathrm{c}(\pi)$, resp. $\mathrm{c}_{\leq2}(\pi)$ are also the number of cycles, resp. cycles of size smaller than or equal to $2$ of any $\sigma$ such that $G(\sigma)=\pi$. Notice also that, since the matrices are symmetric, for any $p,p'\in\boldsymbol{P}_{2}(2k)$ and any $\sigma\in\mathfrak{S}_{k}$, if $F(p)=F(p')=G(\sigma)$, then $p\left[\boldsymbol{A}\right]=p'\left[\boldsymbol{A}\right]=\sigma\left[\boldsymbol{A}\right]$.
Hence:
\[
\sum_{p\in\boldsymbol{P}_{2}(2k)}p\left[\boldsymbol{A}\right]=\sum_{p\in\boldsymbol{P}_{2}(2k)}\sum_{\mathrm{\pi}=F(p)}p\left[\boldsymbol{A}\right]=\sum_{\pi\in\mathrm{Po}}\sum_{p : F(p)=\pi}\pi\left[\boldsymbol{A}\right]=\sum_{\pi\in\mathrm{Po}}2^{k-\mathrm{c}_{\leq2}(\pi)}\pi\left[\boldsymbol{A}\right]
\]
hence
\[
\sum_{p\in\boldsymbol{P}_{2}(2k)}p\left[\boldsymbol{A}\right]=\sum_{\pi\in\mathrm{Po}}2^{k-\mathrm{c}_{\leq2}(\pi)}\frac{1}{2^{\mathrm{c}(\pi)-\mathrm{c}_{\leq2}(\pi)}}\sum_{\sigma : G(\sigma)=\pi}\pi\left[\boldsymbol{A}\right]=\sum_{\sigma\in\mathfrak{S}_{k}}2^{k-\mathrm{c}(\pi)}\sigma\left[\boldsymbol{A}\right],
\]
as required.
\end{proof}

\subsubsection{Bound on derivatives}

Given a bound on a holomorphic function, one can obtain a bound on its derivative.

\begin{lem}\label{lem:cauchy_inequality}
Let $f,g:\mathbb{H}_{<0}\to\mathbb{C}$ be two holomorphic functions such that for any $z\in \mathbb H_{<0}$,
\begin{align*}
\left| f(z)-g(z) \right| \leq F(\mid \!z\! \mid),
\end{align*}
where $F:\mathbb{R}^+ \to \mathbb{R}$ is a decreasing function,  then for any  $z\in \mathbb H_{<0}$:
\begin{align*}
\left| \partial_z f(z)-\partial_z g(z) \right| \leq \frac{2}{-\Re(z)} F\left(\frac{\mid \! z\! \mid}{2}\right),
\end{align*}
\end{lem}

\begin{proof}
This is a consequence of Cauchy's inequality: for any $r < -\Re(z)$ (so that the circle of center $z$ and radius $r$ lies inside $\mathbb{H}_{<0}$),
\begin{align*}
\left|\partial_{z}f(z)-\partial_{z}g(z)\right| \leq \frac{1}{r} \sup_{|w-z|=r}\left | f(w)-g(w)\right| \leq  \frac{1}{r} \sup_{|w-z|=r} F(\mid \!w\! \mid).
\end{align*}
The inequality follows by considering $r=-\frac{1}{2}\Re(z)$ and using the fact that $F$ is decreasing.
\end{proof}

\subsubsection{Generalized Cauchy-Schwarz inequality}

Another result that we will use is the following generalization of the Cauchy-Schwarz inequality, which is a consequence of H\"older's inequality.
\begin{lem}
\label{lem:multiple-Cauchy-Schwarz}For complex random variables $a_{1},...,a_{s}$,
we have
\[
\mathbb{E}\left[\left|a_{1}\cdots a_{s}\right|\right]\leq\sqrt[s]{\mathbb{E}\left[\left|a_{1}\right|^{s}\right]\cdots\mathbb{E}\left[\left|a_{s}\right|^{s}\right]}.
\]
\end{lem}
\begin{proof}
The proof is done using an induction argument. The initialization, i.e. when $s=1$, is trivial.

For the induction step, assume that the result is true for $s$ terms and let us prove it for $s+1$ terms. By H\"older's inequality applied for $p=s+1$ and $q=\frac{s+1}{s}$, we obtain:
\begin{eqnarray*}
\mathbb{E}\left[\left|a_{0}a_{1}\cdots a_{s}\right|\right] & \leq & \left(\mathbb{E}\left[\left|a_{0}\right|^{s+1}\right]\right)^{\frac{1}{s+1}}\left(\mathbb{E}\left[\left|a_{1}\cdots a_{s}\right|^{\frac{s+1}{s}}\right]\right)^{\frac{s}{s+1}}\\
 & \leq & \left(\mathbb{E}\left[\left|a_{0}\right|^{s+1}\right]\right)^{\frac{1}{s+1}}\left(\mathbb{E}\left[\left|a_{1}\right|^{s+1}\right]\cdots\mathbb{E}\left[\left|a_{s}\right|^{s+1}\right]\right)^{\frac{1}{s+1}},
\end{eqnarray*}
where the second inequality is obtained by the induction hypothesis.
\end{proof}

\subsubsection{Control on fixed points }
\begin{lem}
\label{lem:distance_cone}Let $z\in\mathbb{H}_{<0}$, let $\left(a_{k}\right)_{k}$ and $\left(b_{k}\right)_{k}$ be sequences of complex numbers in the cone spanned by $1$ and $-\nicefrac{1}{z}$ and let $(d_{k})_{k}$ be positive numbers. Then
\[
\left|z-\sum_{k=1}^{\infty}\frac{d_{k}}{(1+a_{k})(1+b_{k})}\right|\geq\left|z\right|.
\]

\end{lem}
\begin{proof}
For any complex numbers $z_{1}$ and $z_{2}$, let $\Gamma_{z_{1},z_{2}}$ be the cone spanned by $z_{1}$ and $z_{2}$, i.e. $\Gamma_{z_{1},z_{2}}=\{w\in \mathbb C\ : w=a z_1+ b z_2 \text{for}\ a,b\geq0\} $. Since $a_{k},b_{k}\in\Gamma_{1,-\nicefrac{1}{z}}$, $\nicefrac{1}{1+a_{k}}$ and $\nicefrac{1}{1+b_{k}}$ are in $\Gamma_{1,-z}$. All the summands $\frac{d_{k}}{(1+a_{k})(1+b_{k})}$ lie in $\Gamma_{1,z^{2}}$, hence so does $\sum_{k=1}^{\infty}\frac{d_{k}}{(1+a_{k})(1+b_{k})}$. Since $\Re\left(z\right)<0$, the closest point to $z$ in this cone is $0$ and this yields the lower bound:
\[
\left|z-\sum_{k=1}^{\infty}\frac{d_{k}}{(1+a_{k})(1+b_{k})}\right|\geq\left|z\right|,
\]
 hence the result.
\end{proof}
Recall that $\tilde{m}(z)$, resp. $\tilde{m}_{(k)}(z)$, is the unique fixed point of the function $\psi(x):=-\frac{1}{z}\left(1-\frac{1}{N}\sum_{\ell=1}^{\infty}\frac{d_{\ell}x}{1+d_{\ell}x}\right),$ resp. $\psi_{(k)}(x):=-\frac{1}{z}\left(1-\frac{1}{N}\sum_{\ell\neq k}\frac{d_{\ell}x}{1+d_{\ell}x}\right)$, inside the cone spanned by $1$ and $-\nicefrac{1}{z}$. We have the following control on the distance between $\tilde{m}(z)$ and $\tilde{m}_{(k)}(z)$.
\begin{lem}
\label{lem:distance_tilde_m_without_k}For any $z\in\mathbb{H}_{<0}$,
\[
\left|\tilde{m}_{(k)}(z)-\tilde{m}(z)\right|\leq\frac{1}{\left|z\right|N}.
\]
\end{lem}
\begin{proof}
Let $z\in\mathbb{H}_{<0}$ , $\tilde{m}=\tilde{m}(z)$ and $\tilde{m}_{(k)}=\tilde{m}_{(k)}(z)$.
We have:
\begin{align*}
\tilde{m}_{(k)}-\tilde{m} & =-\frac{1}{z}\left(-\frac{1}{N}\sum_{\ell\neq k}\frac{d_{\ell}\tilde{m}_{(k)}}{1+d_{\ell}\tilde{m}_{(k)}}+\frac{1}{N}\sum_{m}\frac{d_{\ell}\tilde{m}}{1+d_{\ell}\tilde{m}}\right)\\
 & =-\frac{1}{z}\left(\frac{1}{N}\sum_{\ell\neq k}^{\infty}\frac{d_{\ell}}{\left(1+d_{\ell}\tilde{m}_{(k)}\right)\left(1+d_{\ell}\tilde{m}\right)}(\tilde{m}-\tilde{m}_{(k)})+\frac{1}{N}\frac{d_{k}\tilde{m}}{1+d_{k}\tilde{m}}\right)
\end{align*}
 which allows us to express the difference $\tilde{m}_{(k)}-\tilde{m}$
as
\[
\tilde{m}_{(k)}-\tilde{m}=\frac{\frac{1}{N}\frac{d_{k}\tilde{m}}{1+d_{k}\tilde{m}}}{\left(\frac{1}{N}\sum_{\ell\neq k}^{\infty}\frac{d_{\ell}}{\left(1+d_{\ell}\tilde{m}_{(k)}\right)\left(1+d_{\ell}\tilde{m}\right)}-z\right)}.
\]
Since $\tilde{m}_{(k)}$ and $\tilde{m}$ lie in the cone spanned by $1$ and $-\frac{1}{z}$, from Lemma \ref{lem:distance_cone}, we have the lower bound on the norm of the denominator:
\[
\left|\frac{1}{N}\sum_{\ell\neq k}^{\infty}\frac{d_{\ell}}{\left(1+d_{\ell}\tilde{m}_{(k)}\right)\left(1+d_{\ell}\tilde{m}\right)}-z\right|\geq\mid z\mid.
\]
 Since $\Re(\tilde{m})\geq0$, $\left|1+d_{k}\tilde{m}\right|\geq\left|d_{k}\tilde{m}\right|$ and hence $\left|\frac{1}{N}\frac{d_{k}\tilde{m}}{1+d_{k}\tilde{m}}\right|\leq\frac{1}{N}$ . This yields the inequality $\left|\tilde{m}_{(k)}(z)-\tilde{m}(z)\right|\leq\frac{1}{N\left|z\right|}.$
\end{proof}

\end{document}